\documentclass{article}

\usepackage[final]{neurips_2025}

\usepackage[utf8]{inputenc} % allow utf-8 input
\usepackage[T1]{fontenc}    % use 8-bit T1 fonts
\usepackage{hyperref}       % hyperlinks
\usepackage{url}            % simple URL typesetting
\usepackage{booktabs}       % professional-quality tables
\usepackage{amsfonts}       % blackboard math symbols
\usepackage{nicefrac}       % compact symbols for 1/2, etc.
\usepackage{microtype}      % microtypography
\usepackage{xcolor}         % colors

\usepackage{graphicx}
\usepackage{xspace}
\newcommand{\ie}{\emph{i.e.}\xspace}
\newcommand{\eg}{\emph{e.g.}\xspace}
\newcommand{\etc}{\emph{etc.}\xspace}

\AtEndPreamble{
    \usepackage[capitalize]{cleveref}
    \crefname{section}{Sec.}{Secs.}
    \Crefname{section}{Section}{Sections}
    \Crefname{table}{Table}{Tables}
    \crefname{table}{Tab.}{Tabs.}
}
\usepackage{tabularx}
\usepackage{multirow}
\usepackage{makecell}
\usepackage{subcaption}
\usepackage{wrapfig}

\title{Autoregressive Adversarial Post-Training \\ for Real-Time Interactive Video Generation}

\author{%
  \textbf{Shanchuan Lin\thanks{Shanchuan Lin: Corresponding author: \href{mailto:peterlin@bytedance.com}{\texttt{peterlin@bytedance.com}}} \quad Ceyuan Yang \quad Hao He\thanks{Hao He: The Chinese University of Hong Kong. Internship at ByteDance Seed.} \quad Jianwen Jiang\thanks{Jianwen Jiang: ByteDance Intelligent Creation Lab.}} \\
  \textbf{Yuxi Ren \quad Xin Xia \quad Yang Zhao \quad Xuefeng Xiao \quad Lu Jiang} \\
  ByteDance Seed\\
  \href{https://seaweed-apt.com/2}{https://seaweed-apt.com/2}
}

\begin{document}

\maketitle

\begin{abstract}
Existing large-scale video generation models are computationally intensive, preventing adoption in real-time and interactive applications. In this work, we propose autoregressive adversarial post-training (AAPT) to transform a pre-trained latent video diffusion model into
a real-time, interactive video generator. Our model autoregressively generates a latent frame at a time using a single neural function evaluation (1NFE). The model can stream the result to the user in real time and receive interactive responses as controls to generate the next latent frame. Unlike existing approaches, our method explores adversarial training as an effective paradigm for autoregressive generation. This not only allows us to design an architecture that is more efficient for one-step generation while fully utilizing the KV cache, but also enables training the model in a student-forcing manner that proves to be effective in reducing error accumulation during long video generation. Our experiments demonstrate that our 8B model achieves real-time, 24fps, streaming video generation at 736$\times$416 resolution on a single H100, or 1280$\times$720 on 8$\times$H100 up to a minute long (1440 frames).
\end{abstract}
In recent years, the field of visual content creation has been transformed by the rise of foundation models for video generation~\cite{brooks2024video,seawead2025seaweed,polyak2024movie,kong2024hunyuanvideo,wang2025wan}. These models have enabled a wide range of powerful applications, including text-to-video generation, image-to-video synthesis, and controllable video creation conditioned on various multi-modal signals.

Building on this progress, researchers are beginning to explore more ambitious applications. One exciting direction is using video generation models as interactive game engines and world simulators~\cite{valevski2024diffusion,bruce2024genie,parkerholder2024genie2,brooks2024video}. Unlike offline video synthesis, interactive video generation requires the model to respond to user inputs in real time and continuously generate coherent video as the world evolves.

While diffusion models produce high-quality videos, they are very expensive for real-time interactive video generation. Early approaches applied diffusion models frame-by-frame~\cite{valevski2024diffusion,yan2023magicprop}. However, these approaches incur high redundancy due to the need to reprocess the context frames at every frame generation step. To address this, diffusion forcing~\cite{chen2024diffusion,xie2024progressive,kim2024fifo,guo2025long} introduced progressive noise to parallelize denoising across frames. Recent work further reduced inference costs by incorporating causal attention, KV caching, and step distillation~\cite{yin2024slow,magi1}, with the current best model~\cite{yin2024slow} achieving four denoising steps.

Meanwhile, token-based autoregressive generation—popularized by large language models (LLMs)~\cite{brown2020language,achiam2023gpt,grattafiori2024llama}—offers an alternative. Models like VideoPoet~\cite{kondratyuk2023videopoet} treat video generation as a next-token prediction task, which can straightforwardly leverage KV caching to improve generation efficiency. However, per-token decoding remains sequential, limiting parallelism and making it difficult to meet real-time demands.

% At its core, interactive video generation must address three main challenges: (1) achieving real-time video generation throughput, (2) maintaining a low latency for interactive signals, and (3) enabling causal video generation of an extended duration. Our work explores adversarial training as a new paradigm for autoregressive video generation, which offers appealing properties for real-time and interactive applications.

In this work, we aim to address the three core challenges of interactive video generation: (1) achieving real-time video generation throughput, (2) maintaining a low latency for interactive signals, and (3) enabling causal video generation of an extended duration. To this end, we explore adversarial training as a new paradigm and propose autoregressive adversarial post-training (AAPT) as an effective strategy for transforming a pretrained video diffusion transformer into a highly efficient autoregressive generator.

Our approach offers several advantages. First, it is fast. Our model autoregressively predicts each latent frame in a single forward pass (1NFE) while fully exploiting the KV cache. Our architecture design further enables 2$\times$ higher efficiency than equivalent diffusion-forcing models distilled to one step. Second, it maintains better quality over long durations. Our adversarial approach enables full student-forcing training, which mitigates error accumulation for long video generation. Furthermore, our student-forcing approach does not require paired ground-truth targets, allowing us to train long video generators and bypass the limitations of short-duration training data. This is important, as single continuous shots of tens of seconds are extremely rare in most datasets.

We demonstrate these benefits empirically. In terms of speed, our 8B-parameter model achieves real-time 24fps video generation at 736$\times$416 resolution on a single H100 GPU, and 1280$\times$720 resolution on 8$\times$H100 GPUs, with a latency of only 0.16 seconds, substantially outperforming CausVid~\cite{yin2024slow}, a 5B model that operates at 640$\times$352 9.4fps with a 1.30-second latency. In terms of duration, our model can generate continuous 60-second (1440-frame) video streams while fully utilizing the KV cache. This significantly exceeds the previous best one-step generator, APT~\cite{lin2025diffusion}, which supports only 49 frames. 

Our experiments focus on the image-to-video (I2V) generation scenario, where the first frame is provided by the user, as most interactive applications adopt this setting. We showcase our method on two interactive applications—pose-conditioned virtual human generation and camera-controlled world exploration—where users can steer video generation in real time through interactive inputs. Evaluations show that our model achieves performance comparable to the state of the art.
\section{Related Work}
\label{sec:related-work}

% \vspace{-5pt}
\paragraph{One-Step Video Generation}

Early video generation models~\cite{brooks2022generating,skorokhodov2022stylegan} using generative adversarial networks (GANs)~\cite{goodfellow2014generative} can achieve fast generation using a single network evaluation. However, the quality, duration, and resolution are poor by modern standards. Diffusion models~\cite{ho2020denoising,song2020score} are the current state-of-the-art, yet their iterative generation process is slow and expensive. Generating a few seconds of high-resolution videos can take minutes. Existing research has attempted to reduce the inference cost by proposing more efficient formulations~\cite{lipman2022flow,liu2022flow,jin2024pyramidal}, samplers~\cite{lu2022dpm,lu2022dpmpp,tian2025training}, architecture~\cite{xie2024sana,zhang2025fast,zhang2024sageattention,zhang2024sageattention2,wang2024lingen,gao2024matten,mo2024scaling}, caching~\cite{ma2024learning,liu2024timestep,zou2024accelerating}, and distillation, \etc In particular, step distillation~\cite{salimans2022progressive,song2023consistency,song2023improved,lu2024simplifying,lin2025diffusion,lin2024sdxl,ren2024hyper,yin2024one,yin2024improved,sauer2024adversarial,sauer2024fast,wang2024animatelcm,lin2024animatediff,liu2022flow,xu2024ufogen,chen2024nitrofusion,luo2023diff,liu2023instaflow,yan2024perflow,kohler2024imagine,kang2024distilling} emerges as one of the most effective approaches and has been widely studied in the image domain and is also adopted in video models. Seaweed~\cite{seawead2025seaweed} and FastHunyuan~\cite{fasthunyuan} report that the generation of 5-second 1280$\times$720 24fps videos can be distilled to 8 or 6 steps without much degradation in quality. For further reduction in steps, SF-V~\cite{zhang2024sf} and OSV~\cite{mao2024osv} explore 2 seconds of 1024$\times$768 7fps image-to-video generation using only a single step. Recently, APT~\cite{lin2025diffusion} achieves real-time text-to-video generation of 2-second 1280$\times$720 24fps videos on 8$\times$H100 GPUs using a single step. This has inspired more downstream applications to explore one-step video generation~\cite{wang2025seedvr2,chen2025dove}. Our method extends adversarial post-training (APT) to the autoregressive video generation scenario.

\vspace{-8pt}
\paragraph{Streaming Long-Video Generation}

Early research in streaming and long video generation~\cite{henschel2024streamingt2v,kodaira2023streamdiffusion,wang2023gen} applies training-free or pipeline approaches on small-scale image and video generation models but is limited in quality. Modern large-scale video diffusion models, \eg MovieGen~\cite{polyak2024movie}, Hunyuan~\cite{kong2024hunyuanvideo}, Wan2.1~\cite{wang2025wan}, and Seaweed~\cite{seawead2025seaweed}, adopt transformer architecture and are trained on much higher resolutions and frame rates. However, due to the quadratic increase in attention computation, these models are commonly trained to only generate videos up to 5 seconds. To support long-video generation, these models are also trained on the video extension task, which gives the model the first few frames as a condition. At inference, this allows the model to extend the generation and stream the result to users as 5-second chunks. The extension can only be performed a few times before the error accumulation catches up. Recent works have also explored architectures with linear complexity to directly generate long videos~\cite{wang2024lingen,gao2024matten,mo2024scaling}, but they are not designed for streaming applications.

More recently, diffusion forcing~\cite{chen2024diffusion} has been proposed for video generation. It assigns progressive noise levels to frame chunks so the decoding proceeds in a causal streaming fashion. Earlier work uses bidirectional attention~\cite{xie2024progressive,kim2024fifo}. Recent works have moved toward causal attention with KV cache~\cite{yin2024slow,chen2025skyreels,magi1,guo2025long}. Most notably, SkyReel-2~\cite{chen2025skyreels} and MAGI-1~\cite{magi1} are diffusion-forcing video generation models trained from scratch. CausVid~\cite{yin2024slow} explores converting existing bidirectional video diffusion models to causal diffusion-forcing generators. Some of these methods also apply step distillation to improve speed. MAGI-1~\cite{magi1} distills the model to 8 steps and outputs 24 frames as a chunk. It reports real-time 1280$\times$720 24fps generation on 24$\times$H100 GPUs. However, this amount of computation limits wide adoption. CausVid~\cite{yin2024slow} distills the model to 4 steps and outputs 16 frames as a chunk. It can generate 640$\times$320 videos at 9.4fps on a single H100 GPU. In comparison, our method is significantly faster. Our model uses only a single step and achieves 24fps streaming at 736$\times$416 resolution on a single H100 GPU, or 1280$\times$720 on 8$\times$H100 GPUs. Moreover, ours generates a single latent frame (4 video frames) at a time to minimize latency.

It is important to note that these diffusion-forcing models are still only trained up to a fixed-duration window, \eg 5 seconds. Early approaches without KV cache can run a sliding window, but this becomes an issue for KV cache because the receptive field grows indefinitely. Applying a sliding window and dropping out KV tokens can't help because the remaining tokens in the cache were computed in the past and still carry the receptive field. Naive extrapolation at inference leads to out-of-distribution behaviors. Therefore, methods like CausVid~\cite{yin2024slow}, SkyReel-V2~\cite{chen2025skyreels}, and MAGI-1~\cite{magi1} still need to apply the extension technique at inference by restarting and re-computing some overlapping context frames to generate long videos. Except that the diffusion forcing objective naturally supports input tokens with different noise levels, so the context frames can be given as clean latent frames at the beginning, with no additional training necessary. However, this is not ideal as it causes wait time on real-world streaming applications. In contrast, our method supports streaming generations of minute-long videos using KV cache without stopping and reprocessing.

% \vspace{-8pt}
\paragraph{LLMs for Video Generation}

Large language models (LLMs)~\cite{brown2020language,achiam2023gpt,grattafiori2024llama} have widely adopted the causal transformer architecture~\cite{vaswani2017attention} for autoregressive generation. Most notably, attention is masked to prevent attending to future tokens, the inputs are past predictions, and the output targets are shifted by one for predicting the next tokens. Recent research has shown that images and videos can also be generated in such an autoregressive fashion~\cite{team2024chameleon,wang2024emu3,wu2024janus,chen2025janus}. Although causal generation with KV cache is computationally efficient, generating token-by-token prevents parallelization and is slow for high-resolution generation. Some research has explored the decoding of multiple tokens at once during inference~\cite{wang2024parallelized,ren2025next,ye2025fast}, but there is a tradeoff for quality, and it is challenging to decode an entire frame at once. Our architecture is inspired by LLMs, but ours generates a frame of tokens at a time, trained using an adversarial objective. This is optimized for fast generation.

% \vspace{-8pt}
\paragraph{Interactive Video Generation}

Our paper showcases our model's real-time interactive generation ability on two applications: pose-controlled virtual human video generation and camera-controlled world exploration. We briefly introduce the related works in each subfield.

Recent research has explored the use of video generation models to create interactive environments for gameplay and world simulation~\cite{alonso2024diffusion,bruce2024genie,valevski2024diffusion,guo2025mineworld,parkerholder2024genie2,feng2024matrix,oasis2024}. Typically, the first frame is given, and the model continuously predicts the next frame given user control (image-to-world). The control can be the discrete states in an action space or general-purpose camera position embeddings~\cite{he2024cameractrl,he2025cameractrl}. However, the high computation cost of the existing video generation approaches greatly limits the resolution and frame rates. For example, GameNGen~\cite{valevski2024diffusion} and MineWorld~\cite{guo2025mineworld} only generate videos around 320$\times$240 resolution at 6$\sim$20fps with small models of a few hundred million parameters. Recent works, \eg Genie-2~\cite{parkerholder2024genie2}, Oasis~\cite{oasis2024}, Matrix~\cite{feng2024matrix}, \etc, have moved toward large-scale architectures and higher resolutions. Though many report their methods can operate in real-time, the specific hardware requirements are not specified.

Interactive video generation also holds significant potential in the domain of virtual humans. Typically, the first frame is given to establish the identity, then the pose~\cite{hu2024animate,luo2025dreamactor} or other multimodal~\cite{jiang2024loopy,lin2025omnihuman,lin2024cyberhost,tian2024emo} conditions are given to drive the subject. Existing works employ diffusion models with the extension technique to generate long videos~\cite{stypulkowski2024diffused}. The inference speed remains a major bottleneck that limits their applicability to offline human video generation tasks.

\section{Method}

\begin{figure*}
    \centering
    \includegraphics[width=\linewidth]{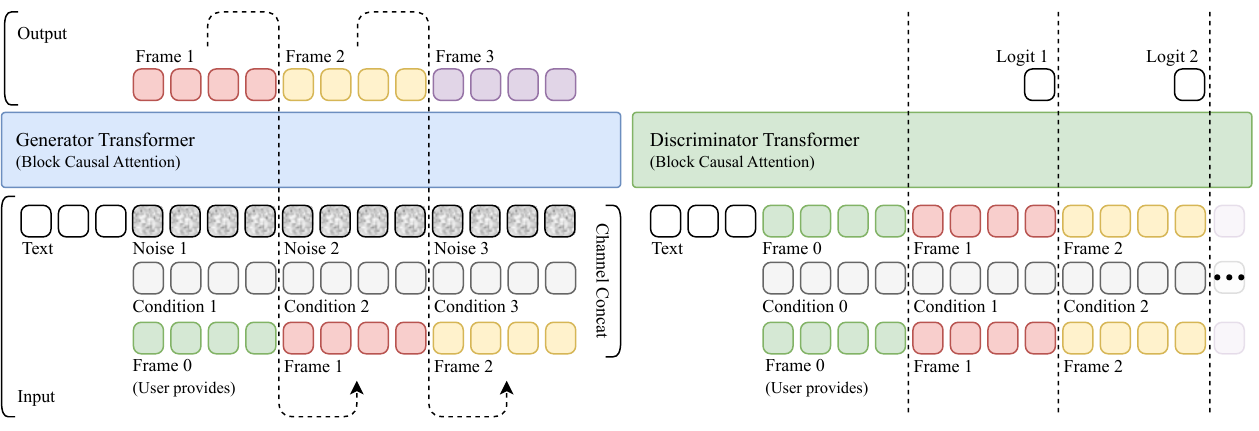}
    \caption{\textbf{Generator (left) } is a block causal transformer. The initial frame 0 is provided by the user at the first autoregressive step, along with text, condition, and noise as inputs to the model to generate the next frame in a single forward pass. Then, the generated frame is recycled as input, along with new conditions and noise, to recursively generate further frames. KV cache is used to avoid recomputation of past tokens. A sliding window is used to ensure constant speed and memory for the generation of arbitrary lengths. \textbf{Discriminator (right)} uses the same block causal architecture. Condition inputs are shifted to align with the frame inputs. Since it is initialized from the diffusion weights, we replace the noise channels with frame inputs following APT.}
    \label{fig:architecture}
    \vspace{-10pt}
\end{figure*}

\begin{wrapfigure}[12]{r}{0.35\textwidth}
    \vspace{-40pt}
    \centering
    \includegraphics[width=\linewidth]{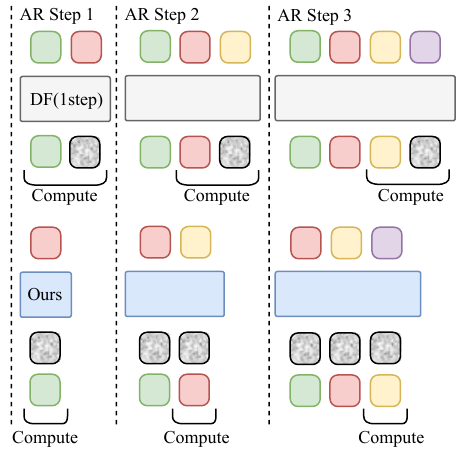}
    \caption{Ours is more efficient than one-step diffusion forcing (DF).}
    \label{fig:architecture-compare-df}
\end{wrapfigure}

Our objective is to transform a pre-trained video diffusion model into a fast, per-latent-frame causal generator suitable for real-time interactive applications. We achieve this through a new method called autoregressive adversarial post-training (AAPT). This section discusses AAPT's architectural transformations and training procedures.

\subsection{Causal Architecture}
\label{sec:method-architecture}

We build our method on a pre-trained video diffusion model that employs a diffusion transformer (DiT)~\cite{peebles2023scalable} architecture and operates in a spatially and temporally compressed latent space through a 3D variational autoencoder (VAE)~\cite{yu2023language}. Since our model operates in the latent space, we will refer to latent frames simply as frames unless otherwise specified. Our diffusion transformer has 8 billion (8B) parameters. It takes text embedding tokens, noisy visual tokens, and diffusion timesteps as input, and calculates bidirectional full attention over all the text and video tokens.

First of all, we transform the bidirectional DiT into a causal autoregressive architecture by replacing full attention with block causal attention. Specifically, text tokens only attend to themselves, and visual tokens attend to text tokens and visual tokens of previous and current frames. Afterward, we change the model inputs. As illustrated in \cref{fig:architecture}, in addition to the regular noise and conditional inputs used by the original diffusion model, we change the model to also take in the past generated frame from the previous autoregressive step through channel concatenation, except the first autoregressive step where the input frame given by the user is used instead. During inference, our model runs autoregressively. At each autoregressive step, it reuses the attention KV cache and generates the next frame in a single forward pass. The generated frame is recycled, along with a new control condition, as inputs for the next autoregressive step.

To prevent the unbounded growth of attention computation and KV cache size,
visual tokens attend to at most $N$ past frames while always attending to the text tokens and the first frame. It is worth noting that although each attention layer uses a window size of $N$, stacking multiple layers results in a much larger effective receptive field.

Our architecture resembles that of large language models (LLMs), but with one important distinction: unlike conventional next-token prediction that outputs the token probabilities using a softmax layer, our model generates all tokens for the next frame in a single forward pass sampled by noise. In addition, our input recycling approach is also more efficient than the one-step diffusion forcing, as shown in \cref{fig:architecture-compare-df}. Diffusion forcing is not optimized for the one-step generation scenario. When using KV cache, diffusion forcing requires computation on two frames on every autoregressive step, while ours only needs one.

\subsection{Training Procedure}
\label{sec:method-procedure}

To create a one-step, per-frame, autoregressive generator, our training process involves three sequential stages: (1) diffusion adaptation, (2) consistency distillation, and (3) adversarial training.

\vspace{-5pt}
\paragraph{Diffusion adaptation} We load the pre-trained weights and finetune the model with the diffusion objective for architectural adaptation. We apply teacher-forcing training, where the ground-truth frames from the dataset are given as past-frame inputs. The output target is shifted by one frame to let the model perform next-frame prediction. Instead of pure noise, the noisy latent and the diffusion timestep $t\sim \mathcal{U}(0, T)$ are still used per regular diffusion training. The same noise level is applied for all frames. This resembles LLMs training, where all the autoregressive steps are trained in parallel.

\vspace{-5pt}
\paragraph{Consistency distillation} We apply consistency distillation~\cite{song2023consistency} before adversarial training as an initialization step to accelerate convergence following APT~\cite{lin2025diffusion}. Our modified formulation and architecture are fully compatible with the original consistency distillation process without the need for modification. We omit classifier-free guidance (CFG) as we find it introduces artifacts in our autoregressive generation setting.

\vspace{-5pt}
\paragraph{Adversarial training} We extend APT~\cite{lin2025diffusion} to the autoregressive setting with improved discriminator design, training strategy, and loss objective.

For the discriminator model, we use the same causal generator architecture as our discriminator backbone, initialize it from the diffusion weights post-adaptation, and insert logit output projection layers. We replace the noise input to frames and randomly sample timestep $t\sim\mathcal{U}(0,T)$ for fast adaptation. A notable difference to APT discriminator design is that ours computes output logit for every frame instead of for the whole clip. This design naturally enables parallel multi-duration discrimination, as inspired by multi-resolution discrimination~\cite{karras2017progressive,karnewar2020msg}.

We find models trained with teacher-forcing incur significant error accumulation at inference. 
To address this, we introduce a student-forcing approach within the adversarial training framework.
Specifically, the generator only uses the ground-truth first frame and recycles the actual generated results as input for the next autoregressive step. In each training step, the generator is autoregressively invoked with KV cache to produce the video, exactly matching the inference behavior, while the discriminator evaluates all the generated frames in a single forward pass in parallel. We find detaching the pass-frame input from the gradient graph improves stability. We allow the gradient to flow through the KV cache to update all the parameters.

For the loss, we use R3GAN~\cite{huang2024gan} objective as our preliminary experiments find that it is more stable than the non-saturating loss~\cite{goodfellow2014generative}. Specifically, we adopt the relativistic loss~\cite{jolicoeur2018relativistic} and apply both the approximated R1 and R2 regularizations~\cite{roth2017stabilizing,mescheder2018training} as proposed in APT~\cite{lin2025diffusion}.

\paragraph{Long-Video Training}
\label{sec:method-extension}

For the model to learn continuous generation of long videos, one must train it on single-shot videos of long duration (\eg, 30–60 seconds). However, such long single-shot videos are rare in most training datasets, where the average shot duration is only 8 seconds. The lack of long-duration training leads to poor temporal extrapolation during inference.

To address the data limitation, we let the generator produce a long video, \eg 60 seconds, and break it down into short segments, \eg 10 seconds, for discriminator evaluation. We keep an overlapping 1-second duration for discriminator evaluation to encourage segment continuation. The discriminator is trained on generated segments and real videos from the dataset. This objective ensures that every segment of a generated long video fits the data distribution. 

To fit the GPU memory, we also let the generator only produce a segment at a time to be evaluated by the discriminator. To produce the next segment, the generator reuses the detached KV cache from the last segment. The gradient is backpropagated after every segment evaluation for loss accumulation.

This technique can be used to train very long generators, with the trade-off of an increase in training time. We find this technique significantly improves the quality of long-duration video generation. This is made possible by the discriminator in adversarial training. Unlike supervised objectives that require ground-truth targets, the discriminator does not need explicit supervision for each input frame. Instead, it learns to distinguish real videos from generated ones. As a result, the model can learn from every video sample, rather than relying on a limited number of long-duration videos.

\subsection{Interactive Generation Applications}
\label{sec:method-application}

We first train a model for the general image-to-video generation task without interactive conditions. This allows us to evaluate the generation quality on standard benchmarks. We then train two separate models on the pose-conditioned human generation task and the camera-conditioned world exploration task. This allows us to evaluate the controllability using two distinct condition signals. For the pose-conditioned human video generation task, we extract and encode the human pose from the training videos and provide it as a per-frame condition to the model following~\cite{lin2025omnihuman}. Similarly, for the camera-conditioned world exploration generation task, we follow~\cite{he2025cameractrl} to extract and encode the camera origin and orientation as Plücker embeddings, with a few modifications to have it better support causal generation. We use similar training datasets as used in these prior works~\cite{lin2025omnihuman,he2025cameractrl}. We refer readers to our supplementary materials for additional details on our architecture, implementation, and training parameters.
\section{Evaluation}

\paragraph{Experimental Setups}

We use causal 3D convolution VAE~\cite{yu2023language} to compress the video temporally by 4 and spatially by 8. Therefore, our model autoregressively generates 4 video frames. The first input frame is independently compressed as a latent frame by the VAE. Since our VAE is causal, it naturally supports streaming decoding. We use attention window size $N=30$ to attend to 30 latent frames (5 seconds). Additional details on the training setup are provided in the supplementary materials.

\paragraph{Baseline and Metrics}
Following prior work~\cite{yin2024slow}, we evaluate our method on the standard VBench-I2V benchmark~\cite{huang2024vbench} on both 120-frame short-video generation and 1440-frame long-video generation. For comparison, we select CausVid~\cite{yin2024slow}, Wan2.1~\cite{wang2025wan}, Hunyuan~\cite{kong2024hunyuanvideo}, MAGI-1~\cite{magi1}, SkyReel-V2, and our own diffusion model as baseline. These models are selected because CausVid is the state-of-the-art for fast streaming generation, and other models are available open-source video generation foundational models that support I2V. Note, CausVid is a closed-source model and only reports VBench-I2V for 120-frame 12fps generation. Wan2.1 and Hunyuan are bidirectional diffusion models that only support up to 120-frame generation. MAGI-1 and SkyReel-V2 are diffusion-forcing models that support arbitrary-length streaming decoding, so we include them for the 1440-frame comparison. Our model is evaluated and compared at 736$\times$416 resolution. Additional inference settings and 1280$\times$720 results are provided in the supplementary materials.

\begin{figure*}[t]
    \centering
    \small
    \captionsetup{justification=raggedright,singlelinecheck=false}
    \begin{tabularx}{\linewidth}{|X|X|X|X|X|X|X|}
        0s (Input) & 10s & 20s & 30s & 40s & 50s & 60s
    \end{tabularx}
    \begin{subfigure}[b]{\textwidth}
        \centering
        \setlength\tabcolsep{1pt}
        \begin{tabularx}{\linewidth}{XXXXXXX}
            \includegraphics[width=\linewidth]{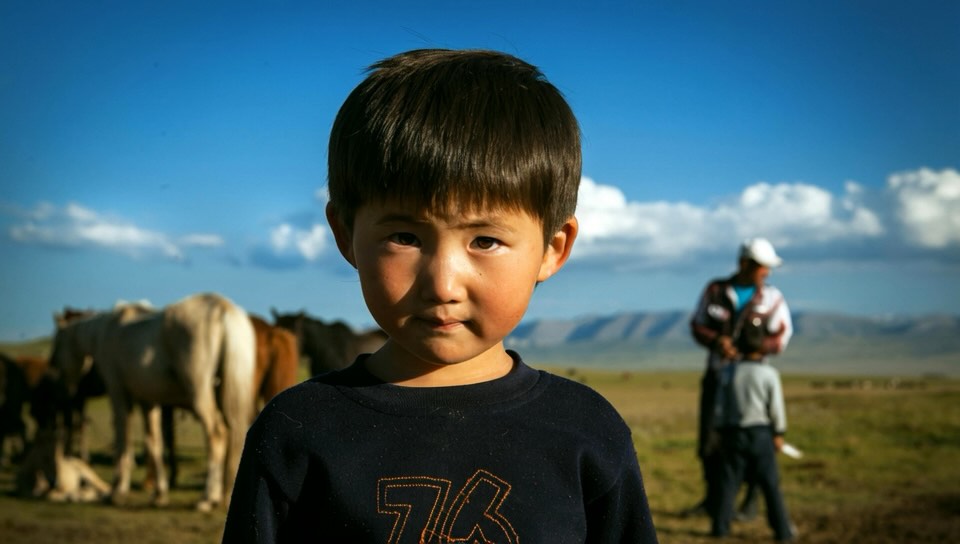} &
            \includegraphics[width=\linewidth]{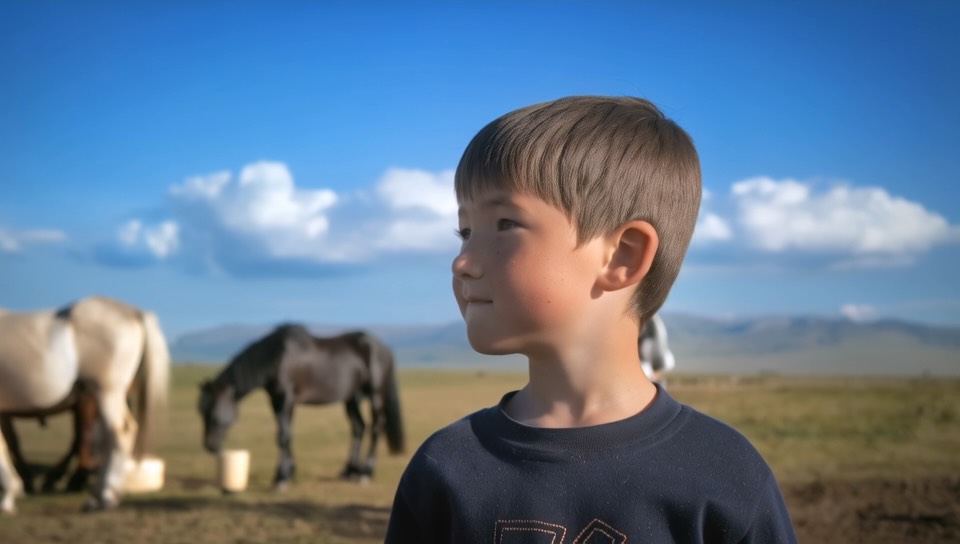} &
            \includegraphics[width=\linewidth]{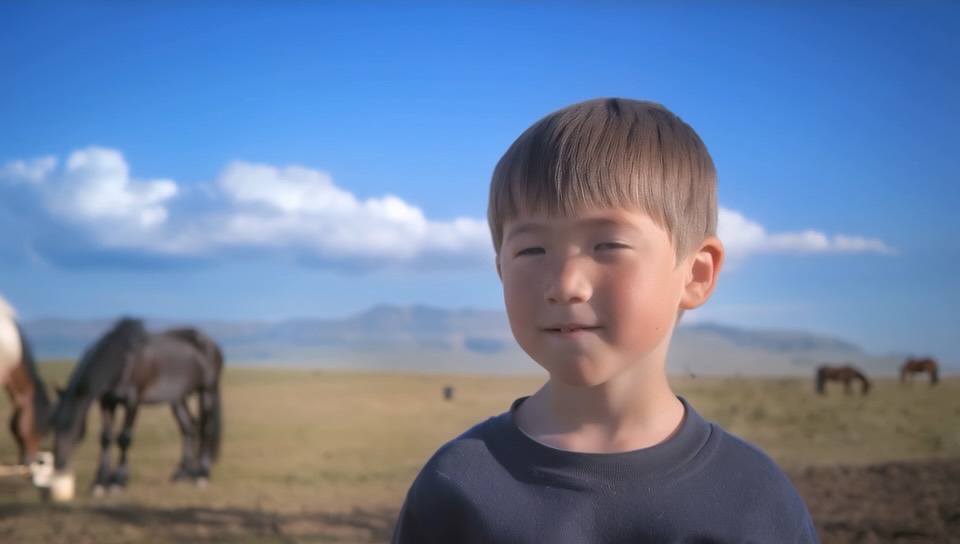} &
            \includegraphics[width=\linewidth]{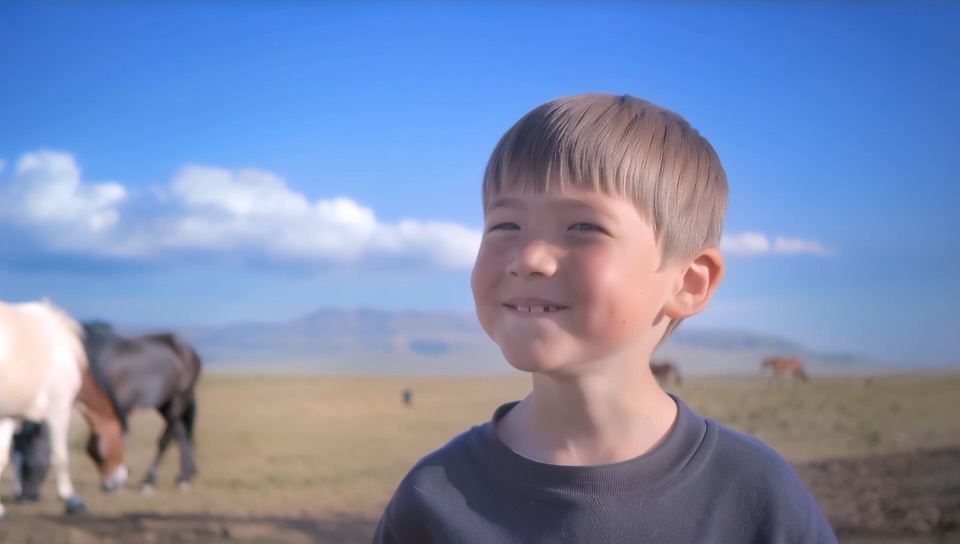} &
            \includegraphics[width=\linewidth]{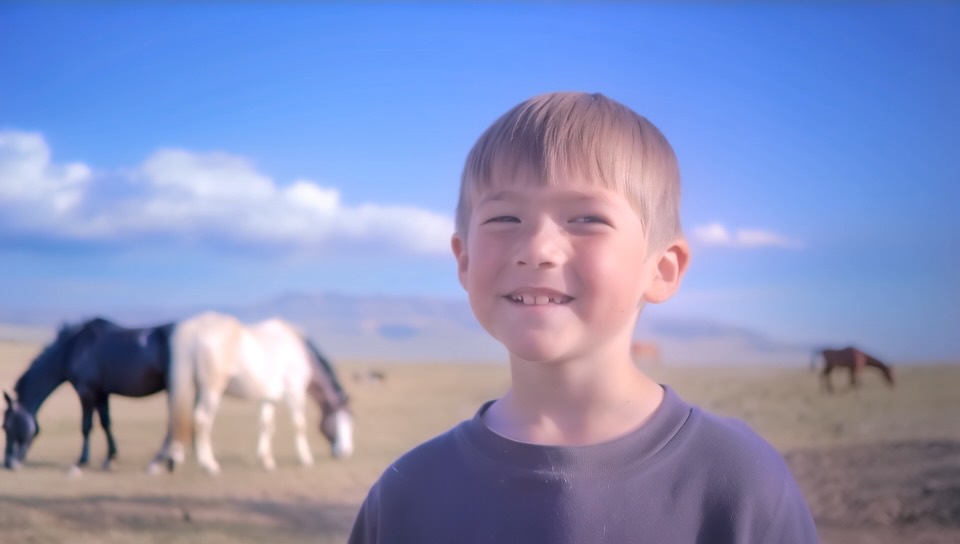} &
            \includegraphics[width=\linewidth]{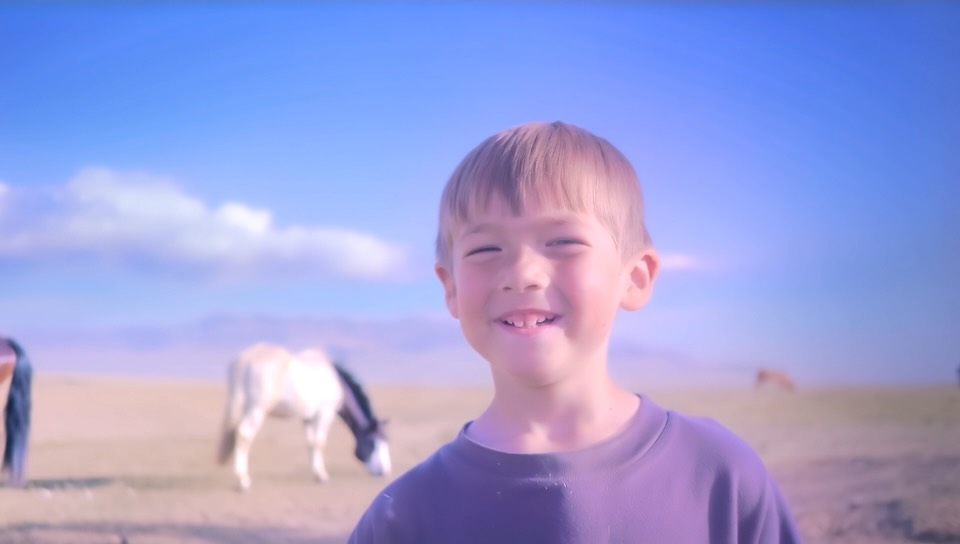} &
            \includegraphics[width=\linewidth]{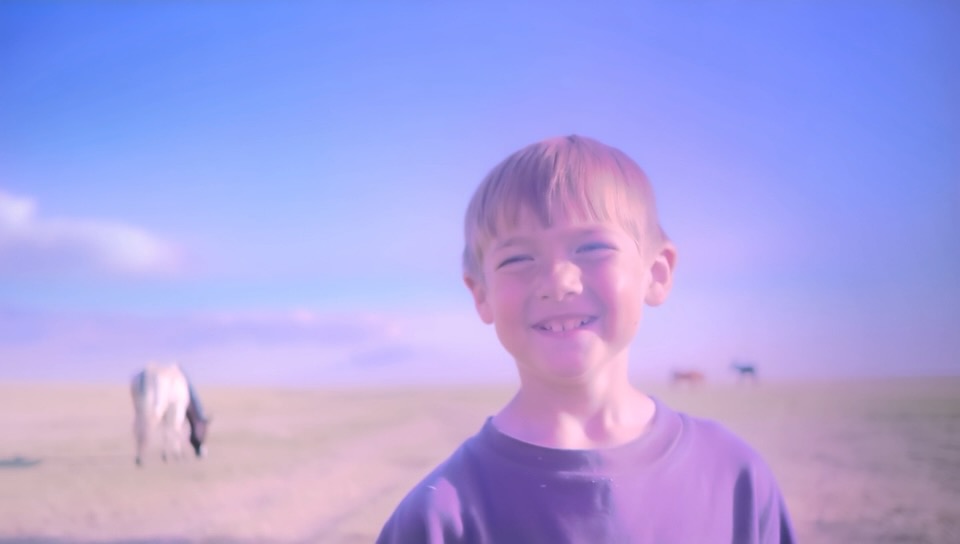} \\
        \end{tabularx}
        \vspace{-8pt}
        \caption{SkyReel-V2 (14B-540P Variant)}
        \label{fig:qualitative-cross-model-magi}
    \end{subfigure}
    \begin{subfigure}[b]{\textwidth}
        \centering
        \setlength\tabcolsep{1pt}
        \begin{tabularx}{\linewidth}{XXXXXXX}
            \includegraphics[width=\linewidth]{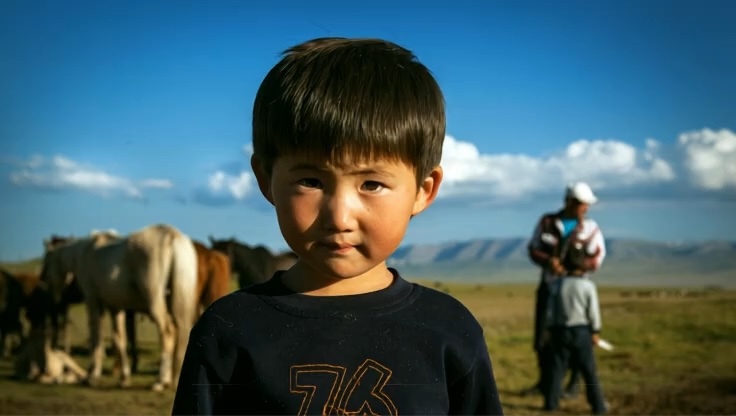} &
            \includegraphics[width=\linewidth]{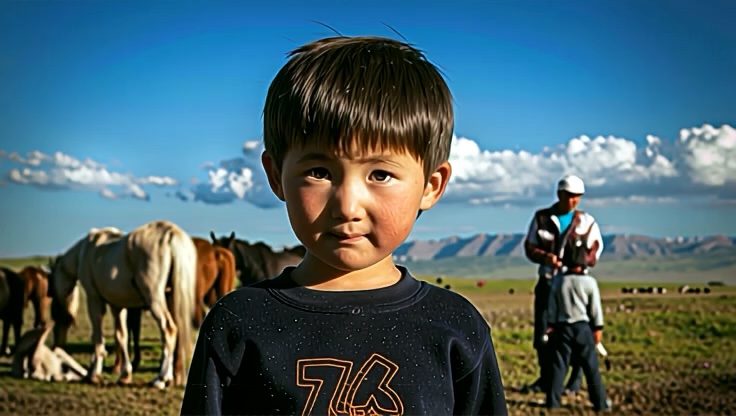} &
            \includegraphics[width=\linewidth]{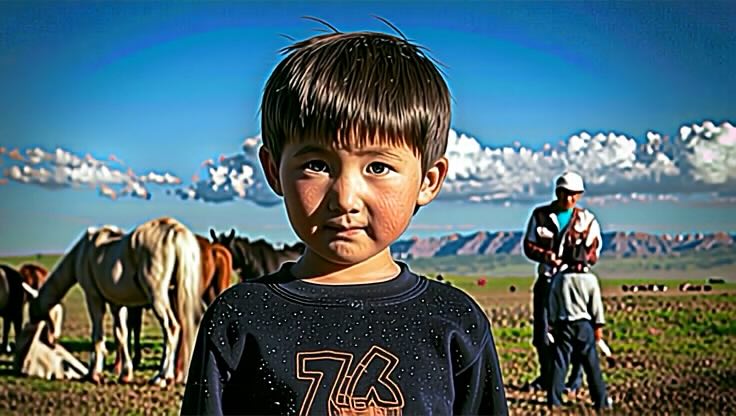} &
            \includegraphics[width=\linewidth]{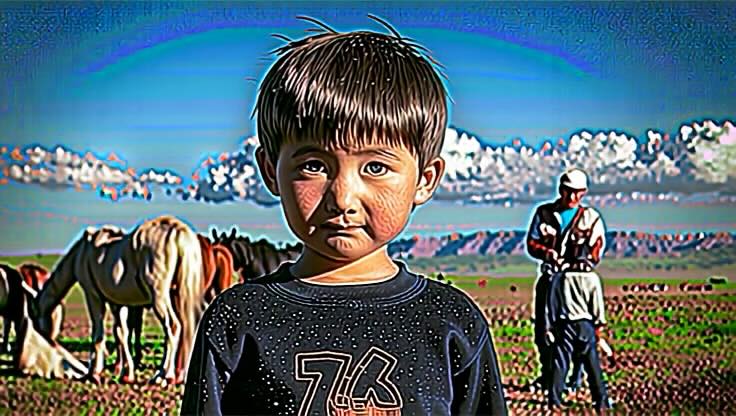} &
            \includegraphics[width=\linewidth]{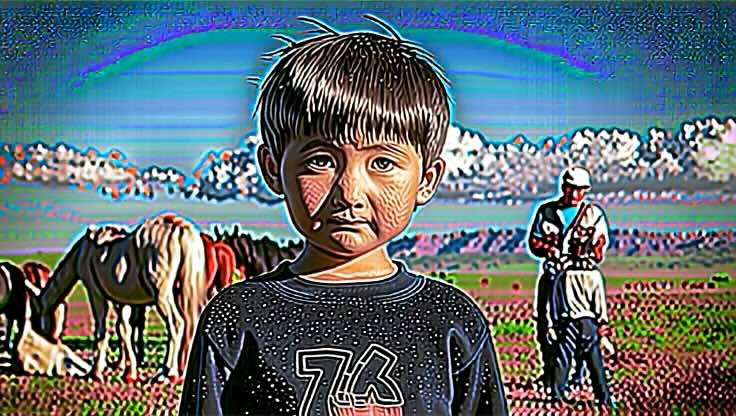} &
            \includegraphics[width=\linewidth]{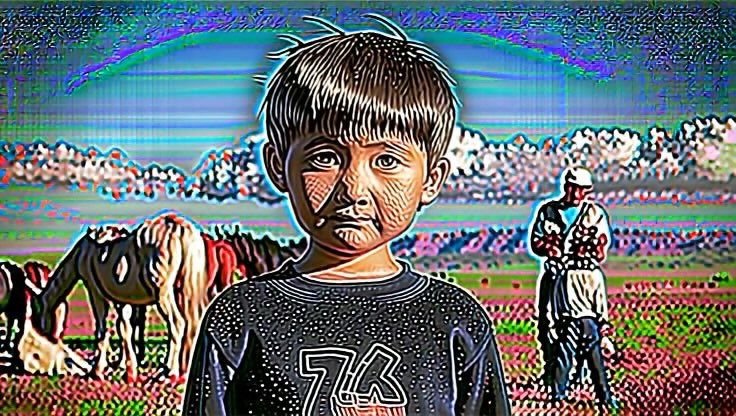} &
            \includegraphics[width=\linewidth]{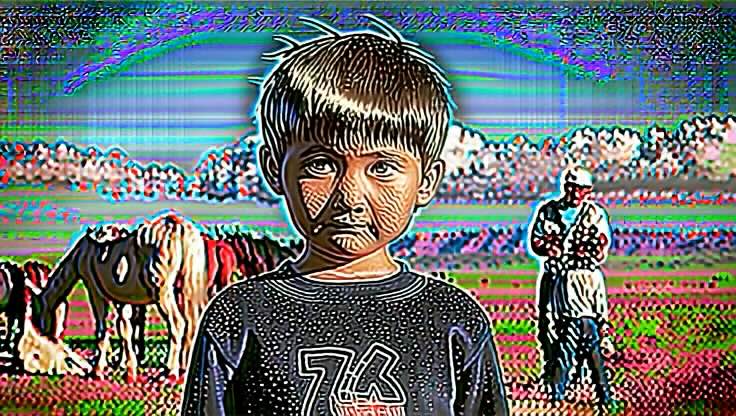} \\
        \end{tabularx}
        \vspace{-8pt}
        \caption{MAGI-1 (24B-Distill Variant)}
        \label{fig:qualitative-cross-model-magi}
    \end{subfigure}
    \begin{subfigure}[b]{\textwidth}
        \centering
        \setlength\tabcolsep{1pt}
        \begin{tabularx}{\linewidth}{XXXXXXX}
            \includegraphics[width=\linewidth]{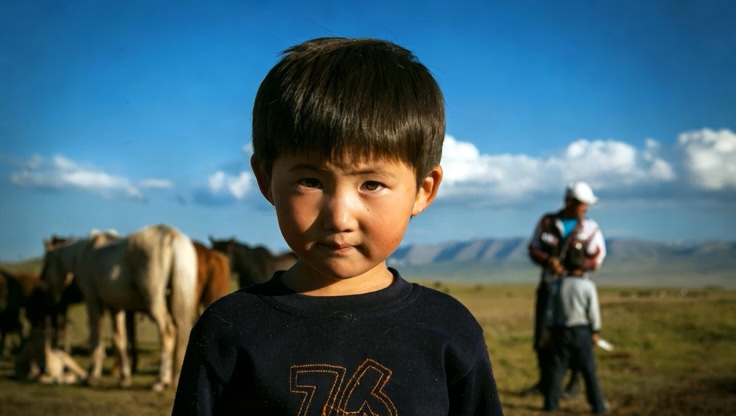} &
            \includegraphics[width=\linewidth]{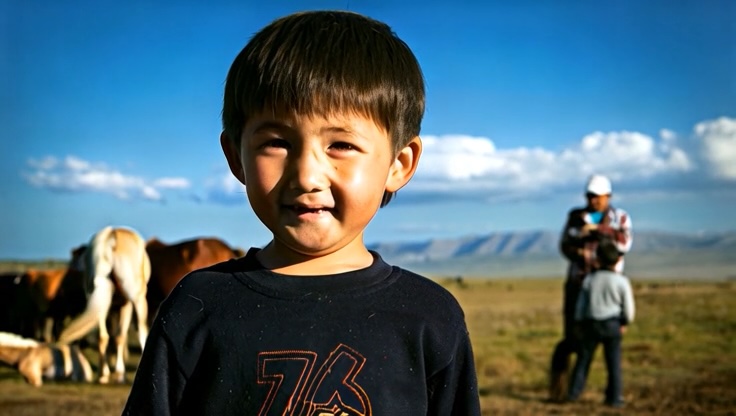} &
            \includegraphics[width=\linewidth]{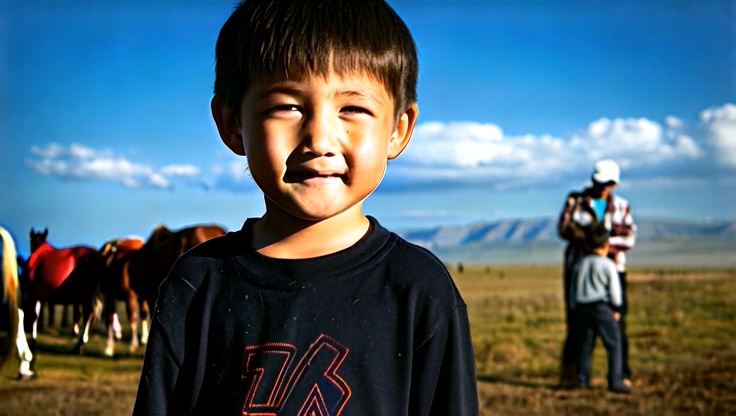} &
            \includegraphics[width=\linewidth]{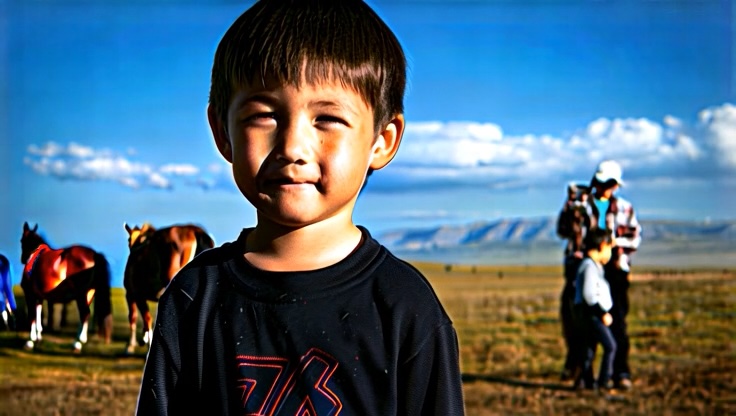} &
            \includegraphics[width=\linewidth]{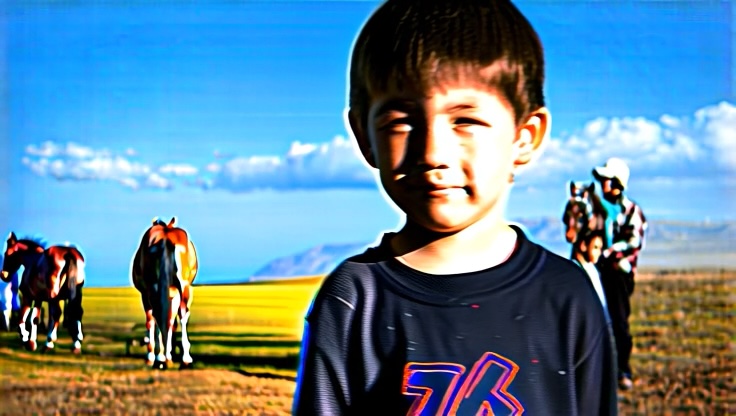} &
            \includegraphics[width=\linewidth]{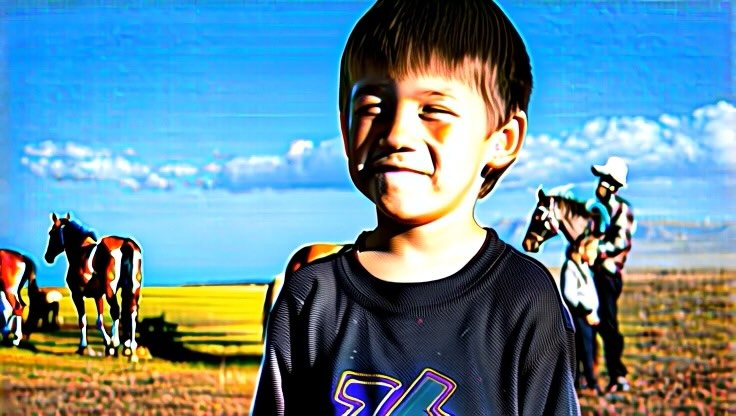} &
            \includegraphics[width=\linewidth]{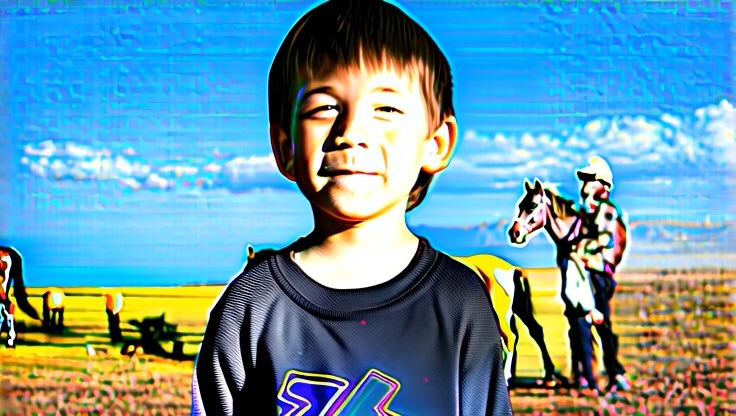} \\
        \end{tabularx}
        \vspace{-8pt}
        \caption{Ours (Diffusion - Extension)}
        \label{fig:qualitative-cross-model-diffusion}
    \end{subfigure}
    \begin{subfigure}[b]{\textwidth}
        \centering
        \setlength\tabcolsep{1pt}
        \begin{tabularx}{\linewidth}{XXXXXXX}
            \includegraphics[width=\linewidth]{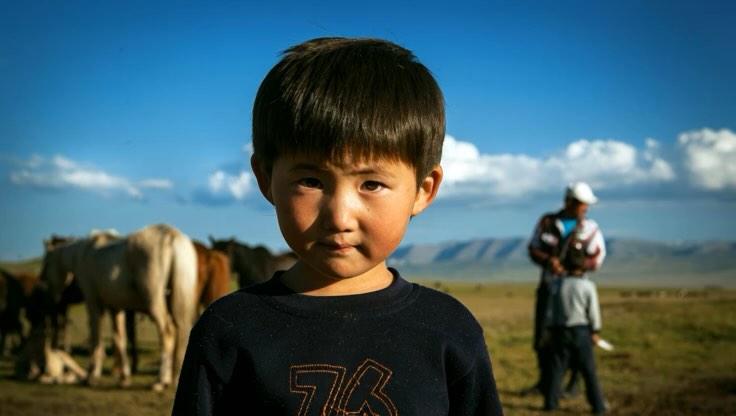} &
            \includegraphics[width=\linewidth]{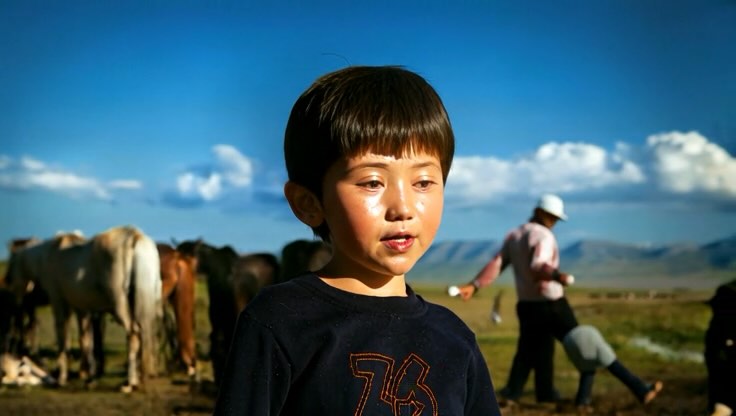} &
            \includegraphics[width=\linewidth]{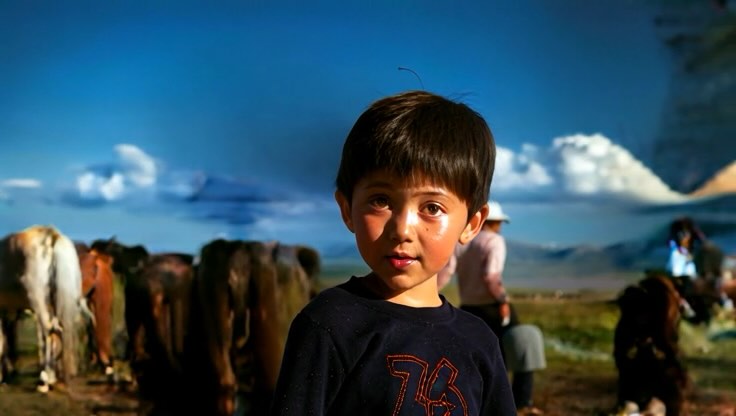} &
            \includegraphics[width=\linewidth]{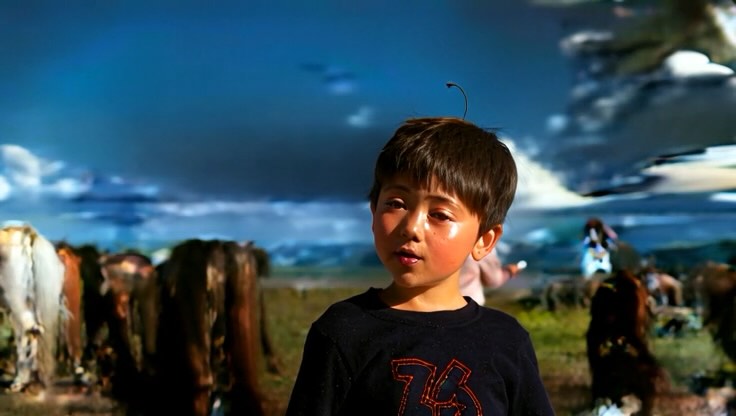} &
            \includegraphics[width=\linewidth]{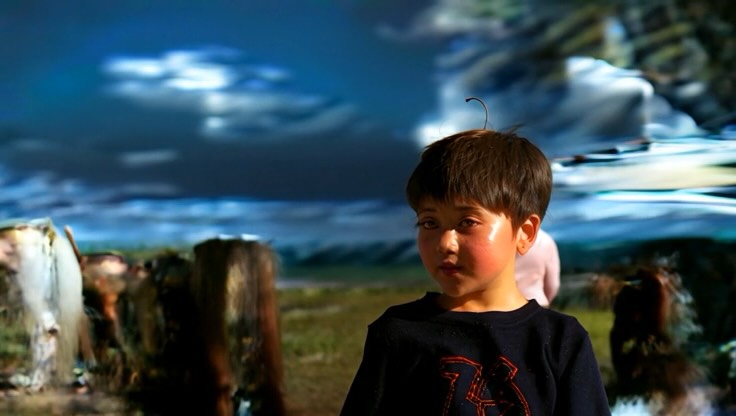} &
            \includegraphics[width=\linewidth]{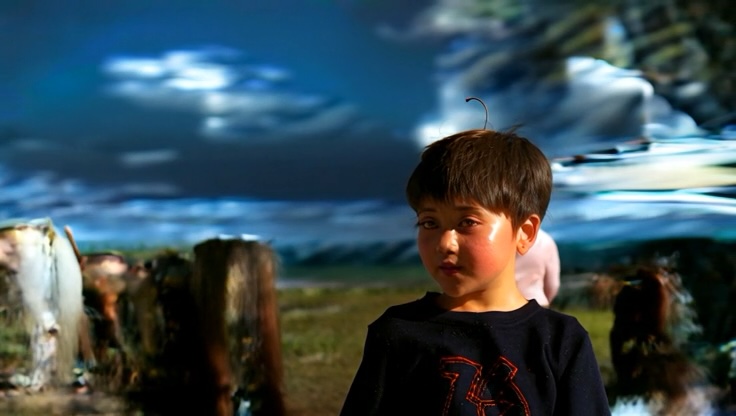} &
            \includegraphics[width=\linewidth]{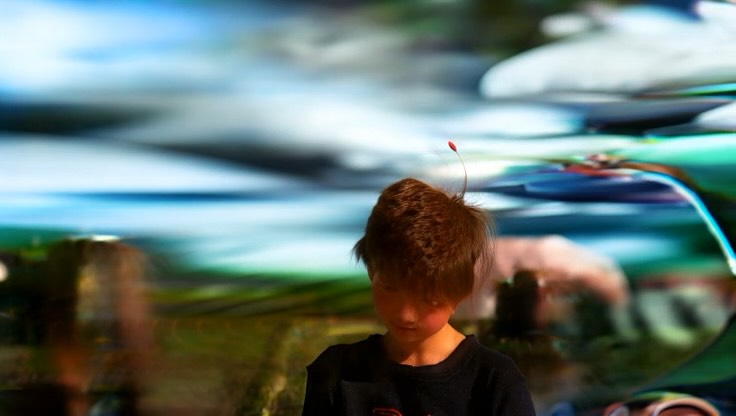} \\
        \end{tabularx}
        \vspace{-8pt}
        \caption{Ours (AAPT - No Long Video Training)}
        \label{fig:qualitative-cross-model-nolong}
    \end{subfigure}
    \begin{subfigure}[b]{\textwidth}
        \centering
        \setlength\tabcolsep{1pt}
        \begin{tabularx}{\linewidth}{XXXXXXX}
            \includegraphics[width=\linewidth]{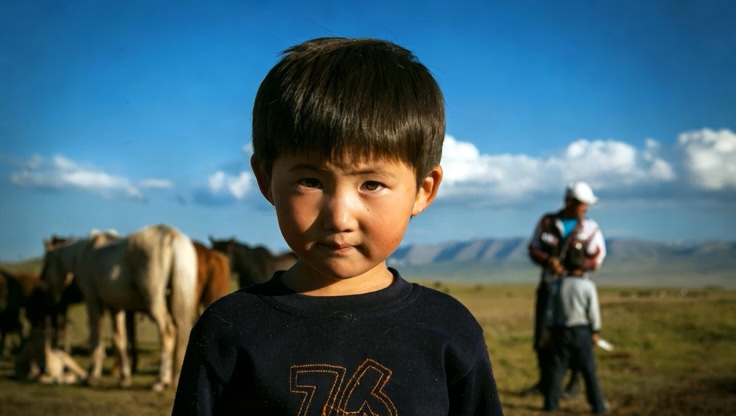} &
            \includegraphics[width=\linewidth]{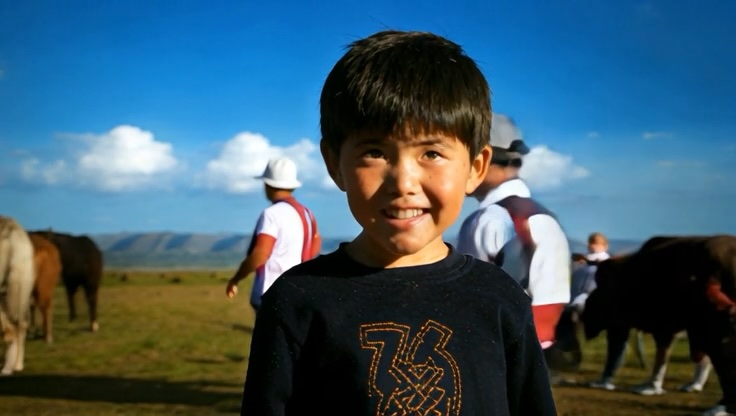} &
            \includegraphics[width=\linewidth]{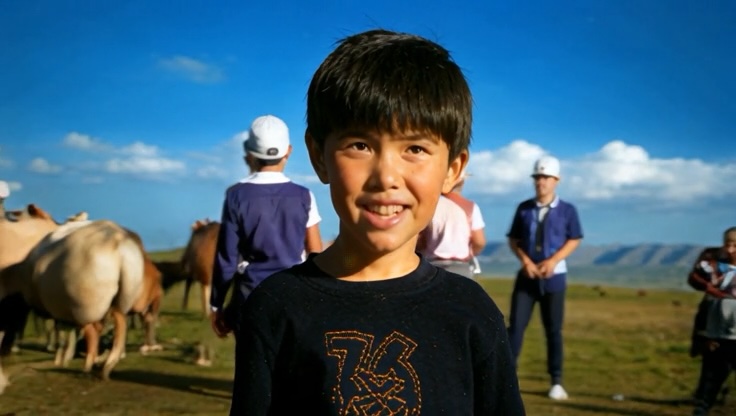} &
            \includegraphics[width=\linewidth]{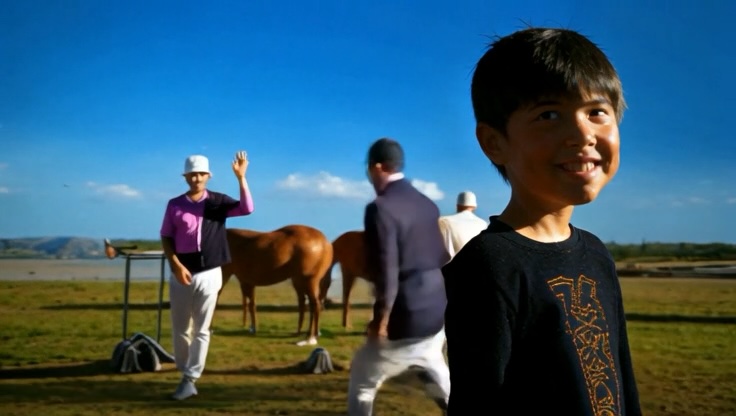} &
            \includegraphics[width=\linewidth]{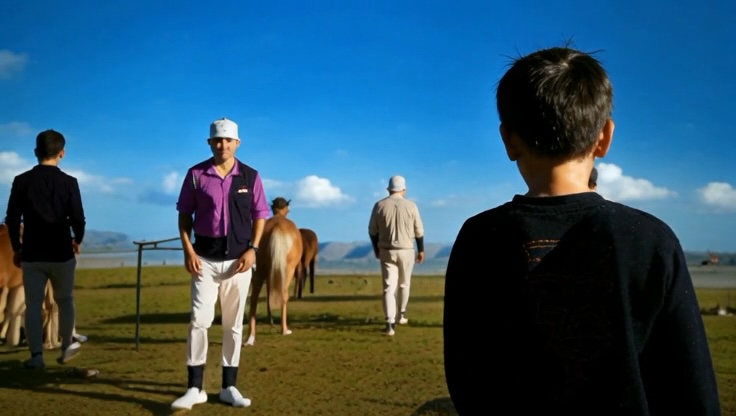} &
            \includegraphics[width=\linewidth]{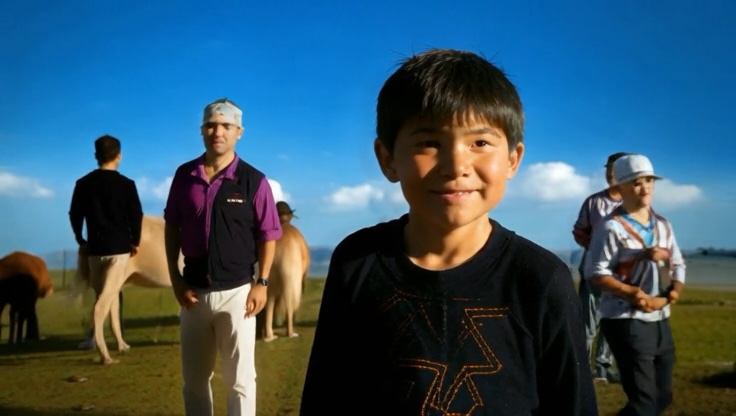} &
            \includegraphics[width=\linewidth]{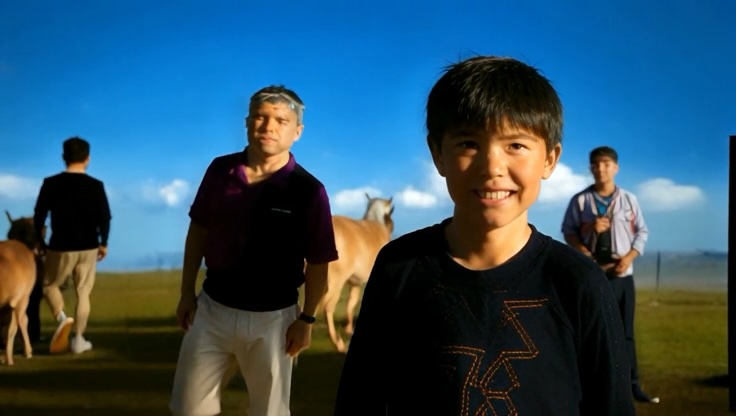} \\
        \end{tabularx}
        \vspace{-8pt}
        \caption{\textbf{Ours (AAPT)}}
        \label{fig:qualitative-cross-model-final}
    \end{subfigure}
    \captionsetup{justification=centering,singlelinecheck=true}
    \vspace{-10pt}
    \caption{Qualitative comparison on one-minute, 1440-frame, VBench-I2V generation.}
    \label{fig:qualitative-cross-model}
    \vspace{-5pt}
\end{figure*}
\begin{figure*}[t]
    \centering
    \small
    \begin{tabularx}{\linewidth}{|X@{\hspace{8pt}}|X|X|X|X|X|X|}
        0s (Input) & 10s & 20s & 30s & 40s & 50s & 60s
    \end{tabularx}
    \setlength\tabcolsep{1pt}
    \begin{tabularx}{\linewidth}{XXXXXXX}

        \includegraphics[width=\linewidth]{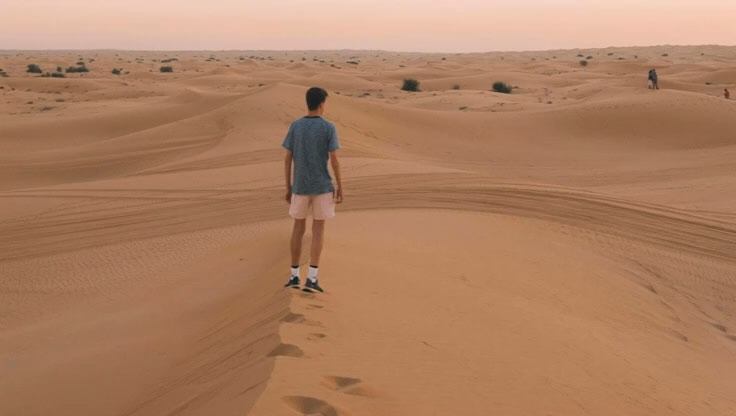} &
        \includegraphics[width=\linewidth]{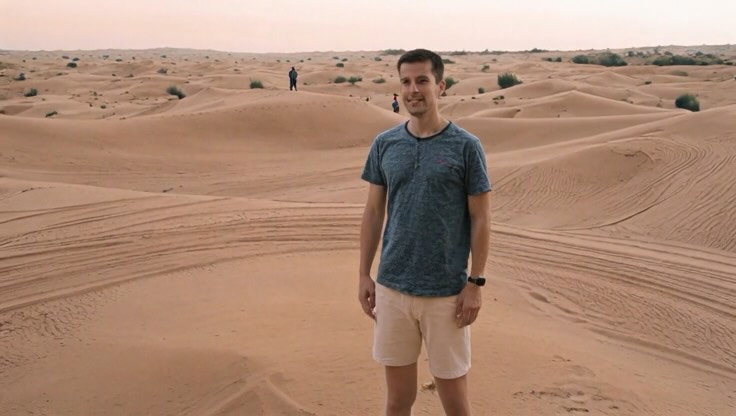} &
        \includegraphics[width=\linewidth]{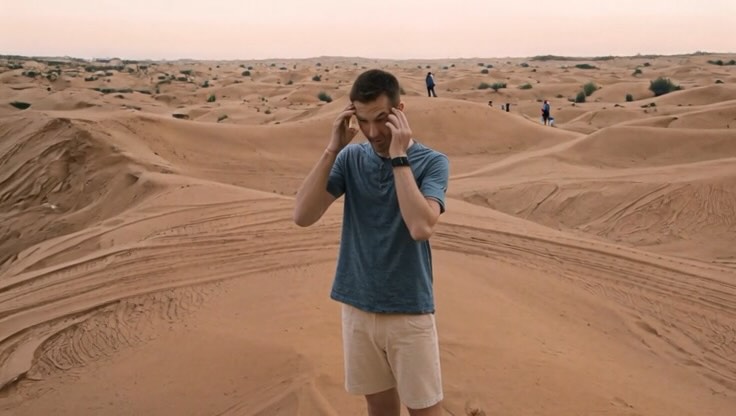} &
        \includegraphics[width=\linewidth]{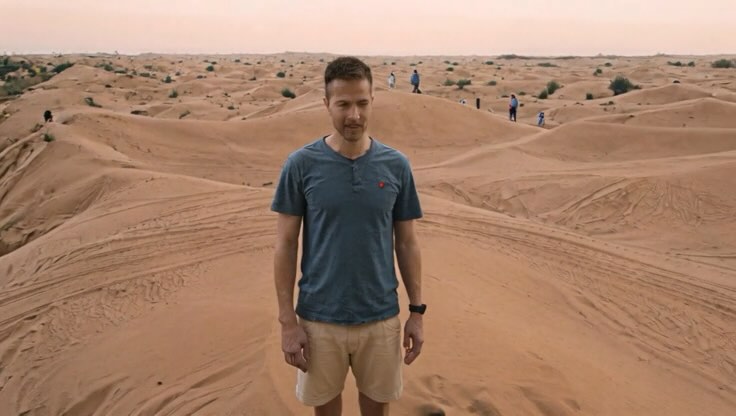} &
        \includegraphics[width=\linewidth]{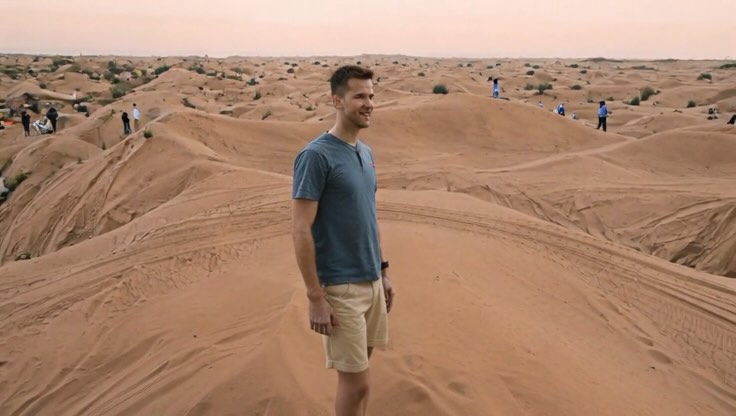} &
        \includegraphics[width=\linewidth]{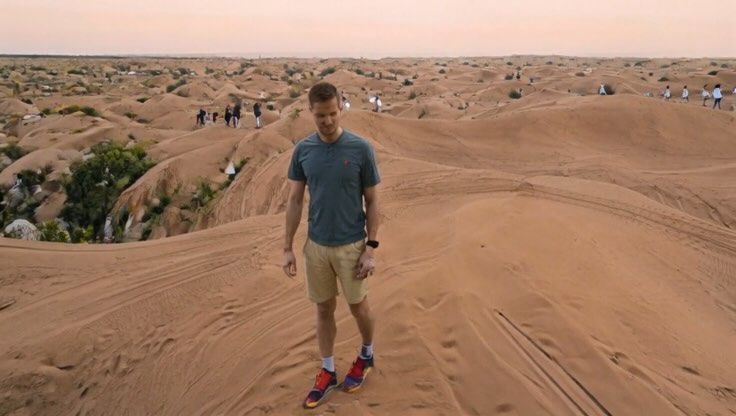} &
        \includegraphics[width=\linewidth]{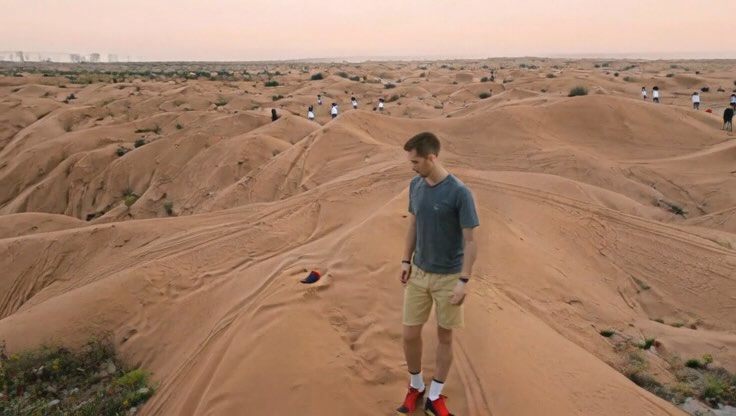} \\

        \includegraphics[width=\linewidth]{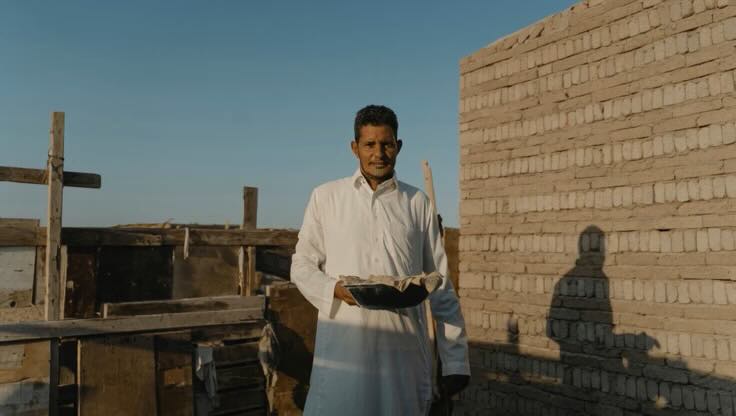} &
        \includegraphics[width=\linewidth]{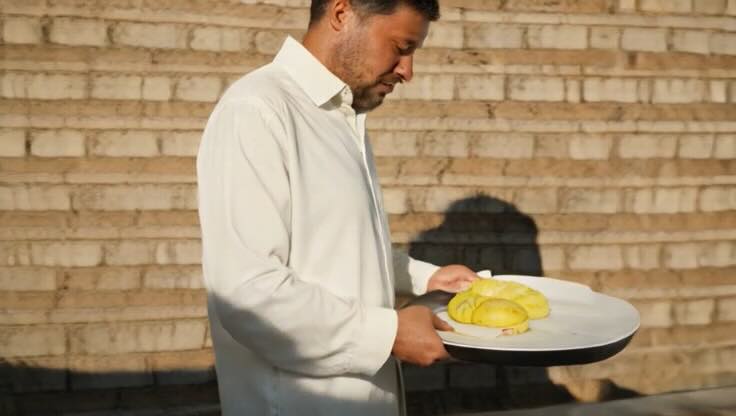} &
        \includegraphics[width=\linewidth]{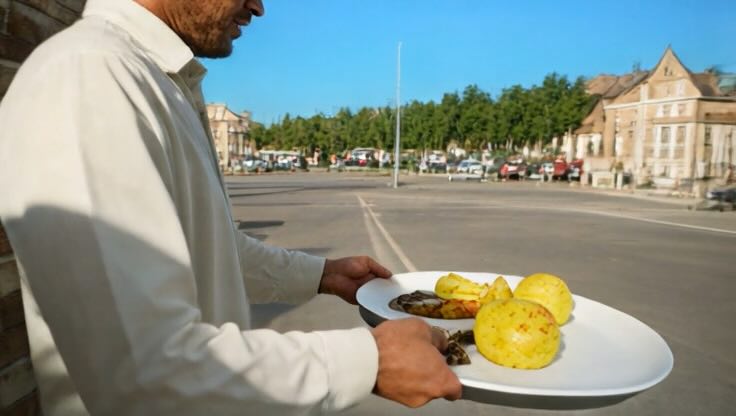} &
        \includegraphics[width=\linewidth]{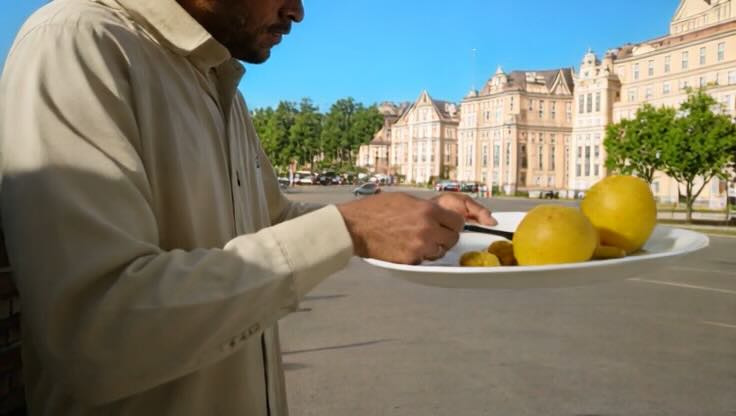} &
        \includegraphics[width=\linewidth]{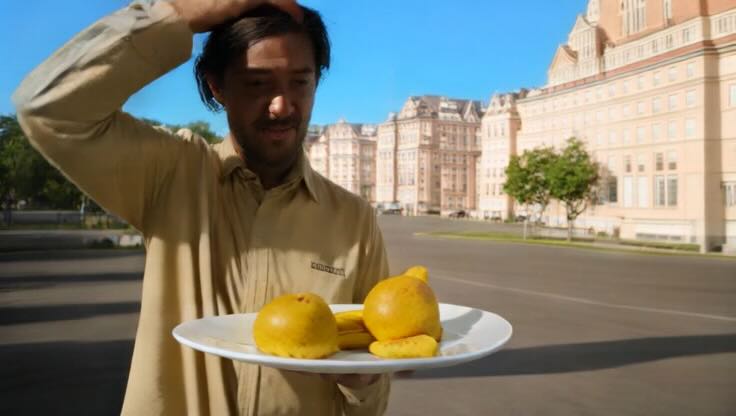} &
        \includegraphics[width=\linewidth]{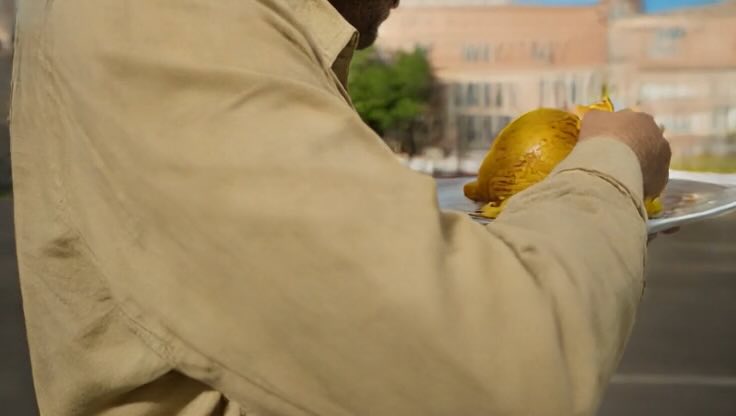} &
        \includegraphics[width=\linewidth]{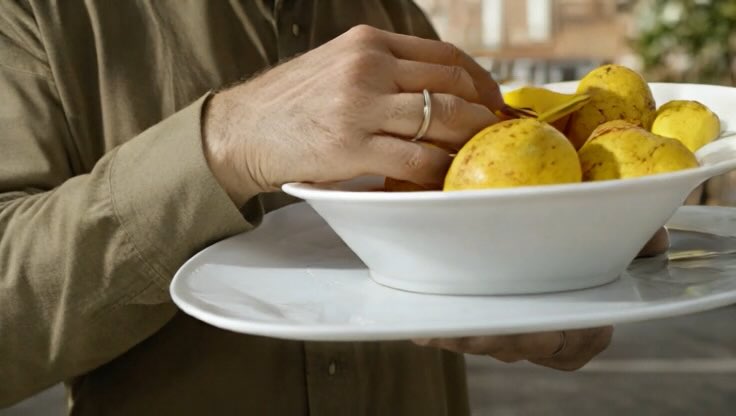} \\

        \includegraphics[width=\linewidth]{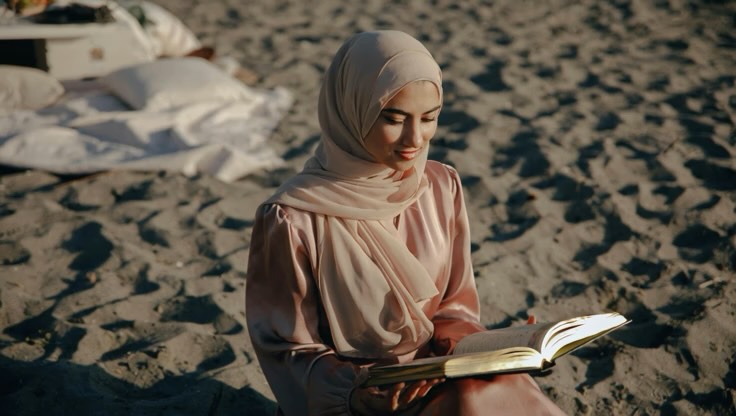} &
        \includegraphics[width=\linewidth]{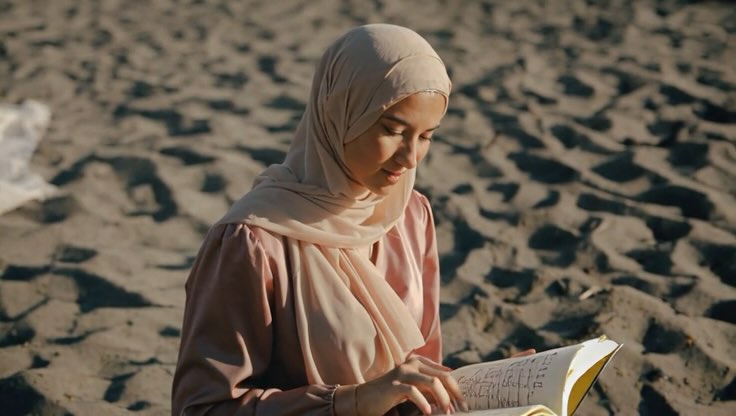} &
        \includegraphics[width=\linewidth]{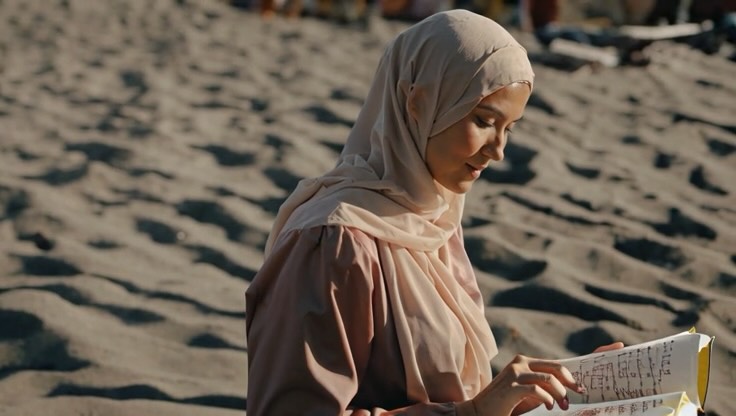} &
        \includegraphics[width=\linewidth]{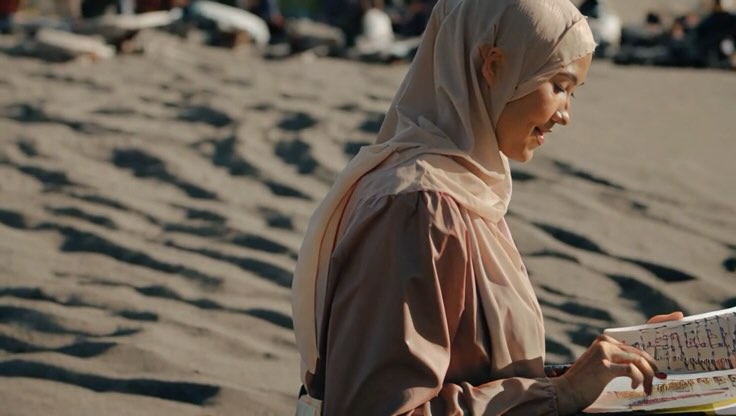} &
        \includegraphics[width=\linewidth]{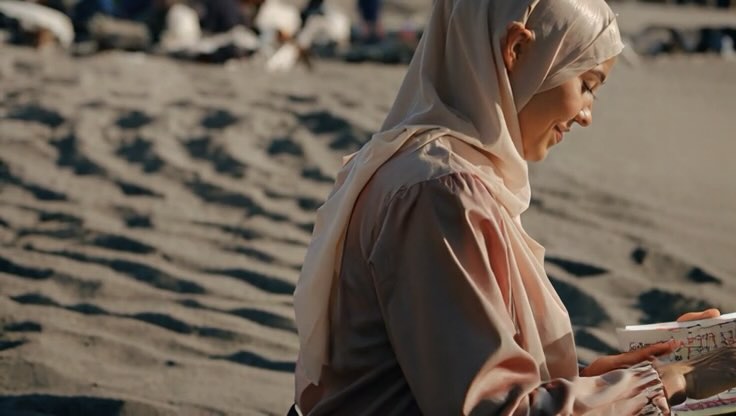} &
        \includegraphics[width=\linewidth]{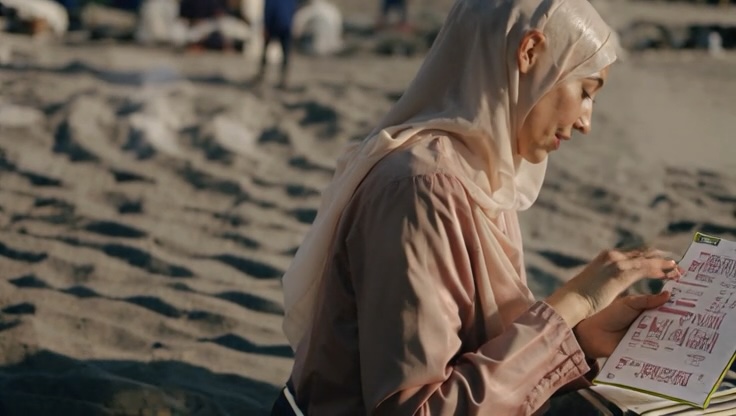} &
        \includegraphics[width=\linewidth]{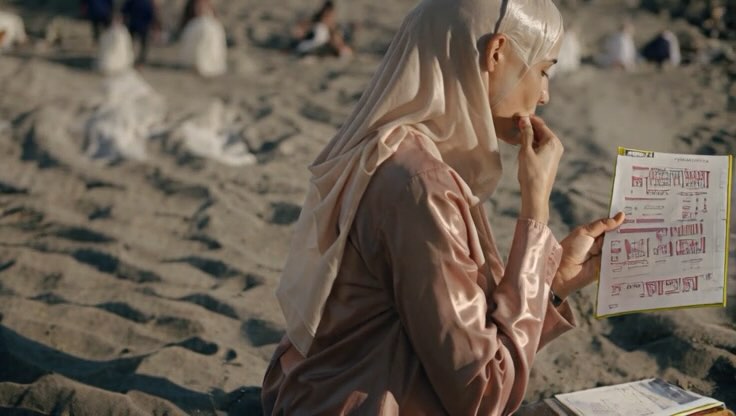} \\
        
        \includegraphics[width=\linewidth]{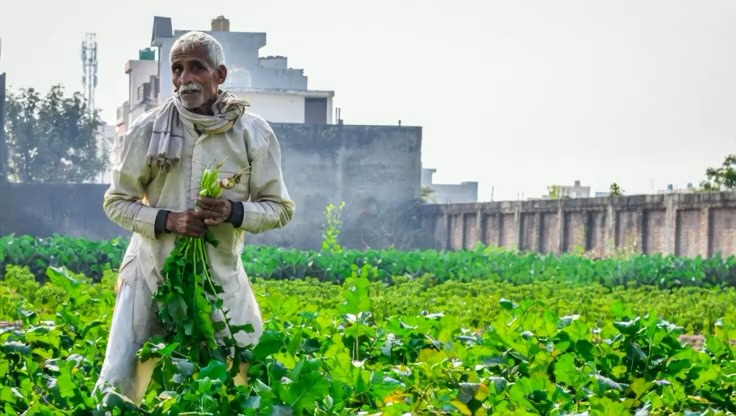} &
        \includegraphics[width=\linewidth]{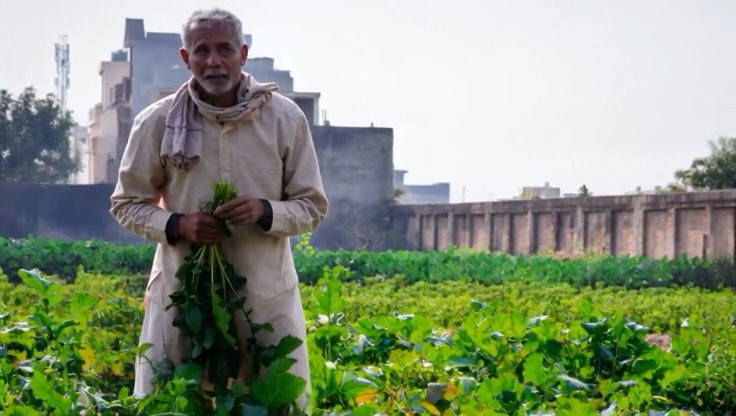} &
        \includegraphics[width=\linewidth]{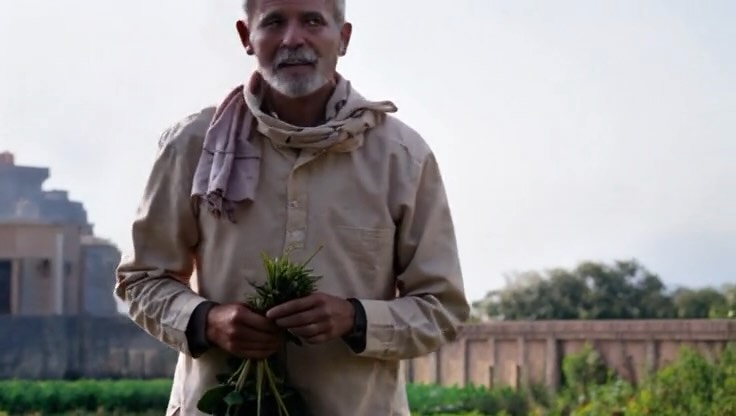} &
        \includegraphics[width=\linewidth]{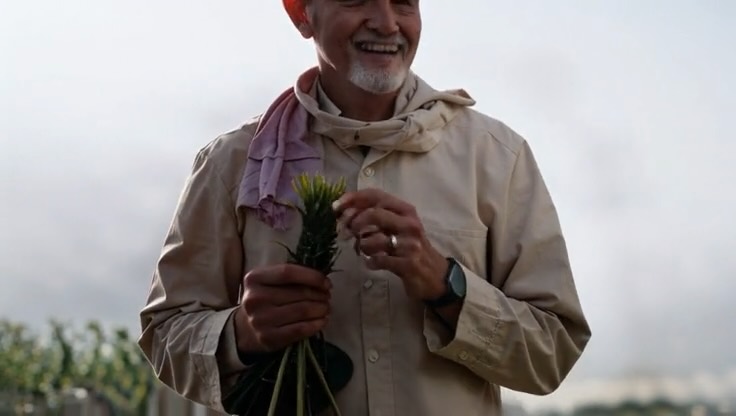} &
        \includegraphics[width=\linewidth]{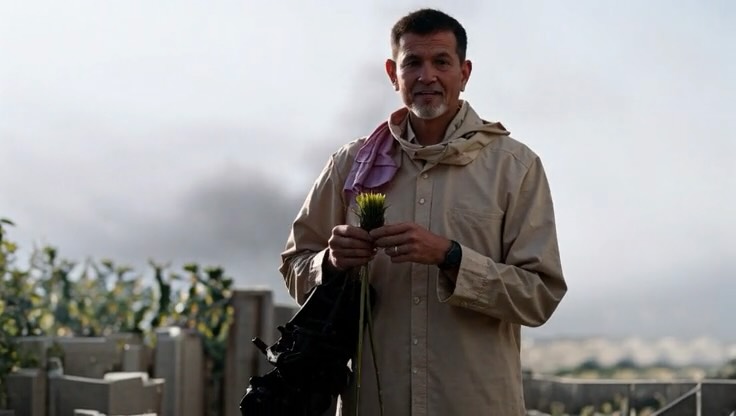} &
        \includegraphics[width=\linewidth]{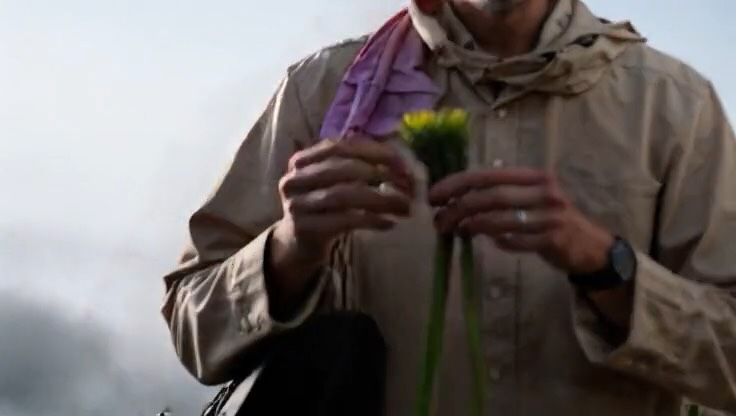} &
        \includegraphics[width=\linewidth]{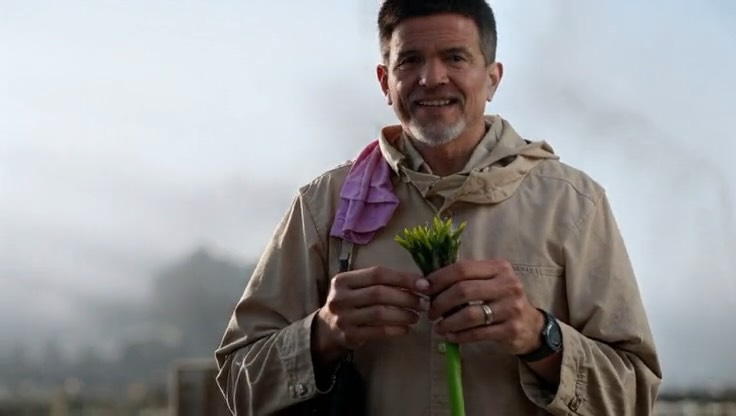} \\

        \includegraphics[width=\linewidth]{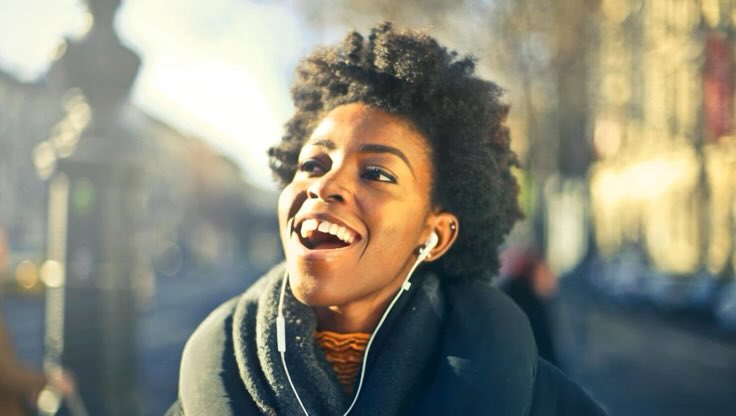} &
        \includegraphics[width=\linewidth]{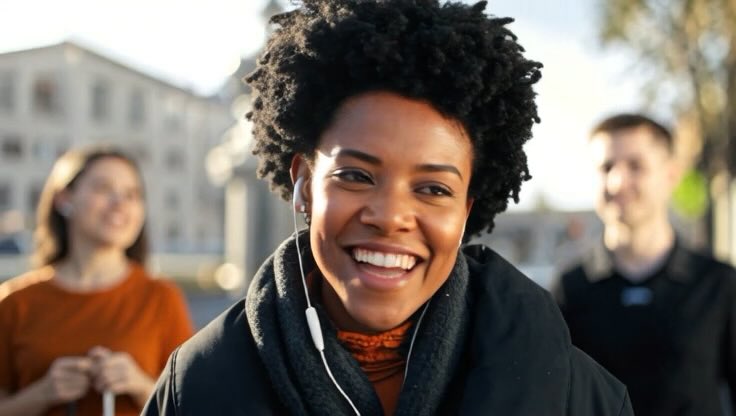} &
        \includegraphics[width=\linewidth]{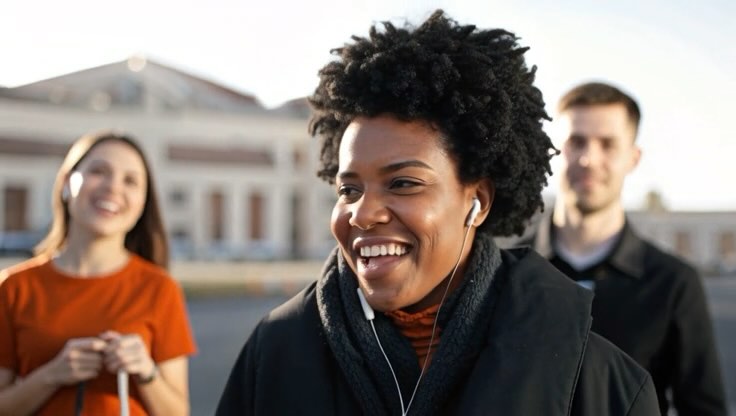} &
        \includegraphics[width=\linewidth]{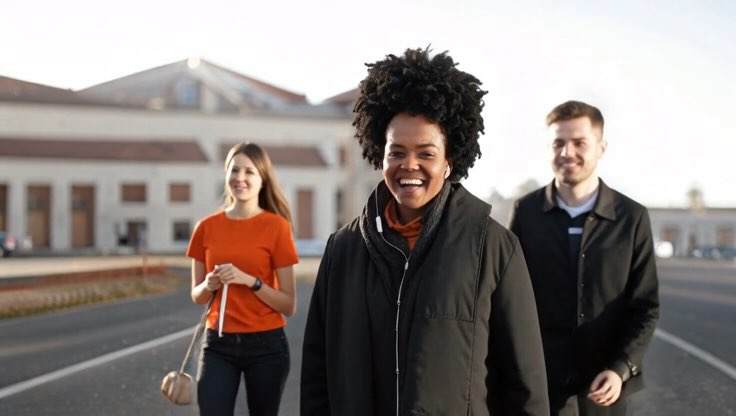} &
        \includegraphics[width=\linewidth]{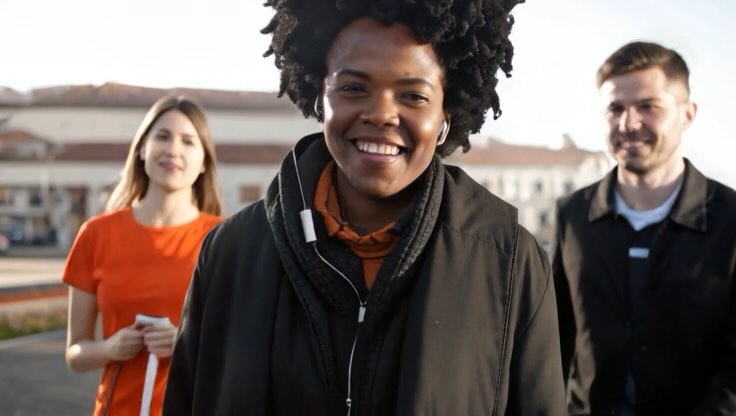} &
        \includegraphics[width=\linewidth]{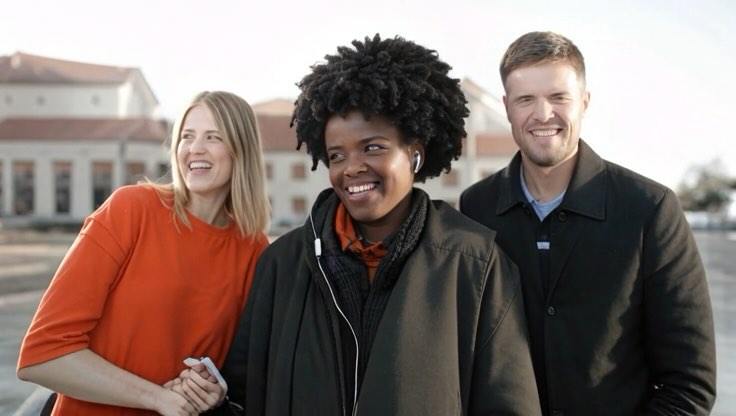} &
        \includegraphics[width=\linewidth]{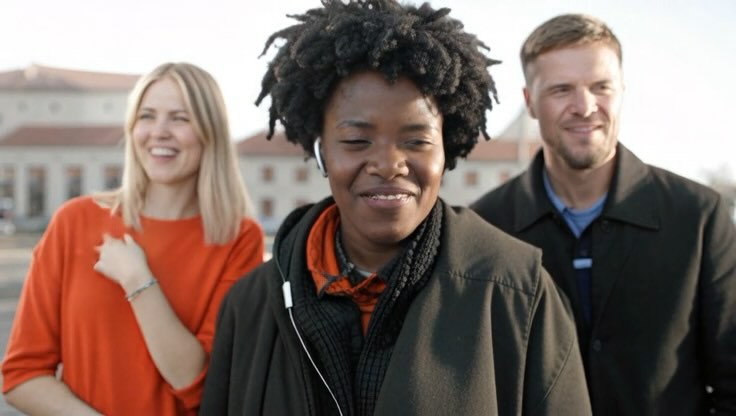} \\

        \includegraphics[width=\linewidth]{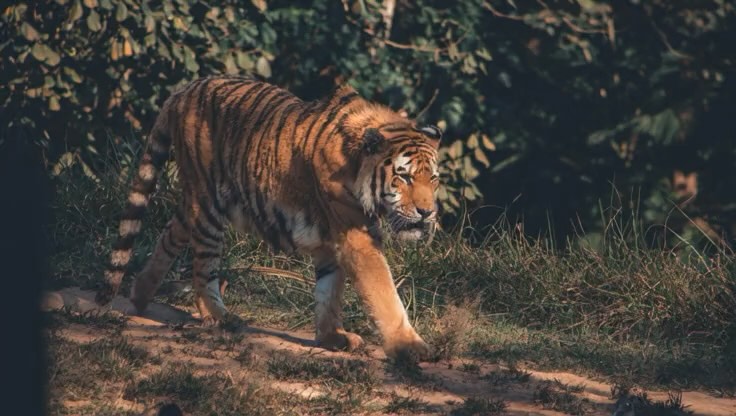} &
        \includegraphics[width=\linewidth]{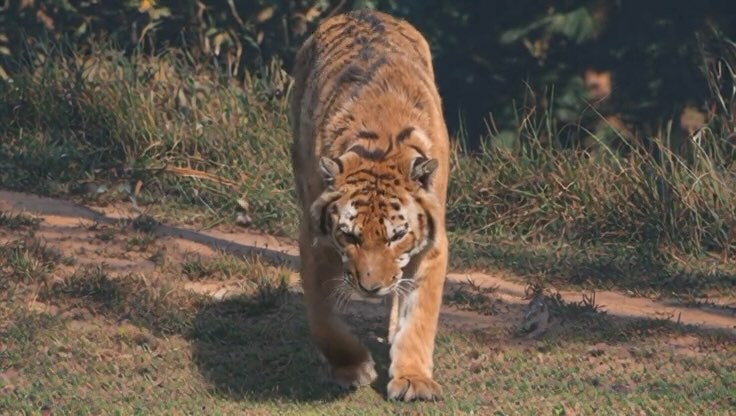} &
        \includegraphics[width=\linewidth]{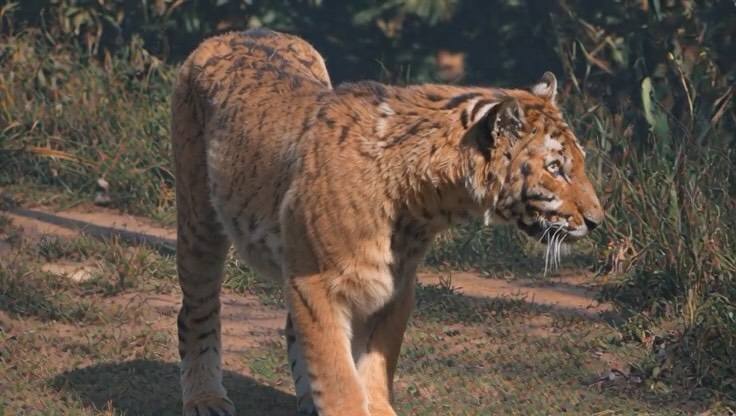} &
        \includegraphics[width=\linewidth]{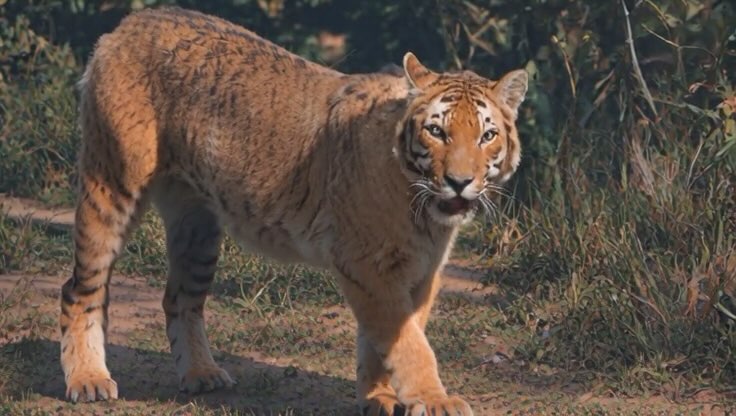} &
        \includegraphics[width=\linewidth]{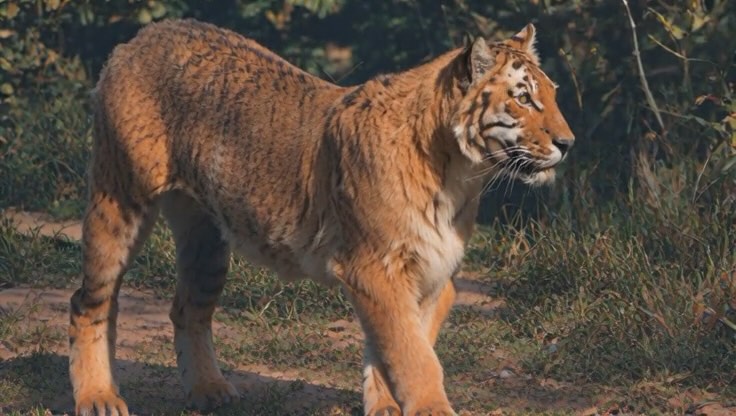} &
        \includegraphics[width=\linewidth]{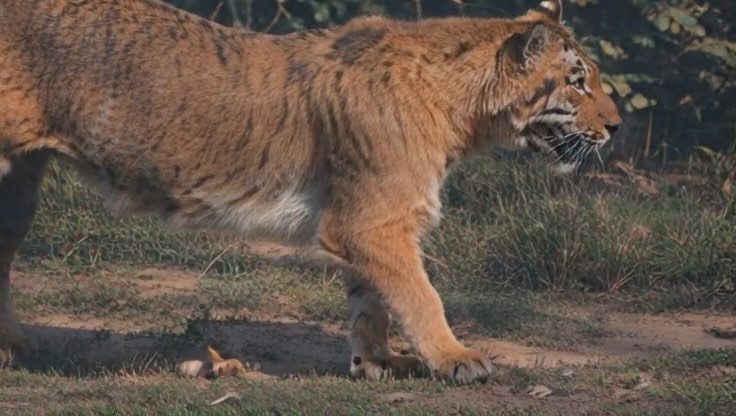} &
        \includegraphics[width=\linewidth]{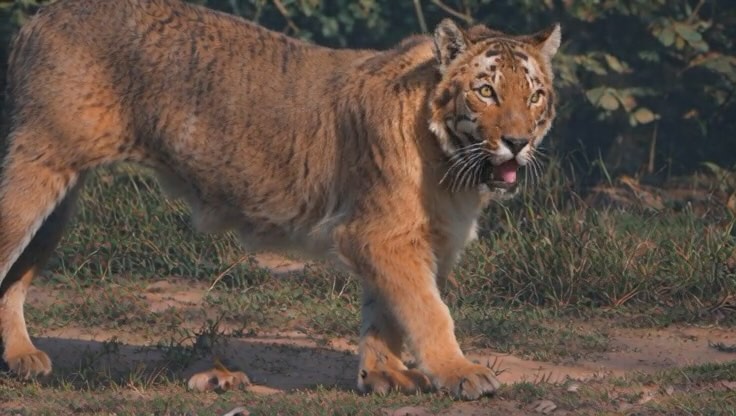} \\

        \includegraphics[width=\linewidth]{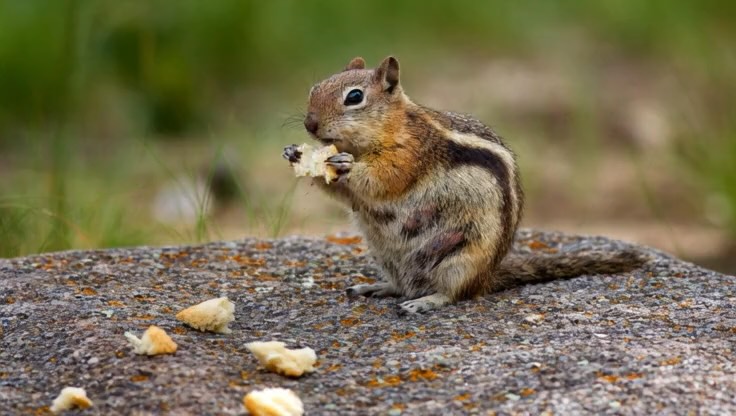} &
        \includegraphics[width=\linewidth]{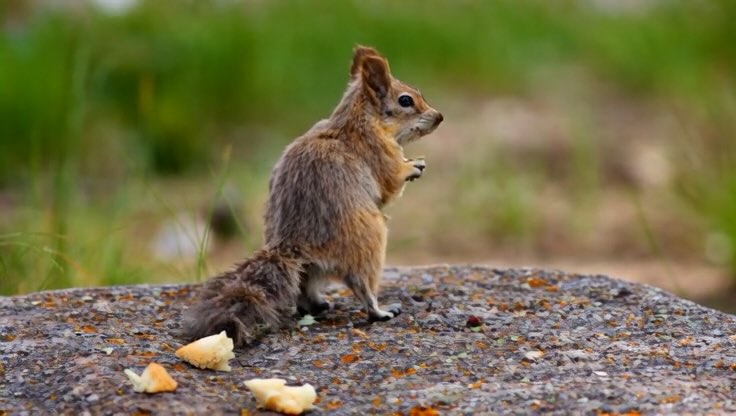} &
        \includegraphics[width=\linewidth]{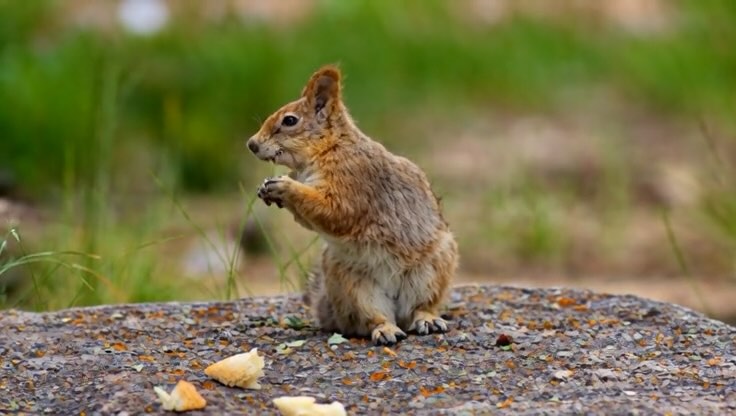} &
        \includegraphics[width=\linewidth]{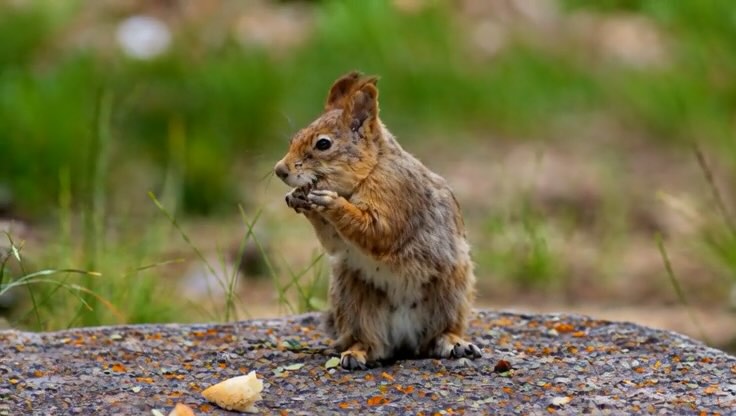} &
        \includegraphics[width=\linewidth]{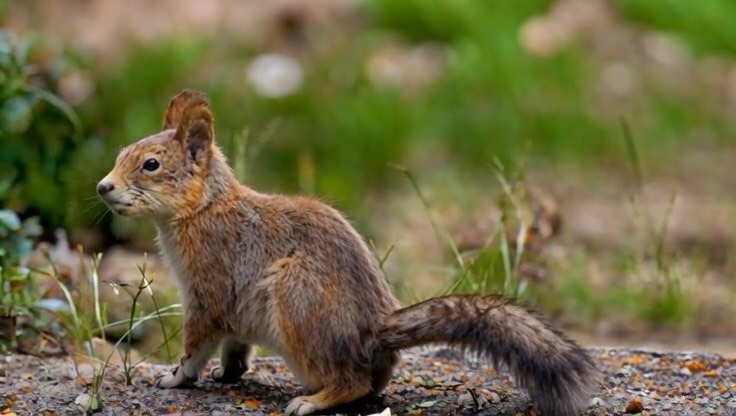} &
        \includegraphics[width=\linewidth]{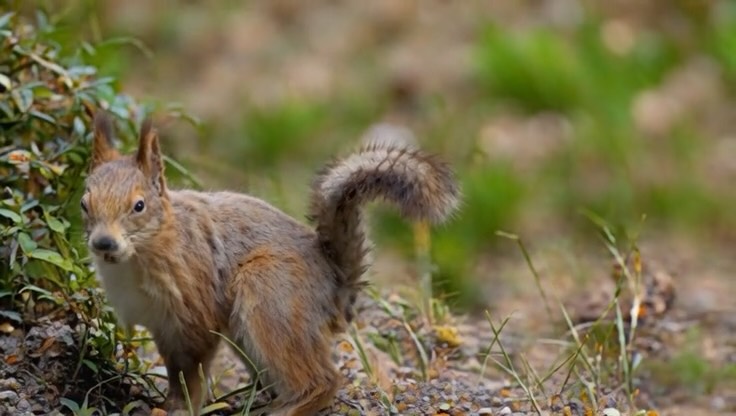} &
        \includegraphics[width=\linewidth]{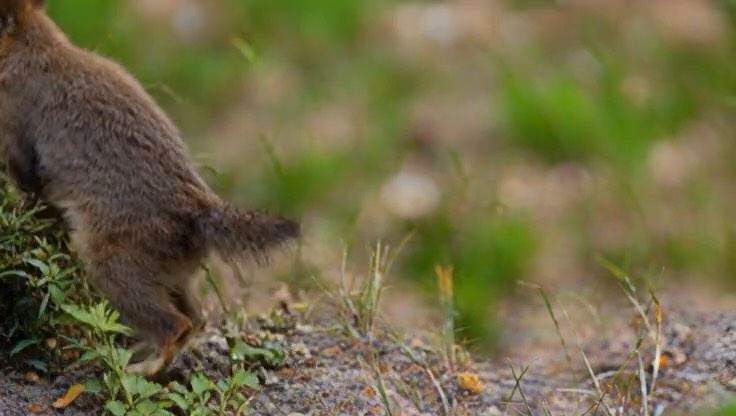} \\

        \includegraphics[width=\linewidth]{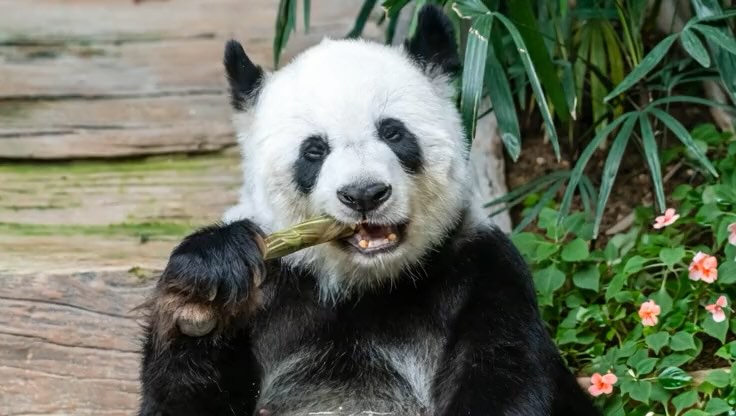} &
        \includegraphics[width=\linewidth]{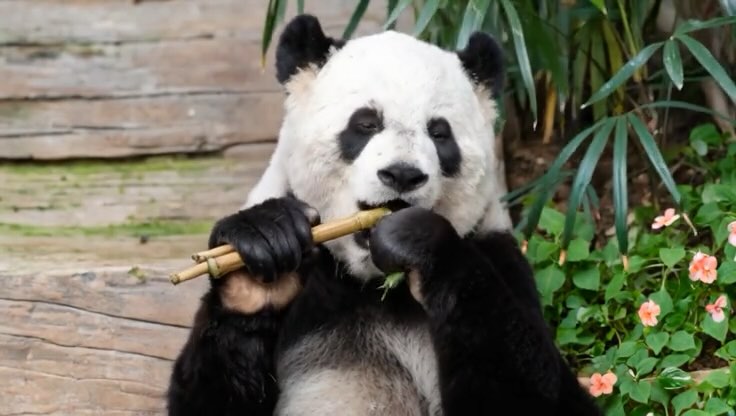} &
        \includegraphics[width=\linewidth]{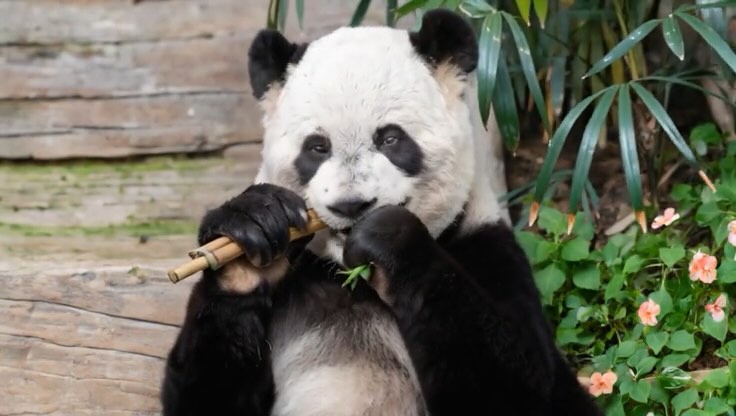} &
        \includegraphics[width=\linewidth]{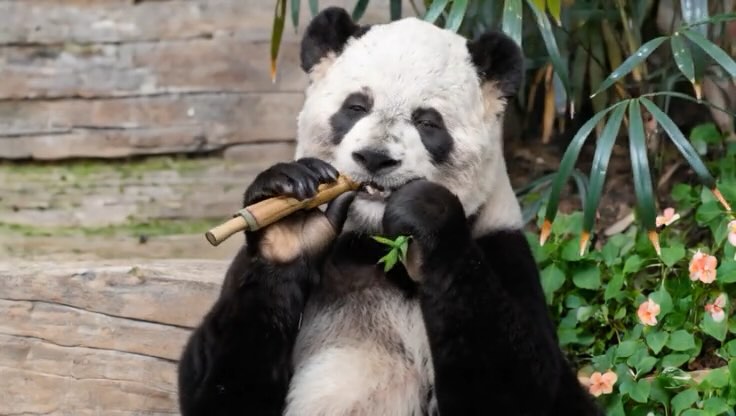} &
        \includegraphics[width=\linewidth]{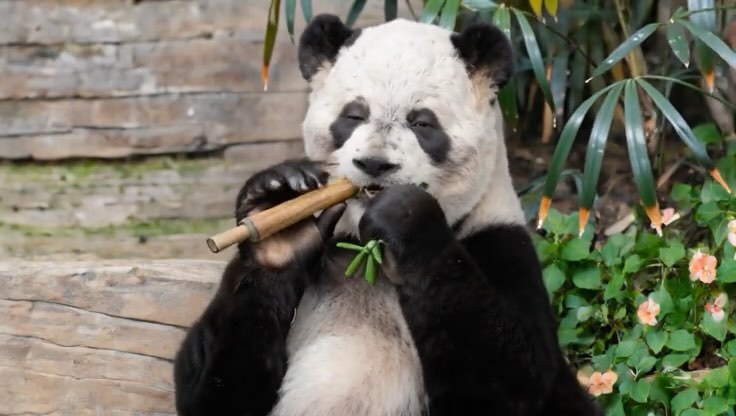} &
        \includegraphics[width=\linewidth]{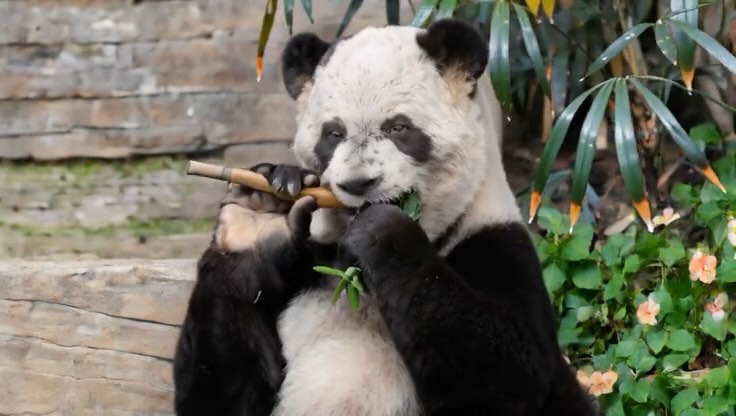} &
        \includegraphics[width=\linewidth]{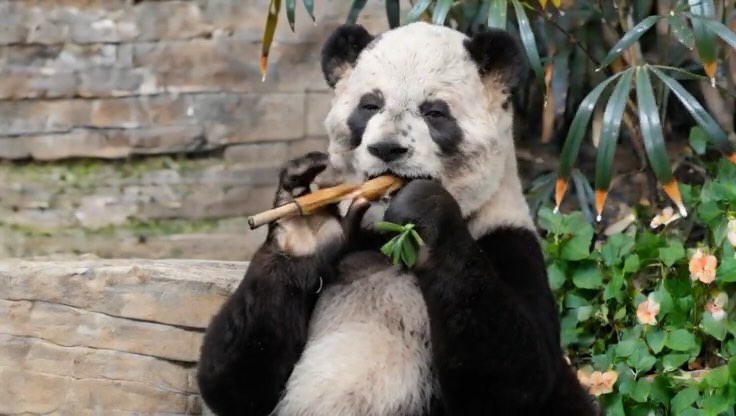} \\

    \end{tabularx}
    \caption{More results of our AAPT model for one-minute, 1440-frame, VBench-I2V generation.}
    \label{fig:qualitative-i2v}
    \vspace{-15pt}
\end{figure*}

\paragraph{Main Results} \Cref{fig:qualitative-cross-model} qualitatively compares our method on one-minute (1440-frame) video generation against SkyReel-V2, MAGI-1, and our diffusion baseline. All three of them exhibit strong error accumulation after 20 to 30 seconds. For our diffusion baseline, we experiment using a lower CFG scale or using rescale~\cite{lin2024common} but it does not mitigate the exposure problem and can further cause more structural deformation, so we keep it at CFG 10. We also show that our AAPT model trained on only a 10-second duration cannot generalize to long videos in \cref{fig:qualitative-cross-model-nolong}. Long video training is critical, as shown in \cref{fig:qualitative-cross-model-final}. \Cref{fig:qualitative-i2v} shows more results of our model across subjects and scenes.

\Cref{tab:eval-i2v} shows that our method achieves competitive performance compared to the state-of-the-art methods on the quantitative metrics. For 120-frame I2V generation, AAPT improves frame quality score and image conditioning scores compared to the diffusion baseline and is the best across all compared methods. The frame quality improvement concurs with the findings in APT~\cite{lin2025diffusion} that adversarial training can improve visual quality. AAPT has resulted in a slight decline in temporal quality score compared to the diffusion baseline, but is still above Wan and closely follows Hunyuan. We note that CausVid has an exceptionally high temporal quality score, likely because it was trained on 12fps data, which usually results in a higher dynamic degree than other 24fps models, and the dynamic degree score is the main differentiator for the overall temporal quality. For 1440-frame I2V generation, AAPT achieves the best quality scores across the comparison and has improved conditioning scores compared to the diffusion baseline. We note that SkyReel-V2 and MAGI-1 have a higher image-conditioning score compared to our AAPT and diffusion baseline which is because most of the videos by MAGI-1 are stationary. This is reflected in its much lower dynamic degree score and the qualitative visualization in \cref{fig:qualitative-cross-model}.

\begin{minipage}[t]{\linewidth}
    \centering
    \captionof{table}{Quantitative comparisons on VBench-I2V~\cite{huang2024vbench}. * denotes metrics that need special interpretation as discussed in the main text. The 6 quality metrics are aggregated as temporal quality and frame quality according to VBench-Competition. The best metrics are highlighted in bold.}
    \small
    \setlength\tabcolsep{2pt}
    \begin{tabularx}{\linewidth}{cX|rc|cccccc|rr}
        \toprule
        & & \multicolumn{8}{c|}{Quality} & \multicolumn{2}{c}{Condition} \\
        Frames & Method & \tiny\textbf{\makecell{Temporal\\Quality}} & \tiny\textbf{\makecell{Frame\\Quality}}  & \tiny\makecell{Subject\\Consistency} & \tiny\makecell{Background\\Consistency} & \tiny\makecell{Motion\\Smoothness} & \tiny\makecell{Dynamic\\Degree} & \tiny\makecell{Aesthetic\\Quality} & \tiny\makecell{Imaging\\Quality} & \tiny\textbf{\makecell{I2V\\Subject}} & \tiny\textbf{\makecell{I2V\\Background}} \\
        \midrule
        \multirow{5}{*}{120} & CausVid~\cite{yin2024slow} & \textbf{*92.00} & 65.00 & \multicolumn{8}{c}{Not Reported} \\
         & Wan 2.1~\cite{wang2025wan} & 87.95 & 66.58 & 93.85 & 96.59 & 97.82 & 39.11 & 63.56 & 69.59 & 96.82 & 98.57 \\
         & Hunyuan~\cite{kong2024hunyuanvideo} & 89.80 & 64.18 & 93.06 & 95.29 & 98.53 & 54.80 & 60.58 & 67.78 & 97.71 & 97.97 \\
         & Ours (Diffusion) & 90.40 & 66.08 & 94.58 & 96.76 & 98.80 & 52.52 & 62.44 & 69.71 & 97.89 & 99.14 \\
         & \textbf{Ours (AAPT)} & 89.51 & \textbf{66.58} & 96.22 & 96.66 & 99.19 & 42.44 & 62.09 & 71.06 & \textbf{98.60} & \textbf{99.36} \\
        \midrule
        \multirow{4}{*}{1440} & SkyReel-V2~\cite{chen2025skyreels} & 82.19 & 53.67 & 78.43 & 86.38 & 99.28 & 47.15 & 53.68 & 53.65 & 96.50 & 98.07 \\
        & MAGI-1~\cite{magi1} & 80.79 & 60.01 & 82.23 & 89.27 & 98.54 & 25.45 & 52.26 & 67.75 & \textbf{*96.90} & \textbf{*98.13} \\
        & Ours (Diffusion) & 86.65 & 60.49 & 82.38 & 89.48 & 98.29 & 66.26 & 56.46 & 64.51 & 95.01 & 97.72  \\
        & \textbf{Ours (AAPT)} & \textbf{89.79} & \textbf{62.16} & 87.15 & 89.74 & 99.11 & 76.50 & 56.77 & 67.55 & 96.11 & 97.52 \\
        \bottomrule
    \end{tabularx}
    \label{tab:eval-i2v}
    \vspace{10pt}
\end{minipage}
\begin{minipage}[t]{0.46\textwidth}
    \centering
    \captionof{table}{Quantitative comparison on pose-conditioned human video generation task. Metrics better than ours are highlighted in bold.}
    \setlength\tabcolsep{1.2pt}
    \small
    \begin{tabularx}{\linewidth}{X|ccccr}
         \toprule
         Method & AKD$\downarrow$ & IQA$\uparrow$ & ASE$\uparrow$ & FID$\downarrow$ & FVD$\downarrow$ \\
         \midrule
         DisCo & 9.313 & 3.707 & 2.396 & 57.12 & 64.52 \\
         AnimateAnyone & 5.747& 3.843 & 2.718 & 26.87 & 37.67 \\
         MimicMotion & 8.536& 3.977 & 2.842 & 23.43 & 22.97  \\
         CyberHost & 3.123& \textbf{4.087} & 2.967 & \textbf{20.04} & \textbf{7.72}  \\
         OmniHuman-1 & \textbf{2.136} & \textbf{4.111} & \textbf{2.986} & \textbf{19.50} & \textbf{7.32}  \\
         \midrule
         \textbf{Ours (AAPT)} & \textbf{2.740} & \textbf{4.077} & \textbf{2.973} & \textbf{22.43} & \textbf{11.78}  \\
         \bottomrule
    \end{tabularx}
    \label{tab:eval-pose}

    \centering
    \small
    \vspace{10pt}
    \setlength\tabcolsep{2pt}
    \begin{tabularx}{\linewidth}{|X|XXXX|}
        Input & Generated & & & \\
    \end{tabularx}
    \setlength\tabcolsep{0.5pt}
    \begin{tabularx}{\linewidth}{XXXXX}
        \includegraphics[width=\linewidth]{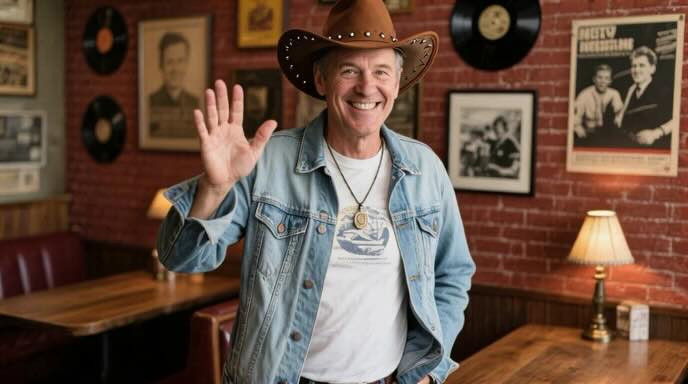} &
        \includegraphics[width=\linewidth]{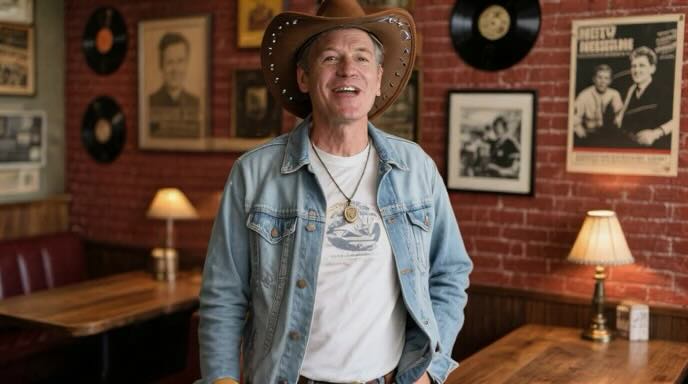} &
        \includegraphics[width=\linewidth]{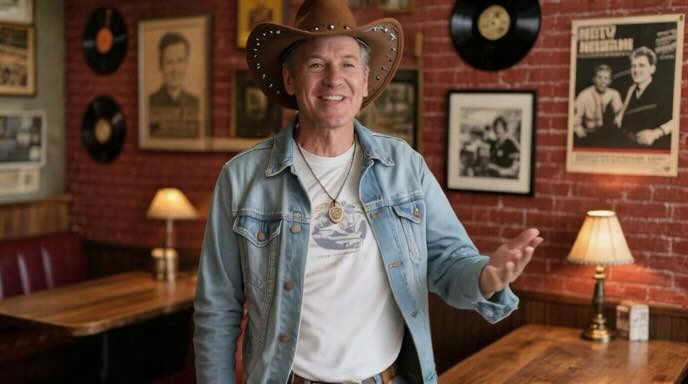} &
        \includegraphics[width=\linewidth]{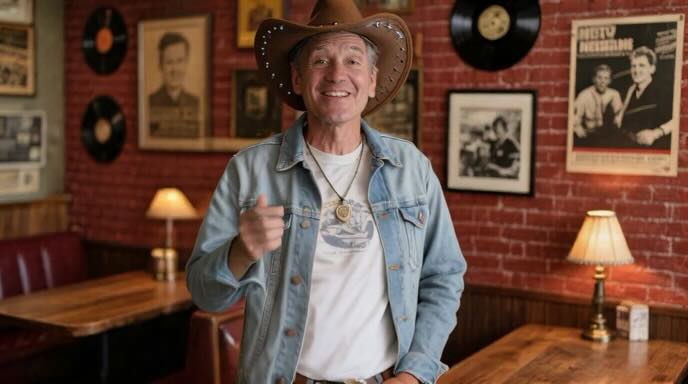} &
        \includegraphics[width=\linewidth]{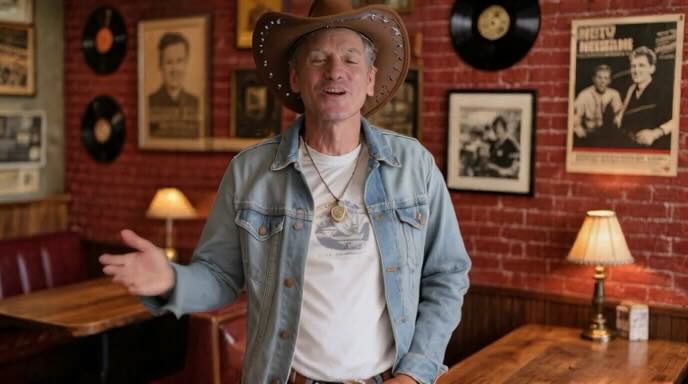} \\
        \includegraphics[width=\linewidth]{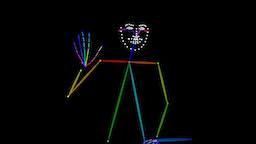} &
        \includegraphics[width=\linewidth]{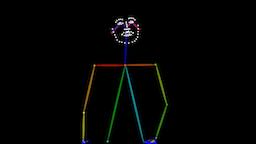} &
        \includegraphics[width=\linewidth]{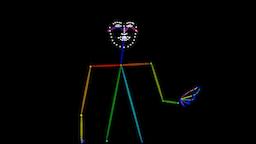} &
        \includegraphics[width=\linewidth]{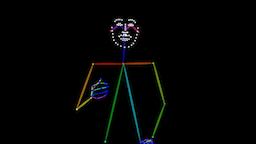} &
        \includegraphics[width=\linewidth]{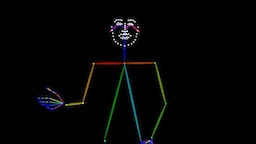} \\

    \end{tabularx}
    \captionof{figure}{Pose-conditioned virtual human}
    \label{fig:eval-pose}
\end{minipage}
\hfill
\begin{minipage}[t]{0.52\textwidth}
    \centering
    \setlength\tabcolsep{1.3pt}
    \small
    \captionof{table}{Quantitative comparison on camera-conditioned world exploration task. Metrics better than ours are highlighted in bold.}
    \begin{tabularx}{\linewidth}{X|rccccc}
        \toprule
        Method & FVD$\downarrow$ & Mov$\uparrow$ & Trans$\downarrow$ & Rot$\downarrow$ & Geo$\uparrow$ & Apr$\uparrow$ \\
        \midrule
        MotionCtrl & 221.23 & 102.21 & 0.3221 & 2.78 & 57.87 & 0.7431 \\
        CameraCtrl & 199.53 & 133.37 &  0.2812 & 2.81 & 52.12 & 0.7784\\
        CameraCtrl2 & 73.11 & \textbf{698.51} & 0.1527 & \textbf{1.58} & \textbf{88.70} & 0.8893 \\
        \midrule
        \textbf{Ours (AAPT)} & \textbf{61.33} & \textbf{521.23} & \textbf{0.1185} & \textbf{1.63} & \textbf{81.25} & \textbf{0.9012} \\
        \bottomrule
    \end{tabularx}
    \label{tab:eval-camera}

    \centering
    \small
    \vspace{10pt}
    \setlength\tabcolsep{2pt}
    \begin{tabularx}{\linewidth}{|X|XXXX|X|}
        Input & Generated & & & & Control \\
    \end{tabularx}
    \setlength\tabcolsep{0.5pt}
    \begin{tabularx}{\linewidth}{XXXXXX}

        \includegraphics[width=\linewidth]{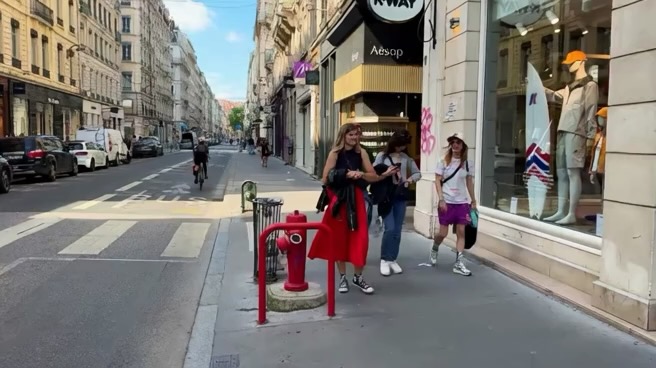} &
        \includegraphics[width=\linewidth]{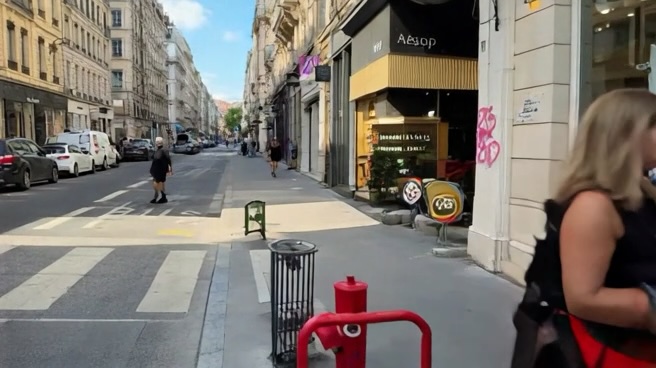} &
        \includegraphics[width=\linewidth]{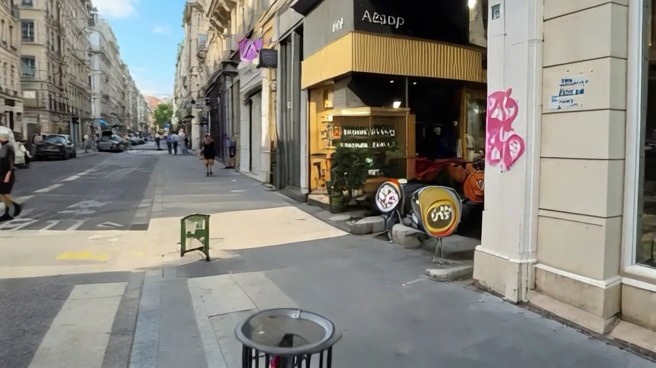} &
        \includegraphics[width=\linewidth]{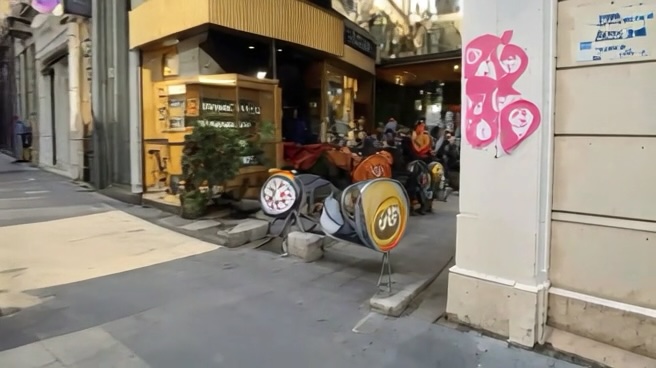} &
        \includegraphics[width=\linewidth]{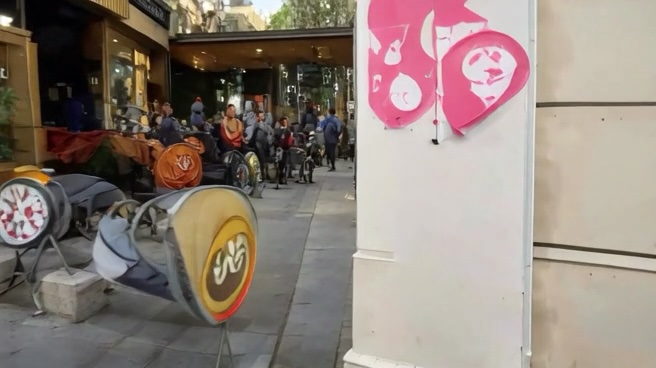} &
        \includegraphics[width=\linewidth]{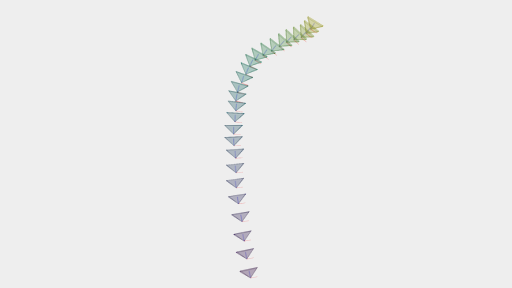} \\

        \includegraphics[width=\linewidth]{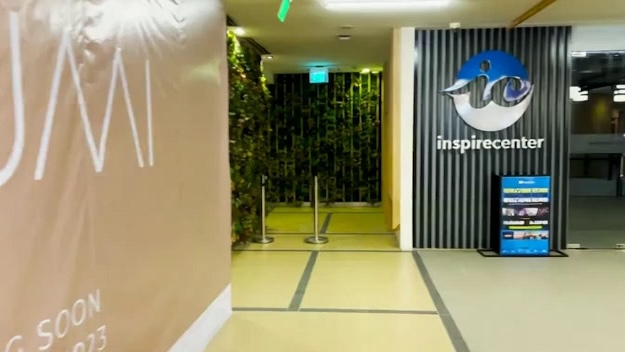} &
        \includegraphics[width=\linewidth]{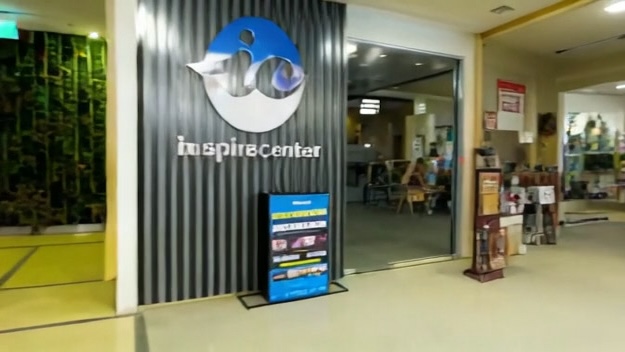} &
        \includegraphics[width=\linewidth]{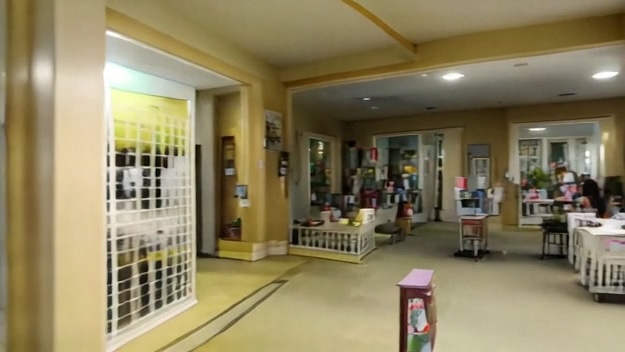} &
        \includegraphics[width=\linewidth]{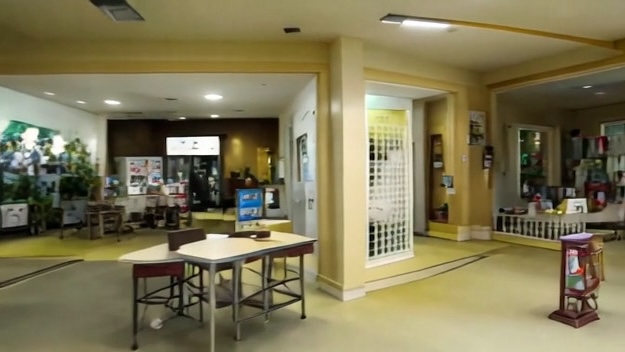} &
        \includegraphics[width=\linewidth]{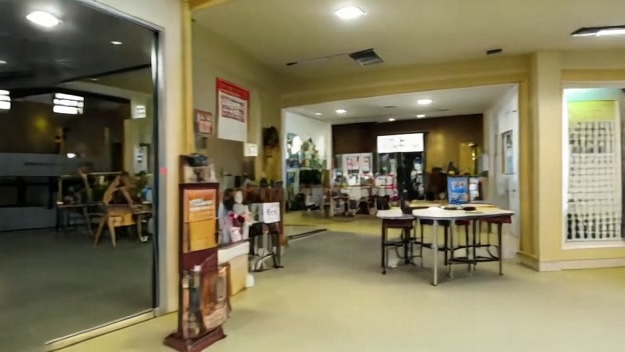} &
        \includegraphics[width=\linewidth]{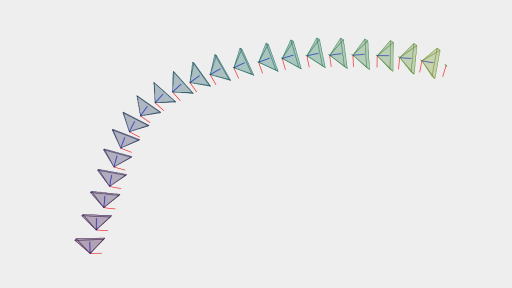} \\

        \includegraphics[width=\linewidth]{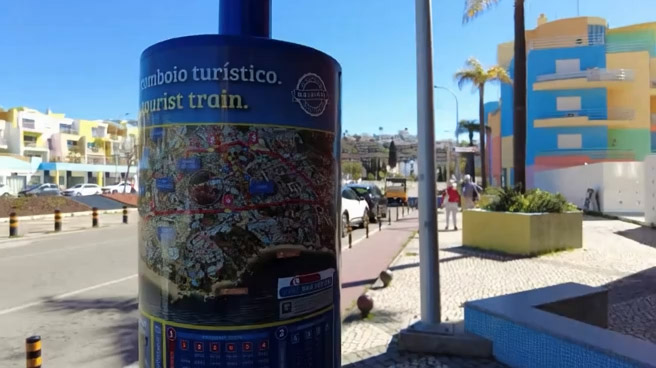} &
        \includegraphics[width=\linewidth]{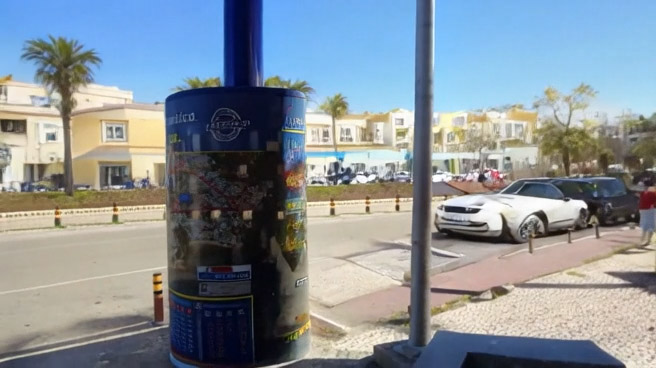} &
        \includegraphics[width=\linewidth]{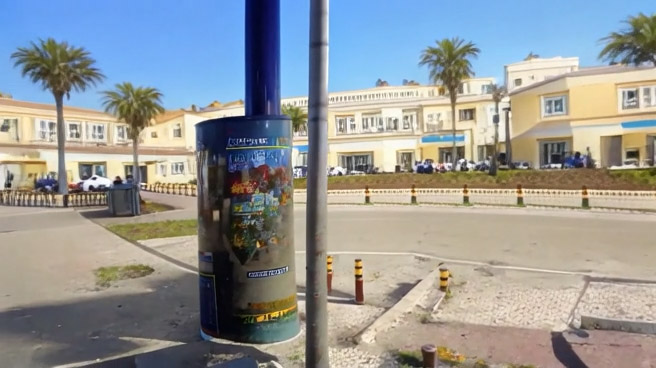} &
        \includegraphics[width=\linewidth]{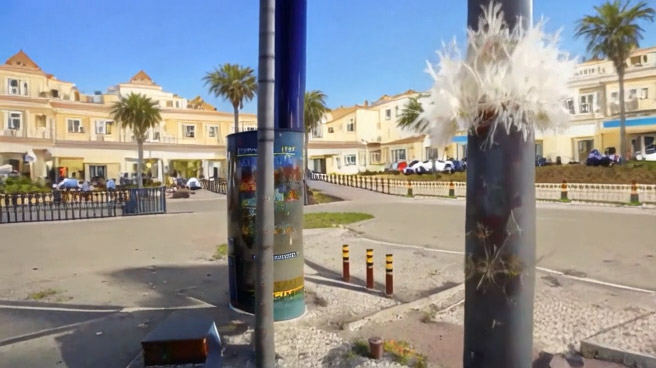} &
        \includegraphics[width=\linewidth]{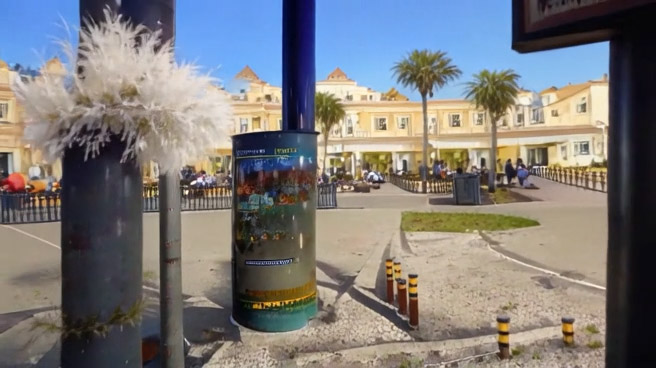} &
        \includegraphics[width=\linewidth]{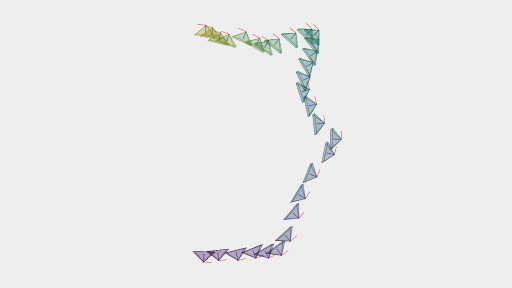} \\
        
    \end{tabularx}
    \captionof{figure}{Camera-controlled world exploration}
    \label{fig:eval-camera}
\end{minipage}

\vspace{-8pt}
\paragraph{Pose-Conditioned Human Video Generation}

We evaluate our method on post-conditioned human video generation using the protocol and test set from previous work~\cite{lin2024cyberhost}. The pose accuracy is assessed via average keypoint distance (AKD) with keypoints extracted using DWPose~\cite{yang2023dwpose}. For visual quality, we use Q-Align~\cite{wu2023q}, a vision-language model, to evaluate image quality (IQA) and aesthetics (ASE). Additionally, Fréchet Inception Distance (FID)~\cite{heusel2017gans} and Frechet Video Distance (FVD)~\cite{unterthiner2019fvd} measure the distributional alignment between the generated and the ground-truth samples. For comparison, we include four recent UNet-based diffusion models, \ie Disco~\cite{wang2024disco}, AnimateAnyone~\cite{hu2024animate}, MimicMotion~\cite{zhang2024mimicmotion}, CyberHost~\cite{lin2024cyberhost}, and the state-of-the-art DiT-based OmniHuman-1~\cite{lin2025omnihuman}. \Cref{tab:eval-pose} presents the quantitative metrics. Among the six compared methods, ours is strong in pose accuracy and is ranked second only after the state-of-the-art baseline OmniHuman-1. In terms of visual quality, ours is consistently ranked second or third and is closely after CyberHost. \Cref{fig:eval-pose} shows visualization of our method.

\vspace{-5pt}
\paragraph{Camera-Conditioned World Exploration} We verify our method on the camera-conditioned world exploration task, also following the protocol of previous work~\cite{he2025cameractrl}. We compute the FVD, movement strength (Mov), translational error (Trans), rotational error (Rot), geometric consistency (Geo), and appearance consistency (Apr). The details are provided in the supplementary materials. We compare against previous state-of-the-arts, \ie MotionCtrl~\cite{wang2024motionctrl}, can CameraCtrl 1 \& 2~\cite{he2024cameractrl,he2025cameractrl}. \Cref{tab:eval-camera} shows that our method achieves new state-of-the-art in three out of six metrics and closely follows CameraCtrl2 for the rest.

\begin{wraptable}[5]{r}{0.5\textwidth}
    \centering
    \vspace{-10pt}
    \caption{Latency and throughput comparison.}
    \small
    \setlength\tabcolsep{1pt}
    \begin{tabularx}{\linewidth}{Xcccc|cr}
        \toprule
        Method & Params & H100 & Resolution & NFE & Latency & FPS \\
        \midrule
        CausVid & 5B & 1$\times$ & 640$\times$352 & 4 & 1.30s & 9.4 \\
        \textbf{Ours} & \textbf{8B} & \textbf{1$\times$} & \textbf{736$\times$416} & \textbf{1} & \textbf{0.16s} & \textbf{24.8} \\
        \midrule
        MAGI-1 & 24B & 8$\times$ & 736 $\times$416 & 8 & 7.00s & 3.43 \\
        SkyReelV2 & 14B & 8$\times$ & 960$\times$544 & 60 & 4.50s & 0.89 \\
        \textbf{Ours} & \textbf{8B} & \textbf{8$\times$} & \textbf{1280$\times$720} & \textbf{1} & \textbf{0.17s} & \textbf{24.2} \\
        \bottomrule
    \end{tabularx}
    \label{tab:speed}
\end{wraptable}

\vspace{-5pt}
\paragraph{Inference speed} We compare the throughput and latency of our method to other streaming video generation methods in \cref{tab:speed}. Our method is significantly faster while achieving performance comparable to the state of the arts.
\section{Ablation Studies}

\begin{wraptable}[10]{r}{0.28\textwidth}
    \vspace{-10pt}
    \centering
    \small
    \captionof{table}{One-minute generation performance using different training durations.}
    \setlength\tabcolsep{2pt}
    \begin{tabularx}{\linewidth}{c|cc}
        \toprule
        \makecell{Training\\Duration} & \makecell{Temporal\\Quality} & \makecell{Frame\\Quality} \\
        \midrule
        10s & 85.86 & 57.92 \\
        20s & 85.60 & 65.69 \\
        60s & 89.79 & 62.16 \\
        \bottomrule
    \end{tabularx}
    \label{tab:ablation-long}
\end{wraptable}

\paragraph{Long Video Training} \Cref{tab:ablation-long} reports VBench-I2V metrics on models trained with different durations for one-minute video generation. Specifically, the model trained for 60s significantly outperforms the model trained for only 10s, showing the effectiveness of long video training. Visualization is provided in \cref{fig:qualitative-cross-model-nolong}.

% \paragraph{Sliding Window}

% During training, we train the model to support different sliding window sizes by uniformly sampling the size $N\sim\mathcal{U}(10, 30)$ for 480p resolution and $N\sim\mathcal{U}(10, 15)$ for 720p in each training iteration, where $N$ denotes the number of latent frames. This corresponds to durations ranging from 1.5 to 5 seconds for 480p and 1.5 to 2.5 seconds for 720p. At inference, we observe that using a larger window size leads to better generation quality and less content drift, while using a smaller window size is faster in computation. Our final model is evaluated using the largest sliding window size, \ie 30 for 480p and 15 for 720p. We do not attempt larger window sizes because the KV cache can no longer fit in the H100 GPU memory.

% We also train the model to always attend to the first frame with 50\% probability. At inference, always attending the first frame reduces content drift for long video generation and is used as our final setting.

% Our experiments only adopt the basic sliding window for simplicity. We leave the exploration of more sophisticated windowing~\cite{willette2024training}, more efficient attention~\cite{shazeer2019fast,liu2024deepseek}, KV-offloading~\cite{magi1}, and the possibility of using state-space modeling~\cite{gu2021efficiently,gu2023mamba} to future works.

\begin{wrapfigure}[11]{r}{0.5\textwidth}
    \vspace{-15pt}
    \centering
    \small
    \setlength\tabcolsep{2pt}
    \begin{tabularx}{\linewidth}{|X|X|X|X|}
        0s (Input) & 1s & 2s & 5s
    \end{tabularx}
    \setlength\tabcolsep{0.5pt}
    \begin{tabularx}{\linewidth}{XXXX}
        \includegraphics[width=\linewidth]{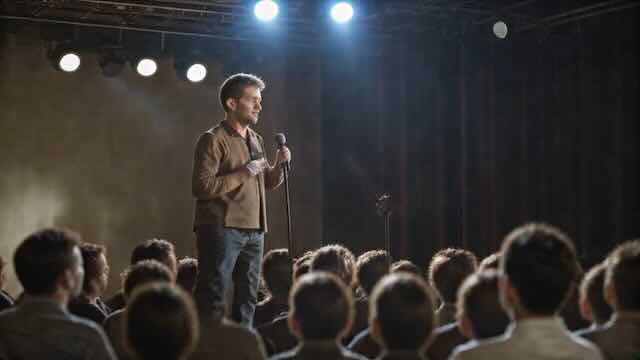} &
        \includegraphics[width=\linewidth]{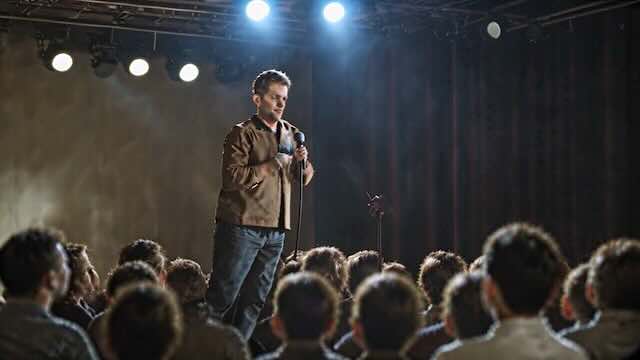} &
        \includegraphics[width=\linewidth]{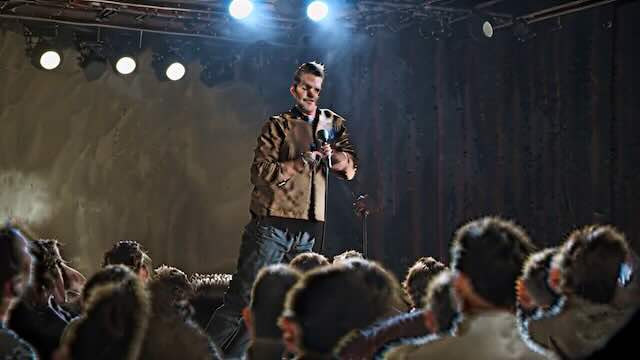} &
        \includegraphics[width=\linewidth]{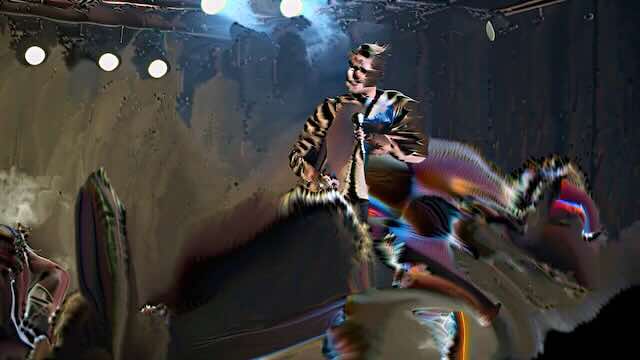} \\
        \includegraphics[width=\linewidth]{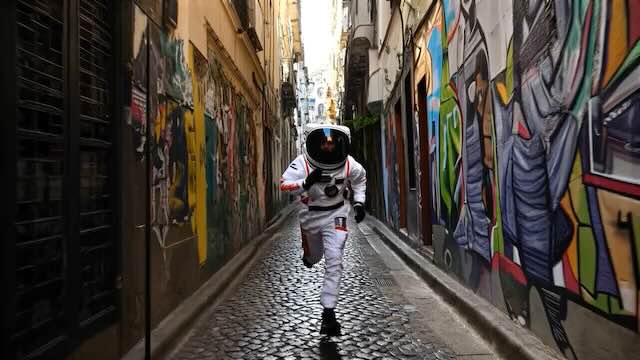} &
        \includegraphics[width=\linewidth]{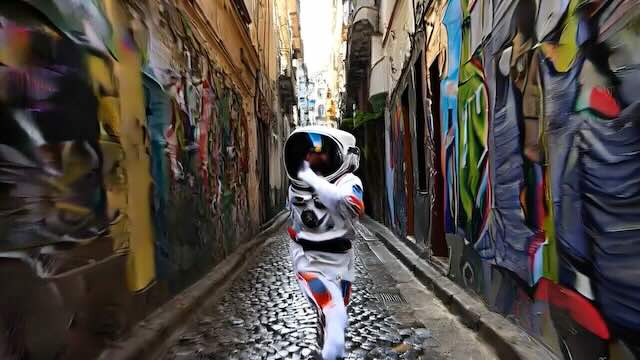} &
        \includegraphics[width=\linewidth]{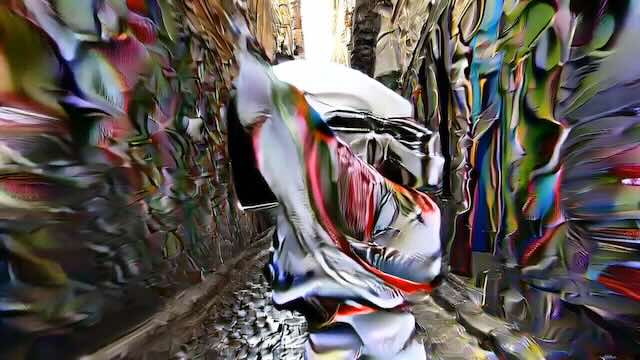} &
        \includegraphics[width=\linewidth]{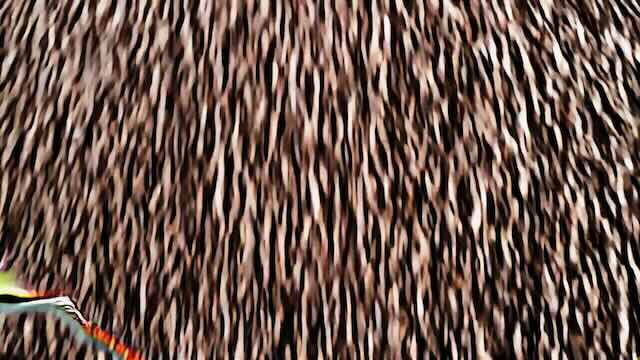} \\
    \end{tabularx}
    \caption{Models trained with teacher-forcing adversarial objective fail to generate proper content at inference.}
    \label{fig:ablation-teacherforcing}
\end{wrapfigure}

\vspace{-5pt}
\paragraph{Teacher-Forcing and Student-Forcing}

Although diffusion adaptation and consistency distillation only support teacher forcing, adversarial training can be done in either teacher-forcing or student-forcing fashion. We describe the setup in the supplementary materials.

% Specifically, in the student forcing mode, the generator runs autoregressively with KV cache to recycle inputs, and the discriminator evaluates the results in parallel. In the teacher forcing mode, the generator takes ground-truth video frames as inputs to predict next frames in parallel, and the discriminator runs autoregressively and always uses the KV cache from the real videos to attend to the ground-truth past frames.

We find that models trained with teacher-forcing adversarial objective fail to generate proper videos at inference time, as shown in \Cref{fig:ablation-teacherforcing}. The content starts to drift significantly only a few frames into the generation process. Student-forcing training is critical in mitigating error accumulation. Although prior work has found that adding Gaussian noise to the input at training can reduce drifting at inference~\cite{valevski2024diffusion}, it does not resolve the distribution gap from a fundamental level as student-forcing training. We leave additional explorations to future works.

\vspace{-5pt}
\paragraph{Limitations} For consistency, our model can have difficulty maintaining the subject and the scene. This is caused by both the generator and the discriminator. Our generator adopts the basic sliding window for simplicity. We leave the exploration of more architectures and optimizations~\cite{willette2024training,shazeer2019fast,liu2024deepseek,gu2021efficiently,gu2023mamba} to future works. For the discriminator, current segment-based discrimination cannot enforce long-range consistency. This may be mitigated by adding identity embeddings~\cite{liu2025phantom} to the discriminator. For training speed, the long video training process can be slow. For quality, we find that one-step generation can still create defects, and once the defects emerge, they can be kept in the scene for a long time since the discriminator also enforces temporal consistency. More research is needed to improve the quality of one-step generation. For the duration, we test our model on zero-shot five-minute generation. Our model can still generate content but with artifacts. We provide examples in the supplementary material.

% \begin{table}[b]
%     \centering
%     \small
%     \captionof{table}{Caption}
%     \begin{tabularx}{\linewidth}{X|cc}
%         \toprule
%         \makecell{Attention} & \makecell{Temporal\\Quality} & \makecell{Frame\\Quality} \\
%         \midrule
%         No First & 86.24 & 57.21 \\
%         Win15 & 90.34 & 63.80 \\
%         Win30 & 89.79 & 62.16 \\
%         \bottomrule
%     \end{tabularx}
%     \label{tab:my_label}
% \end{table}

\section{Conclusion}
We have introduced autoregressive adversarial post-training (AAPT), a method that uses adversarial training as a paradigm to transform video diffusion models into a fast autoregressive generator suitable for real-time interactive applications. Our model achieves performance comparable to that of the best methods while being significantly more efficient. We also analyze its limitations and aim to address them in future work.

\section*{Acknowledgment and Disclosure of Funding}
We thank Weihao Ye for assistance with the evaluation. We thank Zuquan Song and Junru Zheng for assistance with the computing infrastructure. We thank Jianyi Wang and Zhijie Lin for their discussions during the work. This work is fully funded by ByteDance Seed.

% \clearpage

\small
\bibliographystyle{plain}
\bibliography{main}

\begin{thebibliography}{100}

\bibitem{achiam2023gpt}
Josh Achiam, Steven Adler, Sandhini Agarwal, Lama Ahmad, Ilge Akkaya, Florencia~Leoni Aleman, Diogo Almeida, Janko Altenschmidt, Sam Altman, Shyamal Anadkat, et~al.
\newblock Gpt-4 technical report.
\newblock {\em arXiv preprint arXiv:2303.08774}, 2023.

\bibitem{alonso2024diffusion}
Eloi Alonso, Adam Jelley, Vincent Micheli, Anssi Kanervisto, Amos~J Storkey, Tim Pearce, and Fran{\c{c}}ois Fleuret.
\newblock Diffusion for world modeling: Visual details matter in atari.
\newblock {\em Advances in Neural Information Processing Systems}, 37:58757--58791, 2024.

\bibitem{brooks2022generating}
Tim Brooks, Janne Hellsten, Miika Aittala, Ting-Chun Wang, Timo Aila, Jaakko Lehtinen, Ming-Yu Liu, Alexei Efros, and Tero Karras.
\newblock Generating long videos of dynamic scenes.
\newblock {\em Advances in Neural Information Processing Systems}, 35:31769--31781, 2022.

\bibitem{brooks2024video}
Tim Brooks, Bill Peebles, Connor Holmes, Will DePue, Yufei Guo, Li~Jing, David Schnurr, Joe Taylor, Troy Luhman, Eric Luhman, et~al.
\newblock Video generation models as world simulators.
\newblock {\em OpenAI Blog}, 1:8, 2024.

\bibitem{brown2020language}
Tom Brown, Benjamin Mann, Nick Ryder, Melanie Subbiah, Jared~D Kaplan, Prafulla Dhariwal, Arvind Neelakantan, Pranav Shyam, Girish Sastry, Amanda Askell, et~al.
\newblock Language models are few-shot learners.
\newblock {\em Advances in neural information processing systems}, 33:1877--1901, 2020.

\bibitem{bruce2024genie}
Jake Bruce, Michael~D Dennis, Ashley Edwards, Jack Parker-Holder, Yuge Shi, Edward Hughes, Matthew Lai, Aditi Mavalankar, Richie Steigerwald, Chris Apps, et~al.
\newblock Genie: Generative interactive environments.
\newblock In {\em Forty-first International Conference on Machine Learning}, 2024.

\bibitem{chen2024diffusion}
Boyuan Chen, Diego Mart{\'\i}~Mons{\'o}, Yilun Du, Max Simchowitz, Russ Tedrake, and Vincent Sitzmann.
\newblock Diffusion forcing: Next-token prediction meets full-sequence diffusion.
\newblock {\em Advances in Neural Information Processing Systems}, 37:24081--24125, 2024.

\bibitem{chen2024nitrofusion}
Dar-Yen Chen, Hmrishav Bandyopadhyay, Kai Zou, and Yi-Zhe Song.
\newblock Nitrofusion: High-fidelity single-step diffusion through dynamic adversarial training.
\newblock {\em arXiv preprint arXiv:2412.02030}, 2024.

\bibitem{chen2025skyreels}
Guibin Chen, Dixuan Lin, Jiangping Yang, Chunze Lin, Juncheng Zhu, Mingyuan Fan, Hao Zhang, Sheng Chen, Zheng Chen, Chengchen Ma, et~al.
\newblock Skyreels-v2: Infinite-length film generative model.
\newblock {\em arXiv preprint arXiv:2504.13074}, 2025.

\bibitem{chen2025janus}
Xiaokang Chen, Zhiyu Wu, Xingchao Liu, Zizheng Pan, Wen Liu, Zhenda Xie, Xingkai Yu, and Chong Ruan.
\newblock Janus-pro: Unified multimodal understanding and generation with data and model scaling.
\newblock {\em arXiv preprint arXiv:2501.17811}, 2025.

\bibitem{chen2025dove}
Zheng Chen, Zichen Zou, Kewei Zhang, Xiongfei Su, Xin Yuan, Yong Guo, and Yulun Zhang.
\newblock Dove: Efficient one-step diffusion model for real-world video super-resolution.
\newblock {\em arXiv preprint arXiv:2505.16239}, 2025.

\bibitem{cho2023treating}
Suhwan Cho, Minhyeok Lee, Seunghoon Lee, Chaewon Park, Donghyeong Kim, and Sangyoun Lee.
\newblock Treating motion as option to reduce motion dependency in unsupervised video object segmentation.
\newblock In {\em Proceedings of the IEEE/CVF winter conference on applications of computer vision}, pages 5140--5149, 2023.

\bibitem{oasis2024}
Etched Decart, Quinn McIntyre, Spruce Campbell, Xinlei Chen, and Robert Wachen.
\newblock Oasis: A universe in a transformer.
\newblock {\em URL: https://oasis-model. github. io}, 2024.

\bibitem{esser2024scaling}
Patrick Esser, Sumith Kulal, Andreas Blattmann, Rahim Entezari, Jonas M{\"u}ller, Harry Saini, Yam Levi, Dominik Lorenz, Axel Sauer, Frederic Boesel, et~al.
\newblock Scaling rectified flow transformers for high-resolution image synthesis.
\newblock In {\em Forty-first international conference on machine learning}, 2024.

\bibitem{fasthunyuan}
{FastHunyuan}.
\newblock \url{https://huggingface.co/FastVideo/FastHunyuan}.

\bibitem{feng2024matrix}
Ruili Feng, Han Zhang, Zhantao Yang, Jie Xiao, Zhilei Shu, Zhiheng Liu, Andy Zheng, Yukun Huang, Yu~Liu, and Hongyang Zhang.
\newblock The matrix: Infinite-horizon world generation with real-time moving control.
\newblock {\em arXiv preprint arXiv:2412.03568}, 2024.

\bibitem{gao2024matten}
Yu~Gao, Jiancheng Huang, Xiaopeng Sun, Zequn Jie, Yujie Zhong, and Lin Ma.
\newblock Matten: Video generation with mamba-attention.
\newblock {\em arXiv preprint arXiv:2405.03025}, 2024.

\bibitem{goodfellow2014generative}
Ian~J Goodfellow, Jean Pouget-Abadie, Mehdi Mirza, Bing Xu, David Warde-Farley, Sherjil Ozair, Aaron Courville, and Yoshua Bengio.
\newblock Generative adversarial nets.
\newblock {\em Advances in neural information processing systems}, 27, 2014.

\bibitem{grattafiori2024llama}
Aaron Grattafiori, Abhimanyu Dubey, Abhinav Jauhri, Abhinav Pandey, Abhishek Kadian, Ahmad Al-Dahle, Aiesha Letman, Akhil Mathur, Alan Schelten, Alex Vaughan, et~al.
\newblock The llama 3 herd of models.
\newblock {\em arXiv preprint arXiv:2407.21783}, 2024.

\bibitem{gu2023mamba}
Albert Gu and Tri Dao.
\newblock Mamba: Linear-time sequence modeling with selective state spaces.
\newblock {\em arXiv preprint arXiv:2312.00752}, 2023.

\bibitem{gu2021efficiently}
Albert Gu, Karan Goel, and Christopher R{\'e}.
\newblock Efficiently modeling long sequences with structured state spaces.
\newblock {\em arXiv preprint arXiv:2111.00396}, 2021.

\bibitem{guo2025mineworld}
Junliang Guo, Yang Ye, Tianyu He, Haoyu Wu, Yushu Jiang, Tim Pearce, and Jiang Bian.
\newblock Mineworld: a real-time and open-source interactive world model on minecraft.
\newblock {\em arXiv preprint arXiv:2504.08388}, 2025.

\bibitem{guo2025long}
Yuwei Guo, Ceyuan Yang, Ziyan Yang, Zhibei Ma, Zhijie Lin, Zhenheng Yang, Dahua Lin, and Lu~Jiang.
\newblock Long context tuning for video generation.
\newblock {\em arXiv preprint arXiv:2503.10589}, 2025.

\bibitem{he2024cameractrl}
Hao He, Yinghao Xu, Yuwei Guo, Gordon Wetzstein, Bo~Dai, Hongsheng Li, and Ceyuan Yang.
\newblock Cameractrl: Enabling camera control for text-to-video generation.
\newblock {\em arXiv preprint arXiv:2404.02101}, 2024.

\bibitem{he2025cameractrl}
Hao He, Ceyuan Yang, Shanchuan Lin, Yinghao Xu, Meng Wei, Liangke Gui, Qi~Zhao, Gordon Wetzstein, Lu~Jiang, and Hongsheng Li.
\newblock Cameractrl ii: Dynamic scene exploration via camera-controlled video diffusion models.
\newblock {\em arXiv preprint arXiv:2503.10592}, 2025.

\bibitem{henschel2024streamingt2v}
Roberto Henschel, Levon Khachatryan, Hayk Poghosyan, Daniil Hayrapetyan, Vahram Tadevosyan, Zhangyang Wang, Shant Navasardyan, and Humphrey Shi.
\newblock Streamingt2v: Consistent, dynamic, and extendable long video generation from text.
\newblock {\em arXiv preprint arXiv:2403.14773}, 2024.

\bibitem{heusel2017gans}
Martin Heusel, Hubert Ramsauer, Thomas Unterthiner, Bernhard Nessler, and Sepp Hochreiter.
\newblock Gans trained by a two time-scale update rule converge to a local nash equilibrium.
\newblock {\em Advances in neural information processing systems}, 30, 2017.

\bibitem{ho2020denoising}
Jonathan Ho, Ajay Jain, and Pieter Abbeel.
\newblock Denoising diffusion probabilistic models.
\newblock {\em Advances in neural information processing systems}, 33:6840--6851, 2020.

\bibitem{ho2022classifier}
Jonathan Ho and Tim Salimans.
\newblock Classifier-free diffusion guidance.
\newblock {\em arXiv preprint arXiv:2207.12598}, 2022.

\bibitem{hu2024animate}
Li~Hu.
\newblock Animate anyone: Consistent and controllable image-to-video synthesis for character animation.
\newblock In {\em Proceedings of the IEEE/CVF Conference on Computer Vision and Pattern Recognition}, pages 8153--8163, 2024.

\bibitem{huang2024gan}
Nick Huang, Aaron Gokaslan, Volodymyr Kuleshov, and James Tompkin.
\newblock The gan is dead; long live the gan! a modern gan baseline.
\newblock {\em Advances in Neural Information Processing Systems}, 37:44177--44215, 2024.

\bibitem{huang2024vbench}
Ziqi Huang, Yinan He, Jiashuo Yu, Fan Zhang, Chenyang Si, Yuming Jiang, Yuanhan Zhang, Tianxing Wu, Qingyang Jin, Nattapol Chanpaisit, et~al.
\newblock Vbench: Comprehensive benchmark suite for video generative models.
\newblock In {\em Proceedings of the IEEE/CVF Conference on Computer Vision and Pattern Recognition}, pages 21807--21818, 2024.

\bibitem{jacobs2023deepspeed}
Sam~Ade Jacobs, Masahiro Tanaka, Chengming Zhang, Minjia Zhang, Shuaiwen~Leon Song, Samyam Rajbhandari, and Yuxiong He.
\newblock Deepspeed ulysses: System optimizations for enabling training of extreme long sequence transformer models.
\newblock {\em arXiv preprint arXiv:2309.14509}, 2023.

\bibitem{jiang2024loopy}
Jianwen Jiang, Chao Liang, Jiaqi Yang, Gaojie Lin, Tianyun Zhong, and Yanbo Zheng.
\newblock Loopy: Taming audio-driven portrait avatar with long-term motion dependency.
\newblock In {\em The Thirteenth International Conference on Learning Representations}, 2025.

\bibitem{jin2024pyramidal}
Yang Jin, Zhicheng Sun, Ningyuan Li, Kun Xu, Hao Jiang, Nan Zhuang, Quzhe Huang, Yang Song, Yadong Mu, and Zhouchen Lin.
\newblock Pyramidal flow matching for efficient video generative modeling.
\newblock {\em arXiv preprint arXiv:2410.05954}, 2024.

\bibitem{jolicoeur2018relativistic}
Alexia Jolicoeur-Martineau.
\newblock The relativistic discriminator: a key element missing from standard gan.
\newblock {\em arXiv preprint arXiv:1807.00734}, 2018.

\bibitem{kang2024distilling}
Minguk Kang, Richard Zhang, Connelly Barnes, Sylvain Paris, Suha Kwak, Jaesik Park, Eli Shechtman, Jun-Yan Zhu, and Taesung Park.
\newblock Distilling diffusion models into conditional gans.
\newblock In {\em European Conference on Computer Vision}, pages 428--447. Springer, 2024.

\bibitem{karnewar2020msg}
Animesh Karnewar and Oliver Wang.
\newblock Msg-gan: Multi-scale gradients for generative adversarial networks.
\newblock In {\em Proceedings of the IEEE/CVF conference on computer vision and pattern recognition}, pages 7799--7808, 2020.

\bibitem{karras2017progressive}
Tero Karras, Timo Aila, Samuli Laine, and Jaakko Lehtinen.
\newblock Progressive growing of gans for improved quality, stability, and variation.
\newblock {\em arXiv preprint arXiv:1710.10196}, 2017.

\bibitem{kim2024fifo}
Jihwan Kim, Junoh Kang, Jinyoung Choi, and Bohyung Han.
\newblock Fifo-diffusion: Generating infinite videos from text without training.
\newblock {\em arXiv preprint arXiv:2405.11473}, 2024.

\bibitem{kodaira2023streamdiffusion}
Akio Kodaira, Chenfeng Xu, Toshiki Hazama, Takanori Yoshimoto, Kohei Ohno, Shogo Mitsuhori, Soichi Sugano, Hanying Cho, Zhijian Liu, and Kurt Keutzer.
\newblock Streamdiffusion: A pipeline-level solution for real-time interactive generation.
\newblock {\em arXiv preprint arXiv:2312.12491}, 2023.

\bibitem{kohler2024imagine}
Jonas Kohler, Albert Pumarola, Edgar Sch{\"o}nfeld, Artsiom Sanakoyeu, Roshan Sumbaly, Peter Vajda, and Ali Thabet.
\newblock Imagine flash: Accelerating emu diffusion models with backward distillation.
\newblock {\em arXiv preprint arXiv:2405.05224}, 2024.

\bibitem{kondratyuk2023videopoet}
Dan Kondratyuk, Lijun Yu, Xiuye Gu, Jos{\'e} Lezama, Jonathan Huang, Grant Schindler, Rachel Hornung, Vighnesh Birodkar, Jimmy Yan, Ming-Chang Chiu, et~al.
\newblock Videopoet: A large language model for zero-shot video generation.
\newblock {\em arXiv preprint arXiv:2312.14125}, 2023.

\bibitem{kong2024hunyuanvideo}
Weijie Kong, Qi~Tian, Zijian Zhang, Rox Min, Zuozhuo Dai, Jin Zhou, Jiangfeng Xiong, Xin Li, Bo~Wu, Jianwei Zhang, et~al.
\newblock Hunyuanvideo: A systematic framework for large video generative models.
\newblock {\em arXiv preprint arXiv:2412.03603}, 2024.

\bibitem{lin2024cyberhost}
Gaojie Lin, Jianwen Jiang, Chao Liang, Tianyun Zhong, Jiaqi Yang, Zerong Zheng, and Yanbo Zheng.
\newblock Cyberhost: A one-stage diffusion framework for audio-driven talking body generation.
\newblock In {\em The Thirteenth International Conference on Learning Representations}, 2025.

\bibitem{lin2025omnihuman}
Gaojie Lin, Jianwen Jiang, Jiaqi Yang, Zerong Zheng, and Chao Liang.
\newblock Omnihuman-1: Rethinking the scaling-up of one-stage conditioned human animation models.
\newblock {\em arXiv preprint arXiv:2502.01061}, 2025.

\bibitem{lin2024common}
Shanchuan Lin, Bingchen Liu, Jiashi Li, and Xiao Yang.
\newblock Common diffusion noise schedules and sample steps are flawed.
\newblock In {\em Proceedings of the IEEE/CVF winter conference on applications of computer vision}, pages 5404--5411, 2024.

\bibitem{lin2024sdxl}
Shanchuan Lin, Anran Wang, and Xiao Yang.
\newblock Sdxl-lightning: Progressive adversarial diffusion distillation.
\newblock {\em arXiv preprint arXiv:2402.13929}, 2024.

\bibitem{lin2025diffusion}
Shanchuan Lin, Xin Xia, Yuxi Ren, Ceyuan Yang, Xuefeng Xiao, and Lu~Jiang.
\newblock Diffusion adversarial post-training for one-step video generation.
\newblock {\em arXiv preprint arXiv:2501.08316}, 2025.

\bibitem{lin2024animatediff}
Shanchuan Lin and Xiao Yang.
\newblock Animatediff-lightning: Cross-model diffusion distillation.
\newblock {\em arXiv preprint arXiv:2403.12706}, 2024.

\bibitem{lipman2022flow}
Yaron Lipman, Ricky~TQ Chen, Heli Ben-Hamu, Maximilian Nickel, and Matt Le.
\newblock Flow matching for generative modeling.
\newblock {\em arXiv preprint arXiv:2210.02747}, 2022.

\bibitem{liu2024deepseek}
Aixin Liu, Bei Feng, Bing Xue, Bingxuan Wang, Bochao Wu, Chengda Lu, Chenggang Zhao, Chengqi Deng, Chenyu Zhang, Chong Ruan, et~al.
\newblock Deepseek-v3 technical report.
\newblock {\em arXiv preprint arXiv:2412.19437}, 2024.

\bibitem{liu2024timestep}
Feng Liu, Shiwei Zhang, Xiaofeng Wang, Yujie Wei, Haonan Qiu, Yuzhong Zhao, Yingya Zhang, Qixiang Ye, and Fang Wan.
\newblock Timestep embedding tells: It's time to cache for video diffusion model.
\newblock {\em arXiv preprint arXiv:2411.19108}, 2024.

\bibitem{liu2025phantom}
Lijie Liu, Tianxiang Ma, Bingchuan Li, Zhuowei Chen, Jiawei Liu, Qian He, and Xinglong Wu.
\newblock Phantom: Subject-consistent video generation via cross-modal alignment.
\newblock {\em arXiv preprint arXiv:2502.11079}, 2025.

\bibitem{liu2022flow}
Xingchao Liu, Chengyue Gong, and Qiang Liu.
\newblock Flow straight and fast: Learning to generate and transfer data with rectified flow.
\newblock {\em arXiv preprint arXiv:2209.03003}, 2022.

\bibitem{liu2023instaflow}
Xingchao Liu, Xiwen Zhang, Jianzhu Ma, Jian Peng, et~al.
\newblock Instaflow: One step is enough for high-quality diffusion-based text-to-image generation.
\newblock In {\em The Twelfth International Conference on Learning Representations}, 2023.

\bibitem{loshchilov2017decoupled}
Ilya Loshchilov and Frank Hutter.
\newblock Decoupled weight decay regularization.
\newblock {\em arXiv preprint arXiv:1711.05101}, 2017.

\bibitem{lu2024simplifying}
Cheng Lu and Yang Song.
\newblock Simplifying, stabilizing and scaling continuous-time consistency models.
\newblock {\em arXiv preprint arXiv:2410.11081}, 2024.

\bibitem{lu2022dpm}
Cheng Lu, Yuhao Zhou, Fan Bao, Jianfei Chen, Chongxuan Li, and Jun Zhu.
\newblock Dpm-solver: A fast ode solver for diffusion probabilistic model sampling in around 10 steps.
\newblock {\em Advances in Neural Information Processing Systems}, 35:5775--5787, 2022.

\bibitem{lu2022dpmpp}
Cheng Lu, Yuhao Zhou, Fan Bao, Jianfei Chen, Chongxuan Li, and Jun Zhu.
\newblock Dpm-solver++: Fast solver for guided sampling of diffusion probabilistic models.
\newblock {\em arXiv preprint arXiv:2211.01095}, 2022.

\bibitem{luo2023diff}
Weijian Luo, Tianyang Hu, Shifeng Zhang, Jiacheng Sun, Zhenguo Li, and Zhihua Zhang.
\newblock Diff-instruct: A universal approach for transferring knowledge from pre-trained diffusion models.
\newblock {\em Advances in Neural Information Processing Systems}, 36:76525--76546, 2023.

\bibitem{luo2025dreamactor}
Yuxuan Luo, Zhengkun Rong, Lizhen Wang, Longhao Zhang, Tianshu Hu, and Yongming Zhu.
\newblock Dreamactor-m1: Holistic, expressive and robust human image animation with hybrid guidance.
\newblock {\em arXiv preprint arXiv:2504.01724}, 2025.

\bibitem{ma2024learning}
Xinyin Ma, Gongfan Fang, Michael Bi~Mi, and Xinchao Wang.
\newblock Learning-to-cache: Accelerating diffusion transformer via layer caching.
\newblock {\em Advances in Neural Information Processing Systems}, 37:133282--133304, 2024.

\bibitem{mao2024osv}
Xiaofeng Mao, Zhengkai Jiang, Fu-Yun Wang, Jiangning Zhang, Hao Chen, Mingmin Chi, Yabiao Wang, and Wenhan Luo.
\newblock Osv: One step is enough for high-quality image to video generation.
\newblock {\em arXiv preprint arXiv:2409.11367}, 2024.

\bibitem{mescheder2018training}
Lars Mescheder, Andreas Geiger, and Sebastian Nowozin.
\newblock Which training methods for gans do actually converge?
\newblock In {\em International conference on machine learning}, pages 3481--3490. PMLR, 2018.

\bibitem{mo2024scaling}
Shentong Mo and Yapeng Tian.
\newblock Scaling diffusion mamba with bidirectional ssms for efficient image and video generation.
\newblock {\em arXiv preprint arXiv:2405.15881}, 2024.

\bibitem{parkerholder2024genie2}
Jack Parker-Holder, Philip Ball, Jake Bruce, Vibhavari Dasagi, Kristian Holsheimer, Christos Kaplanis, Alexandre Moufarek, Guy Scully, Jeremy Shar, Jimmy Shi, Stephen Spencer, Jessica Yung, Michael Dennis, Sultan Kenjeyev, Shangbang Long, Vlad Mnih, Harris Chan, Maxime Gazeau, Bonnie Li, Fabio Pardo, Luyu Wang, Lei Zhang, Frederic Besse, Tim Harley, Anna Mitenkova, Jane Wang, Jeff Clune, Demis Hassabis, Raia Hadsell, Adrian Bolton, Satinder Singh, and Tim Rockt{\"a}schel.
\newblock Genie 2: A large-scale foundation world model.
\newblock 2024.

\bibitem{peebles2023scalable}
William Peebles and Saining Xie.
\newblock Scalable diffusion models with transformers.
\newblock In {\em Proceedings of the IEEE/CVF international conference on computer vision}, pages 4195--4205, 2023.

\bibitem{polyak2024movie}
Adam Polyak, Amit Zohar, Andrew Brown, Andros Tjandra, Animesh Sinha, Ann Lee, Apoorv Vyas, Bowen Shi, Chih-Yao Ma, Ching-Yao Chuang, et~al.
\newblock Movie gen: A cast of media foundation models.
\newblock {\em arXiv preprint arXiv:2410.13720}, 2024.

\bibitem{radford2021learning}
Alec Radford, Jong~Wook Kim, Chris Hallacy, Aditya Ramesh, Gabriel Goh, Sandhini Agarwal, Girish Sastry, Amanda Askell, Pamela Mishkin, Jack Clark, et~al.
\newblock Learning transferable visual models from natural language supervision.
\newblock In {\em International conference on machine learning}, pages 8748--8763. PmLR, 2021.

\bibitem{ren2025next}
Shuhuai Ren, Shuming Ma, Xu~Sun, and Furu Wei.
\newblock Next block prediction: Video generation via semi-auto-regressive modeling.
\newblock {\em arXiv preprint arXiv:2502.07737}, 2025.

\bibitem{ren2024hyper}
Yuxi Ren, Xin Xia, Yanzuo Lu, Jiacheng Zhang, Jie Wu, Pan Xie, Xing Wang, and Xuefeng Xiao.
\newblock Hyper-sd: Trajectory segmented consistency model for efficient image synthesis.
\newblock {\em arXiv preprint arXiv:2404.13686}, 2024.

\bibitem{roth2017stabilizing}
Kevin Roth, Aurelien Lucchi, Sebastian Nowozin, and Thomas Hofmann.
\newblock Stabilizing training of generative adversarial networks through regularization.
\newblock {\em Advances in neural information processing systems}, 30, 2017.

\bibitem{salimans2022progressive}
Tim Salimans and Jonathan Ho.
\newblock Progressive distillation for fast sampling of diffusion models.
\newblock {\em arXiv preprint arXiv:2202.00512}, 2022.

\bibitem{magi1}
Sand-AI.
\newblock Magi-1: Autoregressive video generation at scale, 2025.

\bibitem{sauer2024fast}
Axel Sauer, Frederic Boesel, Tim Dockhorn, Andreas Blattmann, Patrick Esser, and Robin Rombach.
\newblock Fast high-resolution image synthesis with latent adversarial diffusion distillation.
\newblock In {\em SIGGRAPH Asia 2024 Conference Papers}, pages 1--11, 2024.

\bibitem{sauer2024adversarial}
Axel Sauer, Dominik Lorenz, Andreas Blattmann, and Robin Rombach.
\newblock Adversarial diffusion distillation.
\newblock In {\em European Conference on Computer Vision}, pages 87--103. Springer, 2024.

\bibitem{seawead2025seaweed}
Team Seawead, Ceyuan Yang, Zhijie Lin, Yang Zhao, Shanchuan Lin, Zhibei Ma, Haoyuan Guo, Hao Chen, Lu~Qi, Sen Wang, et~al.
\newblock Seaweed-7b: Cost-effective training of video generation foundation model.
\newblock {\em arXiv preprint arXiv:2504.08685}, 2025.

\bibitem{shah2024flashattention}
Jay Shah, Ganesh Bikshandi, Ying Zhang, Vijay Thakkar, Pradeep Ramani, and Tri Dao.
\newblock Flashattention-3: Fast and accurate attention with asynchrony and low-precision.
\newblock {\em Advances in Neural Information Processing Systems}, 37:68658--68685, 2024.

\bibitem{shazeer2019fast}
Noam Shazeer.
\newblock Fast transformer decoding: One write-head is all you need.
\newblock {\em arXiv preprint arXiv:1911.02150}, 2019.

\bibitem{skorokhodov2022stylegan}
Ivan Skorokhodov, Sergey Tulyakov, and Mohamed Elhoseiny.
\newblock Stylegan-v: A continuous video generator with the price, image quality and perks of stylegan2.
\newblock In {\em Proceedings of the IEEE/CVF conference on computer vision and pattern recognition}, pages 3626--3636, 2022.

\bibitem{song2023improved}
Yang Song and Prafulla Dhariwal.
\newblock Improved techniques for training consistency models.
\newblock {\em arXiv preprint arXiv:2310.14189}, 2023.

\bibitem{song2023consistency}
Yang Song, Prafulla Dhariwal, Mark Chen, and Ilya Sutskever.
\newblock Consistency models.
\newblock 2023.

\bibitem{song2020score}
Yang Song, Jascha Sohl-Dickstein, Diederik~P Kingma, Abhishek Kumar, Stefano Ermon, and Ben Poole.
\newblock Score-based generative modeling through stochastic differential equations.
\newblock {\em arXiv preprint arXiv:2011.13456}, 2020.

\bibitem{stypulkowski2024diffused}
Micha{\l} Stypu{\l}kowski, Konstantinos Vougioukas, Sen He, Maciej Zi{\k{e}}ba, Stavros Petridis, and Maja Pantic.
\newblock Diffused heads: Diffusion models beat gans on talking-face generation.
\newblock In {\em Proceedings of the IEEE/CVF Winter Conference on Applications of Computer Vision}, pages 5091--5100, 2024.

\bibitem{su2024roformer}
Jianlin Su, Murtadha Ahmed, Yu~Lu, Shengfeng Pan, Wen Bo, and Yunfeng Liu.
\newblock Roformer: Enhanced transformer with rotary position embedding.
\newblock {\em Neurocomputing}, 568:127063, 2024.

\bibitem{team2024chameleon}
Chameleon Team.
\newblock Chameleon: Mixed-modal early-fusion foundation models.
\newblock {\em arXiv preprint arXiv:2405.09818}, 2024.

\bibitem{teed2020raft}
Zachary Teed and Jia Deng.
\newblock Raft: Recurrent all-pairs field transforms for optical flow.
\newblock In {\em Computer Vision--ECCV 2020: 16th European Conference, Glasgow, UK, August 23--28, 2020, Proceedings, Part II 16}, pages 402--419. Springer, 2020.

\bibitem{tian2024emo}
Linrui Tian, Qi~Wang, Bang Zhang, and Liefeng Bo.
\newblock Emo: Emote portrait alive generating expressive portrait videos with audio2video diffusion model under weak conditions.
\newblock In {\em European Conference on Computer Vision}, pages 244--260. Springer, 2024.

\bibitem{tian2025training}
Ye~Tian, Xin Xia, Yuxi Ren, Shanchuan Lin, Xing Wang, Xuefeng Xiao, Yunhai Tong, Ling Yang, and Bin Cui.
\newblock Training-free diffusion acceleration with bottleneck sampling.
\newblock {\em arXiv preprint arXiv:2503.18940}, 2025.

\bibitem{unterthiner2019fvd}
Thomas Unterthiner, Sjoerd van Steenkiste, Karol Kurach, Rapha{\"e}l Marinier, Marcin Michalski, and Sylvain Gelly.
\newblock Fvd: A new metric for video generation.

\bibitem{unterthiner2018towards}
Thomas Unterthiner, Sjoerd Van~Steenkiste, Karol Kurach, Raphael Marinier, Marcin Michalski, and Sylvain Gelly.
\newblock Towards accurate generative models of video: A new metric \& challenges.
\newblock {\em arXiv preprint arXiv:1812.01717}, 2018.

\bibitem{valevski2024diffusion}
Dani Valevski, Yaniv Leviathan, Moab Arar, and Shlomi Fruchter.
\newblock Diffusion models are real-time game engines.
\newblock {\em arXiv preprint arXiv:2408.14837}, 2024.

\bibitem{vaswani2017attention}
Ashish Vaswani, Noam Shazeer, Niki Parmar, Jakob Uszkoreit, Llion Jones, Aidan~N Gomez, {\L}ukasz Kaiser, and Illia Polosukhin.
\newblock Attention is all you need.
\newblock {\em Advances in neural information processing systems}, 30, 2017.

\bibitem{wang2025wan}
Ang Wang, Baole Ai, Bin Wen, Chaojie Mao, Chen-Wei Xie, Di~Chen, Feiwu Yu, Haiming Zhao, Jianxiao Yang, Jianyuan Zeng, et~al.
\newblock Wan: Open and advanced large-scale video generative models.
\newblock {\em arXiv preprint arXiv:2503.20314}, 2025.

\bibitem{wang2023gen}
Fu-Yun Wang, Wenshuo Chen, Guanglu Song, Han-Jia Ye, Yu~Liu, and Hongsheng Li.
\newblock Gen-l-video: Multi-text to long video generation via temporal co-denoising.
\newblock {\em arXiv preprint arXiv:2305.18264}, 2023.

\bibitem{wang2024animatelcm}
Fu-Yun Wang, Zhaoyang Huang, Weikang Bian, Xiaoyu Shi, Keqiang Sun, Guanglu Song, Yu~Liu, and Hongsheng Li.
\newblock Animatelcm: Computation-efficient personalized style video generation without personalized video data.
\newblock In {\em SIGGRAPH Asia 2024 Technical Communications}, pages 1--5. 2024.

\bibitem{wang2024lingen}
Hongjie Wang, Chih-Yao Ma, Yen-Cheng Liu, Ji~Hou, Tao Xu, Jialiang Wang, Felix Juefei-Xu, Yaqiao Luo, Peizhao Zhang, Tingbo Hou, et~al.
\newblock Lingen: Towards high-resolution minute-length text-to-video generation with linear computational complexity.
\newblock {\em arXiv preprint arXiv:2412.09856}, 2024.

\bibitem{wang2025seedvr2}
Jianyi Wang, Shanchuan Lin, Zhijie Lin, Yuxi Ren, Meng Wei, Zongsheng Yue, Shangchen Zhou, Hao Chen, Yang Zhao, Ceyuan Yang, et~al.
\newblock Seedvr2: One-step video restoration via diffusion adversarial post-training.
\newblock {\em arXiv preprint arXiv:2506.05301}, 2025.

\bibitem{wang2024vggsfm}
Jianyuan Wang, Nikita Karaev, Christian Rupprecht, and David Novotny.
\newblock Vggsfm: Visual geometry grounded deep structure from motion.
\newblock In {\em Proceedings of the IEEE/CVF conference on computer vision and pattern recognition}, pages 21686--21697, 2024.

\bibitem{wang2024disco}
Tan Wang, Linjie Li, Kevin Lin, Yuanhao Zhai, Chung-Ching Lin, Zhengyuan Yang, Hanwang Zhang, Zicheng Liu, and Lijuan Wang.
\newblock Disco: Disentangled control for realistic human dance generation.
\newblock In {\em Proceedings of the IEEE/CVF Conference on Computer Vision and Pattern Recognition}, pages 9326--9336, 2024.

\bibitem{wang2024emu3}
Xinlong Wang, Xiaosong Zhang, Zhengxiong Luo, Quan Sun, Yufeng Cui, Jinsheng Wang, Fan Zhang, Yueze Wang, Zhen Li, Qiying Yu, et~al.
\newblock Emu3: Next-token prediction is all you need.
\newblock {\em arXiv preprint arXiv:2409.18869}, 2024.

\bibitem{wang2024parallelized}
Yuqing Wang, Shuhuai Ren, Zhijie Lin, Yujin Han, Haoyuan Guo, Zhenheng Yang, Difan Zou, Jiashi Feng, and Xihui Liu.
\newblock Parallelized autoregressive visual generation.
\newblock {\em arXiv preprint arXiv:2412.15119}, 2024.

\bibitem{wang2024motionctrl}
Zhouxia Wang, Ziyang Yuan, Xintao Wang, Yaowei Li, Tianshui Chen, Menghan Xia, Ping Luo, and Ying Shan.
\newblock Motionctrl: A unified and flexible motion controller for video generation.
\newblock In {\em ACM SIGGRAPH 2024 Conference Papers}, pages 1--11, 2024.

\bibitem{willette2024training}
Jeffrey Willette, Heejun Lee, Youngwan Lee, Myeongjae Jeon, and Sung~Ju Hwang.
\newblock Training-free exponential context extension via cascading kv cache.
\newblock {\em arXiv preprint arXiv:2406.17808}, 2024.

\bibitem{wu2024janus}
Chengyue Wu, Xiaokang Chen, Zhiyu Wu, Yiyang Ma, Xingchao Liu, Zizheng Pan, Wen Liu, Zhenda Xie, Xingkai Yu, Chong Ruan, et~al.
\newblock Janus: Decoupling visual encoding for unified multimodal understanding and generation.
\newblock {\em arXiv preprint arXiv:2410.13848}, 2024.

\bibitem{wu2023q}
Haoning Wu, Zicheng Zhang, Weixia Zhang, Chaofeng Chen, Liang Liao, Chunyi Li, Yixuan Gao, Annan Wang, Erli Zhang, Wenxiu Sun, et~al.
\newblock Q-align: Teaching lmms for visual scoring via discrete text-defined levels.
\newblock {\em arXiv preprint arXiv:2312.17090}, 2023.

\bibitem{xie2024progressive}
Desai Xie, Zhan Xu, Yicong Hong, Hao Tan, Difan Liu, Feng Liu, Arie Kaufman, and Yang Zhou.
\newblock Progressive autoregressive video diffusion models.
\newblock {\em arXiv preprint arXiv:2410.08151}, 2024.

\bibitem{xie2024sana}
Enze Xie, Junsong Chen, Junyu Chen, Han Cai, Haotian Tang, Yujun Lin, Zhekai Zhang, Muyang Li, Ligeng Zhu, Yao Lu, et~al.
\newblock Sana: Efficient high-resolution image synthesis with linear diffusion transformers.
\newblock {\em arXiv preprint arXiv:2410.10629}, 2024.

\bibitem{xu2024ufogen}
Yanwu Xu, Yang Zhao, Zhisheng Xiao, and Tingbo Hou.
\newblock Ufogen: You forward once large scale text-to-image generation via diffusion gans.
\newblock In {\em Proceedings of the IEEE/CVF Conference on Computer Vision and Pattern Recognition}, pages 8196--8206, 2024.

\bibitem{yan2023magicprop}
Hanshu Yan, Jun~Hao Liew, Long Mai, Shanchuan Lin, and Jiashi Feng.
\newblock Magicprop: Diffusion-based video editing via motion-aware appearance propagation.
\newblock {\em arXiv preprint arXiv:2309.00908}, 2023.

\bibitem{yan2024perflow}
Hanshu Yan, Xingchao Liu, Jiachun Pan, Jun~Hao Liew, Qiang Liu, and Jiashi Feng.
\newblock Perflow: Piecewise rectified flow as universal plug-and-play accelerator.
\newblock {\em arXiv preprint arXiv:2405.07510}, 2024.

\bibitem{yang2023dwpose}
Zhendong Yang, Ailing Zeng, Chun Yuan, and Yu~Li.
\newblock Effective whole-body pose estimation with two-stages distillation.
\newblock In {\em Proceedings of the IEEE/CVF International Conference on Computer Vision}, pages 4210--4220, 2023.

\bibitem{ye2025fast}
Yang Ye, Junliang Guo, Haoyu Wu, Tianyu He, Tim Pearce, Tabish Rashid, Katja Hofmann, and Jiang Bian.
\newblock Fast autoregressive video generation with diagonal decoding.
\newblock {\em arXiv preprint arXiv:2503.14070}, 2025.

\bibitem{yin2024improved}
Tianwei Yin, Micha{\"e}l Gharbi, Taesung Park, Richard Zhang, Eli Shechtman, Fredo Durand, and Bill Freeman.
\newblock Improved distribution matching distillation for fast image synthesis.
\newblock {\em Advances in neural information processing systems}, 37:47455--47487, 2024.

\bibitem{yin2024one}
Tianwei Yin, Micha{\"e}l Gharbi, Richard Zhang, Eli Shechtman, Fredo Durand, William~T Freeman, and Taesung Park.
\newblock One-step diffusion with distribution matching distillation.
\newblock In {\em Proceedings of the IEEE/CVF conference on computer vision and pattern recognition}, pages 6613--6623, 2024.

\bibitem{yin2024slow}
Tianwei Yin, Qiang Zhang, Richard Zhang, William~T Freeman, Fredo Durand, Eli Shechtman, and Xun Huang.
\newblock From slow bidirectional to fast causal video generators.
\newblock {\em arXiv preprint arXiv:2412.07772}, 2024.

\bibitem{yu2023language}
Lijun Yu, Jos{\'e} Lezama, Nitesh~B Gundavarapu, Luca Versari, Kihyuk Sohn, David Minnen, Yong Cheng, Vighnesh Birodkar, Agrim Gupta, Xiuye Gu, et~al.
\newblock Language model beats diffusion--tokenizer is key to visual generation.
\newblock {\em arXiv preprint arXiv:2310.05737}, 2023.

\bibitem{zhang2024sageattention2}
Jintao Zhang, Haofeng Huang, Pengle Zhang, Jia Wei, Jun Zhu, and Jianfei Chen.
\newblock Sageattention2 technical report: Accurate 4 bit attention for plug-and-play inference acceleration.
\newblock {\em arXiv preprint arXiv:2411.10958}, 2024.

\bibitem{zhang2024sageattention}
Jintao Zhang, Haofeng Huang, Pengle Zhang, Jun Zhu, Jianfei Chen, et~al.
\newblock Sageattention: Accurate 8-bit attention for plug-and-play inference acceleration.
\newblock {\em arXiv preprint arXiv:2410.02367}, 2024.

\bibitem{zhang2025fast}
Peiyuan Zhang, Yongqi Chen, Runlong Su, Hangliang Ding, Ion Stoica, Zhenghong Liu, and Hao Zhang.
\newblock Fast video generation with sliding tile attention.
\newblock {\em arXiv preprint arXiv:2502.04507}, 2025.

\bibitem{zhang2024mimicmotion}
Yuang Zhang, Jiaxi Gu, Li-Wen Wang, Han Wang, Junqi Cheng, Yuefeng Zhu, and Fangyuan Zou.
\newblock Mimicmotion: High-quality human motion video generation with confidence-aware pose guidance.
\newblock {\em arXiv preprint arXiv:2406.19680}, 2024.

\bibitem{zhang2024sf}
Zhixing Zhang, Yanyu Li, Yushu Wu, Anil Kag, Ivan Skorokhodov, Willi Menapace, Aliaksandr Siarohin, Junli Cao, Dimitris Metaxas, Sergey Tulyakov, et~al.
\newblock Sf-v: Single forward video generation model.
\newblock {\em Advances in Neural Information Processing Systems}, 37:103599--103618, 2024.

\bibitem{zhao2023pytorch}
Yanli Zhao, Andrew Gu, Rohan Varma, Liang Luo, Chien-Chin Huang, Min Xu, Less Wright, Hamid Shojanazeri, Myle Ott, Sam Shleifer, et~al.
\newblock Pytorch fsdp: experiences on scaling fully sharded data parallel.
\newblock {\em arXiv preprint arXiv:2304.11277}, 2023.

\bibitem{zou2024accelerating}
Chang Zou, Xuyang Liu, Ting Liu, Siteng Huang, and Linfeng Zhang.
\newblock Accelerating diffusion transformers with token-wise feature caching.
\newblock {\em arXiv preprint arXiv:2410.05317}, 2024.

\end{thebibliography}

\clearpage
\appendix

\section{Model Architecture}

\paragraph{Diffusion Transformer} Our diffusion transformer largely follows the MMDiT design~\cite{esser2024scaling}. It has 8B parameters and 36 transformer blocks. The discriminator adopts the same architecture. Therefore, our generator and our discriminator consist of 16B parameters for the adversarial training.

\paragraph{Block Causal Attention} We implement block causal attention using Flash Attention 3~\cite{shah2024flashattention} in a for-loop. We find it to provide reasonable performance for training. We leave the exploration for more performance implementation to future work. For inference, recurrent autoregressive steps are taken, and Flash Attention 3 can be naturally adopted without a performance penalty.

\paragraph{Positional Embedding} As the duration of the generation becomes agnostic to our causal architecture, we modify the 3D rotary positional embeddings (RoPE)~\cite{su2024roformer}. Specifically, the positional embeddings continue to stretch dynamically along the spatial dimension to help the model generalize to different resolutions, while the positional embeddings are changed to have a fixed interval along the temporal dimension to support arbitrary lengths of training and generation.

\paragraph{Parallelism} We adopt FSDP~\cite{zhao2023pytorch} for data parallelism. We use ZERO 2 for the generator during student-forcing training that requires recurrent forward calls to avoid repeated parameter gathering, and ZERO 3 for all other modules to save memory. We also adopt Ulysses~\cite{jacobs2023deepspeed} as our context parallel strategy. We shard each video sample across 8 GPUs. Gradient checkpointing is also utilized per transformer block to fit the memory requirement.
\section{Training Details}

\paragraph{Diffusion Adaptation} After changing the architecture to block causal attention and adding the recycled input channels, we first adapt the model with diffusion training.

We follow the original model to use the flow-matching parameterization \cite{lipman2022flow}. Specifically, given sample $x_0$ and noise $\epsilon$, input is derived through linear interpolation $x_t = (1 - t) \cdot x_0 + t \cdot \epsilon$. The diffusion timestep is sampled uniformly $t\sim\mathcal{U}(0, 1)$, then passed through a shifting function $\mathrm{shift}(t, s) := (s \times t) / (1 + (s-1) \times t)$, where $s=24$. Note that the same timestep is used for the entire clip without the diffusion-forcing~\cite{chen2024diffusion} approach of assigning independent timesteps for each frame. Our model predicts the velocity $v=\epsilon - x_0$ and is penalized with the mean squared error loss. We apply the teacher-forcing paradigm and provide the ground-truth frames without noise as recycled input. The noisy input and the output target are shifted by one frame to facilitate next frame prediction.

We use AdamW optimizer \cite{loshchilov2017decoupled} with a learning rate of 1e-5 and a weight decay scale of 0.01 throughout the process. We first train on 736$\times$416 (equivalent to 640$\times$480 by area) 5-second videos for 20k iterations with a batch size of 256. Then, we add 1280$\times$720 to the mix for another 6k iterations with a batch size of 128. Finally, we turn up the maximum duration of 736$\times$416 resolution videos to 15 seconds for 4k iterations with a batch size of 32. This curriculum allows our model to see enough samples in the early stages and see longer samples in the final stage.

\paragraph{Consistency Distillation} Then we apply consistency distillation~\cite{song2023improved} to create a one-step generator. Although the results after consistency distillation are blurry, it provides a better initialization for the adversarial training stage, as discovered by APT~\cite{lin2025diffusion}.

We inherit the same AdamW settings and the dataset settings as in the last diffusion adaptation stage. We distill the model on 32 fixed steps, which are uniformly selected and then passed through the shifting function with a shifting factor $s=24$. We do not apply classifier-free guidance~\cite{ho2022classifier}. We continue to use the teacher-forcing paradigm to provide ground-truth frames as recycled inputs, and shift the noisy inputs and output targets by one frame following the diffusion adaptation approach. We follow the improved consistency distillation technique~\cite{song2023improved} and do not apply exponential moving average on the consistency target. No additional modification is needed for consistency distillation. The model is trained for 5k iterations.

\paragraph{Adversarial Training} Finally, we perform adversarial training. In this stage, we switch to the student-forcing paradigm, where the generator only takes the first frame as input and recycles the actual generated frame for the next autoregressive step, strictly following the inference procedure. Then, the discriminator evaluates the generated results in parallel, producing logits after each frame for multi-duration discrimination.

We follow APT~\cite{lin2025diffusion} to initialize the generator from the consistency distillation weights, and to initialize the discriminator from the diffusion adaptation weight. We change to use the relativistic pairing loss~\cite{jolicoeur2018relativistic}:
\begin{equation}
    \mathcal{L}_{RpGAN}(x_0, \epsilon) = f(D(G(\epsilon, c),c) - D(x_0,c)),
\end{equation}
where $G$,$D$ denote the generator and the discriminator respectively, $f_G(x)=-\log(1+e^{-x})$ or $f_D(x)=-\log(1+e^{x})$ is used each of their update steps respectively, $c$ denotes the text condition and other interactive conditions. We calculate R1 and R2 regularization~\cite{roth2017stabilizing,mescheder2018training} through the approximation technique proposed in APT~\cite{lin2025diffusion}:

\begin{equation}
    \mathcal{L}_{aR1} = \lambda \| D(x_0, c) - D(\mathcal{N}(x_0, \sigma \textbf{I}), c) \|^2_2,
\end{equation}
\begin{equation}
    \mathcal{L}_{aR2} = \lambda \| D(G(\epsilon, c), c) - D(\mathcal{N}(G(\epsilon, c), \sigma \textbf{I}), c) \|^2_2,
\end{equation}
where $\epsilon = 0.1$ and $\lambda=1000$. Since the discriminator is initialized from the diffusion model, we follow APT to provide timesteps by random uniform sampling $t~\sim\mathcal{U}(0, 1)$. We do not shift the timestep for the discriminator. We use RMSProp optimizer with $\alpha=0.9$ following APT~\cite{lin2025diffusion}.

We first perform training without the long-video extension training technique. The videos are 5s to 10s in duration. We train it using a low learning rate of 3e-6 following APT~\cite{lin2025diffusion} and a batch size of 256 for 500 generator updates. The resulting model can only generate up to 10 seconds and will drift for videos longer than 10 seconds.

Then we apply the long video training technique. The training videos are still from 5s to 10s, and we extend it once with an overlap of 1s to a total maximum duration of 19s (10 + (10-1)). This stage is trained for 500 generator updates. Then we turn up the extension to 5 times, to a total maximum duration of 55s (10 + 5$\times$(10-1)). We find it necessary to decrease the batch size to 64 and increase the learning rate to 1e-5 for the extension training for the model to make adequate changes in a reasonable amount of time.

Since the generator in student-forcing mode must recurrently perform model forward for each autoregressive step during training, we switch FSDP to ZERO 2 mode to save all the model parameters on each machine. This avoids repeated parameter gathering and improves the training seed. The discriminator and text encoder still adopt ZERO 3 to shard all the model parameters for memory saving.

\paragraph{Computational Resources} We use 256 H100 GPUs for our final training and employ gradient accumulation where necessary to reach our final batch size. The model is trained in approximately 7 days, where the diffusion adaptation and the long-video adversarial training take the majority of the time.
\section{Variational Autoencoder}

We train a lightweight VAE decoder to fit the real-time budget. Specifically, our original VAE decoder has 3 residual blocks per resolution scale, and has channels [128, 256, 512, 512] at each resolution scale. Our lightweight VAE decoder reduces the number of residual blocks per resolution to 2, and reduces the channels to [64, 128, 256, 512]. This results in nearly 3 times speed-up without visible quality degradation.

\section{Teacher-Forcing Adversarial Training}

The adversarial training supports both student-forcing and teacher-forcing modes. To implement student forcing, the generator runs autoregressively with KV cache and recycles the actual generated frame as input for the next autoregressive step. The discriminator evaluates the results in parallel. To implement teacher forcing, the generator takes ground-truth video frames as past prediction inputs and predicts the next frames in parallel. The discriminator runs autoregressively and always uses the KV cache from the real videos to attend to the ground-truth past frames.

\begin{wrapfigure}{r}{0.5\linewidth}
    \centering
    \includegraphics[width=\linewidth]{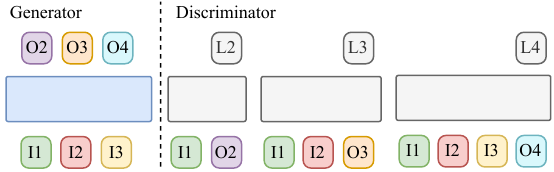}
    \caption{Teacher-forcing adversarial training}
    \label{fig:appendix-teacher-forcing}
\end{wrapfigure}

\Cref{fig:appendix-teacher-forcing} visualizes teacher-forcing adversarial training. Specifically, in teacher-forcing mode, the generator given input $I1,I2,I3$ generates independent output $O2,O3,O4$. Namely, the output $O3$ only has a correlation with $I2$ but not with $O2$. Therefore, the discriminator must independently evaluate the generated results with their correct dependencies to produce logits $L2,L3,L4$. Since the discriminator transformer is causal, the repeated computation can be saved using KV cache.

We have conducted experiments with teacher-forcing adversarial training, and the model fails to generate reasonable videos as discussed in the main paper. We suspect LLMs are able to train with teacher-forcing mode because they use a discrete codebook to encode words, where slight inaccuracy is less relevant. But our model predicts continuous latent values for the entire frame, where slight inaccuracy accumulates.

\section{The Importance of Result Recycling}

We conduct an experiment to study the importance of result recycling. Specifically, we keep the exact architecture and training settings, and we mask the recycled input as zero tensors, except the first frame, which takes in the user image. We find that models trained without recycling input cannot generate large motion. Some of the movements become incohesive as well. The video visualization is provided on our website.

\section{I2V Evaluation}

The table in the main text compares our model under the 736$\times$416 setting. For the other models we compare to, we largely follow the default sampling setting for each model, including the number of steps and CFG~\cite{ho2022classifier}. We also use the default resolution for each model to ensure that the model has been properly trained on the expected resolution. Specifically, we use 896$\times$544 for Hunyuan~\cite{kong2024hunyuanvideo}, 832$\times$464 for Wan2.1~\cite{wang2025wan}, 960$\times$544 for SkyReel-V2~\cite{chen2025skyreels}. We note that we run 5 samples per prompt for all the comparisons per VBench-I2V~\cite{huang2024vbench} requirement, except for SkyReel-V2 which we only run 1 sample per prompt and reduce the sampling steps from its default 50 to 30. This is because SkyReel-V2 is too computationally intensive to generate one-minute videos.

We additionally provide the evaluation metrics under the 1280$\times$720 resolution in \cref{tab:ablation-i2v}. Note that 1280$\times$720 is trained and inference with a smaller attention window size $N=15$ to fit the memory.

\begin{table}[h]
    \centering
    \captionof{table}{Quantitative VBench-I2V~\cite{huang2024vbench} metrics on 1280$\times$720 compared to 736$\times$416.}
    \small
    \setlength\tabcolsep{2pt}
    \begin{tabularx}{\linewidth}{cXr|rc|cccccc|rr}
        \toprule
        & & &  \multicolumn{8}{c|}{Quality} & \multicolumn{2}{c}{Condition} \\
        Frames & Method & Resolution & \tiny\textbf{\makecell{Temporal\\Quality}} & \tiny\textbf{\makecell{Frame\\Quality}}  & \tiny\makecell{Subject\\Consistency} & \tiny\makecell{Background\\Consistency} & \tiny\makecell{Motion\\Smoothness} & \tiny\makecell{Dynamic\\Degree} & \tiny\makecell{Aesthetic\\Quality} & \tiny\makecell{Imaging\\Quality} & \tiny\textbf{\makecell{I2V\\Subject}} & \tiny\textbf{\makecell{I2V\\Background}} \\
        \midrule
        \multirow{2}{*}{1440} & \multirow{2}{*}{Ours} & 736$\times$416 & \textbf{89.79} & 62.16 & 87.15 & 89.74 & 99.11 & 76.50 & 56.77 & 67.55 & 96.11 & 97.52 \\
        & & 1280$\times$720 & 88.24 & \textbf{64.30} & 87.95 & 90.10 & 99.16 & 63.29 & 57.79 & 70.80 & \textbf{96.51} & \textbf{98.18} \\
        \bottomrule
    \end{tabularx}
    \label{tab:ablation-i2v}
\end{table}

\section{Camera-Conditioned World Exploration}

\paragraph{Training} We make a few modifications on CameraCtrl II~\cite{he2025cameractrl} to make it better support causal generation. First, CameraCtrl II uses Plücker embeddings to represent the camera position and orientation, where it treats the first frame as the initial position, and the other frames are relative to the first frame. This is problematic as the value can grow unbounded if the displacement forever increases. We change it so that each frame is only relative to the previous frame. Hence, the Plücker embeddings only represent the camera changes between immediate frames to prevent unbounded growth of values. Second, CameraCtrl II uses the original Plücker coordinate to represent each camera ray, which consists of a direction vector and a moment vector. The moment vector encodes the displacement information, which is computed as the cross product of a point on the line and the direction vector. We find that this implicit representation unnecessarily increases the complexity for the model to learn. Rather, we directly encode the camera ray's origin and direction. Third, the input scaling to the model is in fact a hyperparameter that is not previously explored. We scale the coordinate inputs to roughly 1 standard deviation to simplify model learning. We also drop samples whose camera embeddings have very large values. These outliers are caused by inaccurate camera estimation and are detrimental to the stability of adversarial training. Last, we use random initialization instead of zero initialization for the input projection of the new channels. We find that random initialization helps the model to adapt to the new inputs much more quickly.

The camera-conditioned model is trained separately from the I2V model. We start from the I2V diffusion adaptation weights and continue training on the camera-conditioned task. The consistency distillation and adversarial training are done separately for this dedicated model. The training settings are mostly the same as the I2V model. For the long-video extension training, we randomly sample new camera trajectories for the extended parts.

\paragraph{Evaluation} Our evaluation metrics follow CameraCtrl II~\cite{he2025cameractrl}. Specifically, we compute Fréchet Video Distance (FVD)~\cite{unterthiner2018towards} against the ground-truth videos. We compute the movement strength (Mov) on RAFT-extracted~\cite{teed2020raft} dense optical flow of foreground objects identified by TMO-generated~\cite{cho2023treating} segmentation masks. Translational (Trans) and rotational (Rot) errors are computed by comparing estimated camera parameters using VGGSfM~\cite{wang2024vggsfm} with the ground truth. Geometric Consistency (Geo) is computed as the successful ratio of VGGSfM to estimate camera parameters. This indicates the quality of 3D geometry consistency of the generated scene. Appearance Consistency (Apr) is computed by comparing the cosine distance of each frame's CLIP~\cite{radford2021learning} vision embedding to the average of the entire video clip.
\section{Societal Impacts}

Our work proposes a new approach for real-time streaming video generation for interactive applications. Our approach is faster and more computationally efficient than existing approaches. This potentially enables the adoption of more real-time interactive applications. We do not consider our work to bring risk for significant negative societal impacts. The videos generated by our method still contain imperfections that are easy to identify as generated videos, which prevents the technology from being used for malicious purposes.

\end{document}